\documentclass[final,5p,times,twocolumn,authoryear]{elsarticle}
\usepackage{float}          
\usepackage{pifont}         
\usepackage{threeparttable} 
\usepackage{titlesec}       
\usepackage{geometry}       
\usepackage{amssymb}
\usepackage{amsmath}
\usepackage{hyperref} 
\usepackage{microtype}

\usepackage{booktabs}   
\usepackage{multirow}   
\usepackage{graphicx}   

\usepackage{tabularx}   
\usepackage{makecell}   
\usepackage{ragged2e}   
\usepackage{orcidlink}

\usepackage{titlesec}

\newcommand{\cmark}{\ding{51}}
\newcommand{\xmark}{\ding{55}}

\newcolumntype{L}[1]{>{\raggedright\arraybackslash}m{#1}}
\newcolumntype{C}[1]{>{\centering\arraybackslash}m{#1}}

\titleformat{\paragraph}[block]
  {\normalfont\normalsize\itshape}
  {\theparagraph}
  {1em}
  {}

\titlespacing*{\paragraph}
  {0pt}                             
  {2.0ex plus 1ex minus .2ex}       
  {0pt}                             

\journal{Medical Image Analysis}

\begin{document}
\twocolumn[{
  \begin{center}
    
    {\Large \textbf{Tracing 3D Anatomy in 2D Strokes: Can Projection-Derived Masks Replace 3D Segmentation for Cervical Spine Fracture Recognition?}}
    
    \vspace{1.5em}
    
    {\normalsize
    Fabi Nahian Madhurja\textsuperscript{1,$\dagger$}\orcidlink{0009-0005-2709-5745}, 
    Rusab Sarmun\textsuperscript{2,$\dagger$}\orcidlink{0009-0004-4887-0627}, 
    Muhammad E. H. Chowdhury\textsuperscript{3,*}\orcidlink{0000-0003-0744-8206}, \\
    Adam Mushtak\textsuperscript{4}\orcidlink{0000-0001-6409-135X}, 
    Israa Al-Hashimi\textsuperscript{4}\orcidlink{0000-0001-7901-8841}, 
    Sohaib Bassam Zoghoul\textsuperscript{4}\orcidlink{0000-0001-9024-0670}
    }
    
    \vspace{1em}
    
    {\footnotesize 
    \textsuperscript{1}Department of Computer Science and Engineering, BRAC University, Dhaka 1212, Bangladesh\\
    \textsuperscript{2}Department of Electrical and Electronic Engineering, University of Dhaka, Dhaka 1000, Bangladesh\\
    \textsuperscript{3}Department of Electrical Engineering, Qatar University, Doha 2713, Qatar\\
    \textsuperscript{4}Department of Radiology, Hamad Medical Corporation, Doha, Qatar
    }
    
    \vspace{1em}
    
    {\footnotesize
    \textsuperscript{$\dagger$}These authors contributed equally to the work.\\
    \textsuperscript{*}Corresponding author: \href{mailto:mchowdhury@qu.edu.qa}{mchowdhury@qu.edu.qa}
    }
  \end{center}

  \vspace{0.5em}

  \noindent\rule{\textwidth}{0.5pt} 
  \vspace{0.5em}
  
  \begin{center}
  \begin{minipage}{0.9\textwidth}
    \small
    \textbf{Abstract}
    
Cervical spine fractures require rapid and accurate diagnosis, yet automatic CT interpretation remains challenging as subtle injuries must be assessed across large 3D volumes. We ask whether full 3D vertebra segmentation is necessary for fracture recognition, or whether vertebra masks approximated from 2D projections preserve sufficient diagnostic context. In the proposed multi-stage pipeline, a YOLOv8 detector localizes the cervical spine from multi-view variance projections (3D mean Intersection over Union 94.45\%). Multi-label vertebra segmentation is performed with a DenseNet121-Unet on energy-based sagittal and coronal projections (mean Dice 87.86\%). The predicted 2D masks are back-projected and fused into approximate 3D masks per vertebra to extract volumes of interest from the original CT, which are analyzed by an ensemble of 2.5D spatio-sequential CNN-Transformer models. On the full RSNA 2022 cohort at native prevalence (2,019 patients, five-fold cross-validation), the pipeline yields vertebra-level and patient-level F1-scores of 68.15\% and 82.28\%, area under the receiver operating characteristic curve of 91.63\% and 90.95\%, and area under the precision-recall curve (AUPRC) of 75.64\% and 92.00\%. With the classifier architecture fixed, replacing 3D-segmentation masks by projection-derived masks changed vertebra-level and patient-level AUPRC by $+1.03$ and $+0.65$ points, and the pipeline differed from the reproduced RSNA 2022 winning solution by $-0.55$ and $+0.15$ points. No AUPRC difference was statistically significant, and fold-wise 95\% confidence intervals excluded vertebra-level losses larger than 1.46 and 2.81 points, respectively. Explainability and interobserver analyses examine reliability. Projection-based mask approximation is therefore a viable proxy for 3D vertebra segmentation in cervical fracture recognition.

    \vspace{1em}
  \end{minipage}
  \end{center}
  
  \vspace{0.5em}
  \noindent\rule{\textwidth}{0.5pt} 
  
  \vspace{3em} 
}]


\section{Introduction}
Traumatic cervical spine fractures are a clinical emergency. Injury to the cervical vertebrae (C1--C7) can compromise the underlying spinal cord, and missed or delayed diagnosis may result in permanent neurological deficits or death \citep{Copley2016, Izzo2019, Khanpara2020, Poonnoose2002}. These injuries commonly arise from high-energy mechanisms and carry high morbidity and mortality, so timely and accurate assessment is essential \citep{Inaba2016, Zanza2023, Joaquim2013, Sundstrom2014}. Supplementary Figure S2 summarizes the relevant anatomy of the cervical spine, including the atypical C1 (atlas) and C2 (axis) vertebrae \citep{Neumann2016}. Computed Tomography (CT) is the primary modality for evaluating suspected cervical trauma, owing to its high spatial resolution and diagnostic reliability \citep{Holmes2005, Schwarzenberg2020}. However, the volumetric nature of CT means that a large number of slices must be reviewed manually, increasing radiologist workload, reporting time, and the likelihood of fatigue-related errors \citep{Jalal2021, Krupinski2010, Mendoza2019, Mirvis2013, PerotteAMIA}, while degenerative changes and anatomical variants can mimic or mask fractures and contribute to missed or delayed diagnoses \citep{Small2021, Chen2025MLMissedFractures}.

To alleviate this burden, artificial intelligence (AI) and deep learning (DL) methods have been increasingly adopted to automate fracture detection \citep{HungCervical, Sutradhar2025, Bhavya2022, Gaikwad2024}. Purely 2D slice-wise models are computationally lightweight but discard the inter-slice continuity needed to resolve fractures that span multiple slices \citep{Small2021}. Full-3D volumetric models preserve complete spatial context but are computationally intensive and, given the limited annotated data typical of this domain, are prone to overfitting \citep{Singh2020ThreeDReview, Tajbakhsh2020}. Intermediate 2.5D and sequential hybrids aggregate information across selected slice stacks or vertebra-centered sequences \citep{SalehinejadISBI, HungCervical}. However, many fracture-recognition pipelines still require an intermediate localization or segmentation stage to identify the relevant vertebra before classification, most commonly implemented with object detectors or 3D segmentation networks \citep{Gaikwad2024, Golla2023, Kim2023PlatCon, HaKaggle2022}. Projection-based representations remain underexplored for cervical fracture recognition, which leaves open whether the segmentation stage can be shifted into a lower-dimensional projection domain while preserving enough anatomical context for downstream vertebra-level fracture classification. In addition, many reported models are developed and evaluated on artificially balanced datasets \citep{RasulAnalytics, Esfahani2023, Yaseen2024, Nejad2023}, which do not reflect the severe class imbalance of real-world trauma populations and can overstate generalizability.

To the best of our knowledge, this study is the first to extensively investigate the viability of 2D-projection-driven 3D-mask approximation for downstream cervical vertebra fracture recognition. The motivating question is whether the full spatial detail recovered by volumetric segmentation is necessary for accurate fracture diagnosis, or whether a compact set of optimized 2D projections can approximate the underlying 3D anatomy closely enough to support it. We therefore recast vertebra mask estimation as a multi-label segmentation problem on a small number of optimized orthogonal 2D projections, and we examine whether such approximate masks retain enough diagnostic information to support accurate downstream fracture classification.

The proposed multi-stage pipeline combines established localization, segmentation, and classification components within a largely projection-based framework, so that the source of the vertebra masks can be exchanged while the rest of the pipeline is held fixed. A YOLOv8 detector \citep{YOLOv8} first localizes the cervical spine from variance projections of the axial, sagittal, and coronal views, and the detected region is extracted as a spinal column volume of interest. A Unet-based architecture then performs multi-label segmentation of the individual vertebrae (C1--C7) on energy-based sagittal and coronal projections generated from this volume. The resulting orthogonal 2D masks are fused into an estimated 3D mask for each vertebra, so this stage operates in a lower-dimensional space than traditional 3D segmentation. Finally, the approximated masks are used to extract vertebra-level volumes of interest, which are classified for fractures by an ensemble of 2.5D spatio-sequential models. In a matched comparison with the classifier, training protocol, and folds held fixed, replacing 3D-segmentation masks with projection-derived masks did not significantly change AUPRC at either diagnostic level, nor did the pipeline differ significantly from the reproduced RSNA 2022 winning solution. We further quantify how localization and segmentation errors propagate to the diagnosis against a training-matched ground-truth ceiling, visualize the evidence used by the classifier with saliency maps, and compare its agreement with the dataset reference standard with that of three expert radiologists.

In this study, our main contributions are as follows:
\begin{itemize}
    \item \textbf{Investigation of projection-based viability for classification:} To the best of our knowledge, this is the first work to extensively investigate the diagnostic viability of utilizing 2D projection-derived segmentation masks as a proxy for 3D segmentation masks in the downstream task of fracture classification.

    \item \textbf{Multi-label segmentation on optimized projections:} We approximate 3D vertebral masks (C1--C7) by fusing multi-label segmentations of two optimized orthogonal projections, within a multi-stage pipeline that also localizes the cervical spine from projections, so that the segmentation stage operates in a lower-dimensional space than direct 3D segmentation.

    \item \textbf{Projection technique comparison:} We systematically compare projection operators for each stage of the pipeline, four for localization and 23 for segmentation, adopting variance projection for localization and energy projection for segmentation.

    \item \textbf{2.5D Spatio-Sequential Ensemble with projection stacks:}    An ensemble of two 2.5D models combines complementary features from slice stacks and from projection stacks derived from the approximated vertebral volumes, improving discrimination over either model alone.

    \item \textbf{Evaluation on the full imbalanced cohort:} We evaluate every vertebra of all 2,019 patients at the native fracture prevalence of the dataset, compare class-imbalance strategies, and show that testing the same trained models on a balanced subset inflates their precision and AUPRC.

    \item \textbf{Robustness and error-propagation analysis:} We quantify how inaccuracies in the localization and segmentation stages propagate through the multi-stage pipeline to downstream vertebra-level and patient-level fracture diagnosis, and compare the pipeline with a training-matched ground-truth ceiling.

    \item \textbf{Interobserver analysis:} We conduct an interobserver variability study with three expert radiologists and analyze agreement patterns between readers, the model, and the dataset reference standard.
\end{itemize}

\section{Related Works}
In current clinical practice, cervical spine fractures are identified through radiologist review of CT slices, a time-consuming process in which the reader must integrate information across adjacent slices while also determining which cervical vertebra is involved. These demands have motivated a growing body of artificial intelligence (AI) and deep learning (DL) research for automated cervical spine fracture analysis, which recent reviews describe as active but heterogeneous, with reported performance varying substantially across datasets, annotation levels, and evaluation protocols \citep{Liawrungrueang2025SystematicReview, Gholamrezanezhad2026MetaAnalysis}. Within this literature, vertebra localization and fracture recognition are tightly interconnected problems, because the downstream diagnostic decision is defined largely at the vertebra or patient level.

\subsection{Cervical spine fracture classification}
The most common formulation classifies fractures directly from CT, and published systems differ mainly in the unit they classify. Early CNN systems operated at the slice or volume level. Small et al., for example, applied the FDA-cleared C-Spine network to 665 examinations and reported 92\% accuracy at 76\% sensitivity \citep{Small2021}, and a range of lightweight and transfer-learning models followed, from MobileNetV2-based real-time detectors \citep{SandlerMobileNetV2, RasulAnalytics} to compact depthwise-separable networks \citep{Vel2026LightweightCNN} and, more recently, multi-backbone ensembles \citep{Kanwal2024SpineEDLNet}. Singh et al. instead classified all seven cervical vertebrae jointly with an Inception-ResNet-v2 encoder and a U-Net decoder, reporting 98.44\% accuracy with per-vertebra F1 scores between 0.935 and 0.966 on the 87-patient voxel-annotated RSNA subset \citep{Singh2025HybridUNet}. Fully volumetric 3D CNNs retain complete spatial context at higher computational cost, typically on small cohorts, as in the voxel-wise network that Nicolaes et al. trained for osteoporotic fractures in mostly abdominal CT (AUC 0.95 and 0.93 at patient and vertebra level on 90 cases) \citep{NicolaesCSI, Bhavya2022}.

A persistent problem with this group is that their headline accuracies, several at or above 99\%, are obtained on artificially balanced subsets of the RSNA 2022 dataset \citep{Lin2023RSNA} in which fractured and non-fractured cases are equally represented even in the test sets, unlike the full cohort \citep{Esfahani2023, RasulAnalytics, Vel2026LightweightCNN, Kanwal2024SpineEDLNet, Chlad2023, Yaseen2024}, a limitation also emphasized in recent reviews \citep{Liawrungrueang2025SystematicReview, Gholamrezanezhad2026MetaAnalysis}. The contrast with methods evaluated on the full imbalanced distribution is stark. The CNN-LSTM of Salehinejad et al., which extracts features with a 3D ResNet50 and aggregates them with a bidirectional LSTM, attains a patient-level F1 of only 52.92 under cross-validation \citep{SalehinejadISBI}. Such sequential and 2.5D models recover the inter-slice continuity that single-slice classifiers discard, including the CNN-LSTM that Tomita et al. applied to osteoporotic thoracolumbar fractures on chest, abdomen and pelvis CT (89.2\% accuracy and 90.8\% F1 on a 129-scan test set) \citep{Tomita2018} and the two-stage 2.5D network of Hung et al., which processes stacks of adjacent slices for joint vertebra typing and fracture prediction \citep{HungCervical}. Only recently has class imbalance itself been treated as a modeling target, through cost-sensitive training on the full distribution \citep{Pandey2025BayesianEnsemble} or, for fracture-type classification from C3 to L5, GAN-based augmentation of under-represented types \citep{Sindhura2025ViTEnsemble}. Despite these advances, the evaluation gaps summarized in Supplementary Table~S1 remain only partly addressed, since few studies evaluate large, imbalanced cohorts at both the vertebra and patient levels, and fewer still pair such evaluation with explainability or interobserver analyses.

\begin{figure*}[t]
  \centering
  \includegraphics[width=0.9\textwidth, height=0.9\textheight, keepaspectratio]{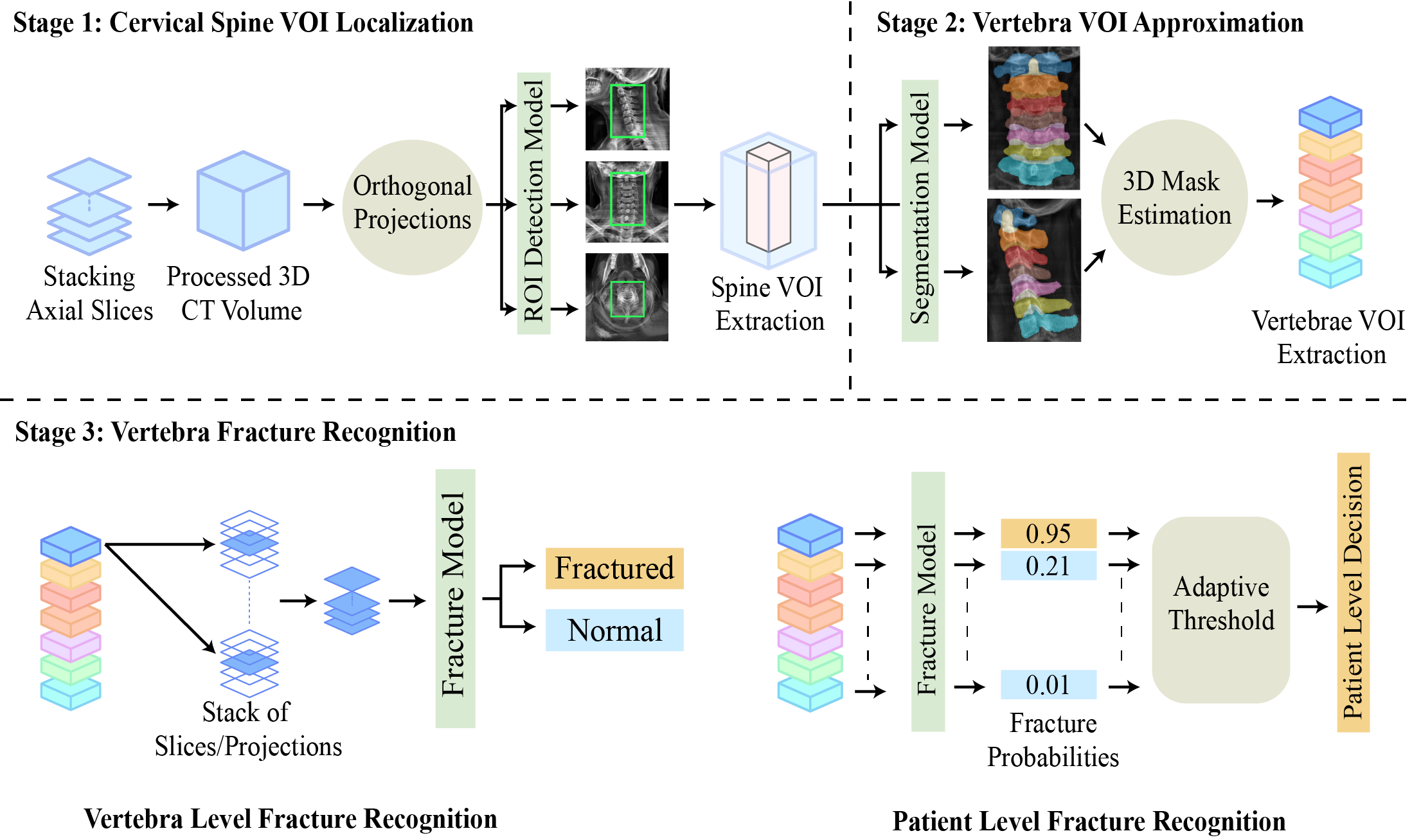}
  \caption{Methodology Overview}
  \label{fig:methodology}
\end{figure*}

\subsection{Vertebra localization and segmentation}

Recognizing a cervical fracture requires more than assigning a binary label to a CT study, since the diagnostic decision must also be linked to the relevant anatomical level. Prior work has therefore approached localization at different granularities. Some classification models infer vertebra involvement indirectly by predicting labels from slices, slice groups, or vertebra-specific crops, while object-detection approaches explicitly localize larger anatomical regions before classification. Gaikwad et al., for example, paired YOLOv5 \citep{YOLOv5} spine localization with a DenseNet201 classifier \citep{HuangDenseNet} and reported a 94\% AUC at 86\% sensitivity \citep{Gaikwad2024}, and similar two-stage designs place a CNN or transformer classifier \citep{Dosovitskiy2020} after a YOLO-based front-end \citep{YOLOv8}, sometimes with Grad-CAM explanation \citep{SelvarajuGradCAM}, as in the pipelines of Chlad et al. and Yaseen et al. \citep{Chlad2023, Yaseen2024}. Other studies localize the suspected fracture itself using bounding boxes or detector outputs \citep{Nejad2023, Sutradhar2025, Boonrod2022}. Our work is positioned at the vertebra level instead, identifying which cervical vertebra is fractured from vertebra-specific volumes extracted with projection-derived masks.

Segmentation provides a different type of localization, because it delineates the vertebral anatomy rather than only enclosing it with a bounding box, and in fracture-recognition cascades this delineation defines the volume of interest that the classifier later examines. The first-place RSNA 2022 solution, for example, segments vertebrae voxel-wise with a 3D ResNet18d-U-Net before a CNN-LSTM classifies each vertebra \citep{HaKaggle2022}, and Kim et al. similarly depend on a 3D EfficientNet-B5 U-Net front-end before downstream fracture detection \citep{Kim2023PlatCon}. Segmentation-then-classification pipelines have also been developed specifically for fracture reporting, including VOI-based fracture detection after coarse-to-fine vertebra segmentation, classification of fresh and old thoracolumbar fractures after 3D segmentation of the fractured vertebral bodies, and recent nnU-Net-plus-classifier frameworks \citep{Golla2023, Chen2024, Glessgen2025VertebraeFx}. These cascades show why segmentation is central to vertebra-level fracture analysis, since it determines the anatomical evidence made available to the classifier.

Dedicated 3D vertebra segmentation studies, supported by public datasets and benchmarks such as the CSI 2014 data, VerSe, and CTSpine1K \citep{ZenodoDataset2016, Sekuboyina2021, Deng2021}, differ mainly in how they separate and label neighboring vertebrae. Some resolve vertebra identity in stages, from the iterative fully convolutional network with memory mechanisms of Lessmann et al., which performs instance-wise segmentation and sequential labeling with an average Dice score of 94.9\% \citep{Lessmann2019}, to multi-stage pipelines combining localization, labeling, and 3D U-Net segmentation \citep{Kawathekar2024} and two-stage Dense U-Net designs for centroid detection and instance segmentation \citep{Cheng2021}. Others couple labeling and segmentation through semantics-instance interaction learning \citep{Mao2025}, and transformer-based designs have been applied to both tasks, from a 3D detector followed by a multi-task encoder-decoder in arbitrary-field-of-view CT \citep{Tao2022} to the recognition of anatomical variants after segmentation \citep{Yang2025WNetViT}. A third group refines the segmentation network itself, with atrous residual paths, 3D attention, and layer normalization \citep{Li2022, Ba2016}, sparse autoencoders \citep{Qadri2022}, or adversarial and recurrent frameworks for spinal structures \citep{Han2018}. These methods achieve strong overlap metrics, but their goal is usually anatomical segmentation itself rather than testing whether a lower-dimensional segmentation surrogate is sufficient for downstream fracture recognition.

Across these localization strategies, the input ultimately presented to the fracture classifier has usually taken one of a few forms, namely raw axial slices, bounding-box crops, vertebra volumes cropped from 3D segmentation, or slice stacks curated from such volumes. Projection-based representations have been used far less extensively for downstream fracture recognition. Mahajan et al. examined non-axial views for direct fracture classification and found that sagittal and coronal slices underperformed axial input \citep{MahajanINCOFT}. Wu et al. provide an important projection-based precedent, localizing lumbar vertebrae from hybrid sagittal and coronal projection images with a 2D projection-location network and a 3D localization criterion, but their final vertebral delineation still relied on a Unet3D-inspired segmentation network and the study did not classify fractures \citep{Wu2022}. Thus, prior projection-based work supports the feasibility of using projections for vertebra localization, while it does not establish whether projection-derived vertebra masks can define VOIs that remain diagnostically viable for downstream cervical fracture recognition.

\subsection{Projection-derived VOIs for fracture recognition}
The preceding literature leaves a specific question unanswered. Projection-based methods have shown value for vertebra localization, but it remains unclear whether vertebra-level VOIs derived from projection-domain mask approximation preserve enough anatomical and fracture-relevant information for downstream classification. This is the central viability question addressed in our study. We approximate each 3D vertebral mask by fusing multi-label segmentations from two orthogonal projections and test whether the resulting vertebra-specific VOIs support cervical fracture recognition. We evaluate this question on the full, imbalanced public RSNA cohort using 5-fold cross-validation at both the vertebra and patient levels, together with Grad-CAM-based explainability, interobserver variability analysis with three expert radiologists, and operating-point analysis of the sensitivity-specificity trade-off for possible screening-oriented use.

The extracted VOIs are classified by a 2.5D spatio-sequential CNN-Transformer ensemble that combines complementary information from conventional slice stacks and maximum-projection slice stacks. Our goal is therefore to measure how much downstream fracture-recognition performance, if any, is lost when vertebra masks are approximated in a lower-dimensional projection domain instead of being obtained by full 3D segmentation, with both mask sources evaluated under the same classifier and imbalanced cross-validation protocol.

\section{Methodology}\label{methodology}

In our study, cervical spine fracture detection was formulated as a projection-driven, three-stage pipeline. First, the 3D CT volume was reconstructed from the DICOM series and converted into orthogonal axial, sagittal, and coronal projections. A YOLOv8 detector was then applied to each view to localize the cervical spine region, and the resulting multi-view detections were fused to define a 3D spinal column volume of interest. Next, sagittal and coronal projections were generated from this volume of interest, and a Unet-based multi-label segmentation model predicted vertebra-specific masks (C1--C7) on the sagittal and coronal projections. These orthogonal masks were combined to approximate 3D vertebral masks, which were subsequently used to extract individual vertebra volumes from the original CT. Finally, the extracted vertebra volumes were evaluated for fracture using an ensemble of two 2.5D CNN-Transformer models. An overview of the complete pipeline is provided in Figure~\ref{fig:methodology}.

\subsection{Dataset Description}\label{dataset-description}

We used the public training set of the RSNA Cervical Spine Fracture Challenge 2022, which contains CT scans of 2,019 patients. The inclusion criteria for the dataset required axial non-contrast CT with 1mm slice thickness, and each volume has on average 350 slices. Bounding-box fracture annotations are provided for 235 patients and voxel-level 3D segmentation masks for 87. The vertebra-level fracture labels were primarily derived from the formal radiology reports submitted by the contributing institutions. Each study carried binary labels denoting fracture presence or absence at every cervical level, and every positive examination was reviewed by a neuroradiologist of the organizing committee against the submitted labels, with discrepancies corrected through adjudication \citep{Lin2023RSNA}. The segmentation masks are NIfTI files that label C1--C7 (values 1--7) and the thoracic vertebrae (values 8--19), with background labeled 0. They denote anatomical vertebral structures rather than fracture regions, and, as detailed in the dataset's supplementary appendix, were generated semi-automatically by a 3D U-Net trained on cervical labels from the VerSe dataset \citep{Sekuboyina2021} and subsequently reviewed and, where necessary, corrected by a radiologist using ITK-SNAP \citep{Lin2023RSNA}. The data were collected at 12 institutions on six continents. Of the 14,133 cervical vertebrae, 1,444 (10.2\%) are fractured, and 961 of the 2,019 patients (47.6\%) have at least one fractured vertebra, with C2, C6 and C7 the most frequently fractured levels. The dataset treats acute and chronic fractures equivalently and excludes post-surgical cases because of streak artifacts and altered anatomy \citep{Lin2023RSNA}. Fracture distribution statistics and sample CT slices are given in Supplementary Figure S1.

\begin{figure}[htbp]
  \centering
  \includegraphics[width=\linewidth]{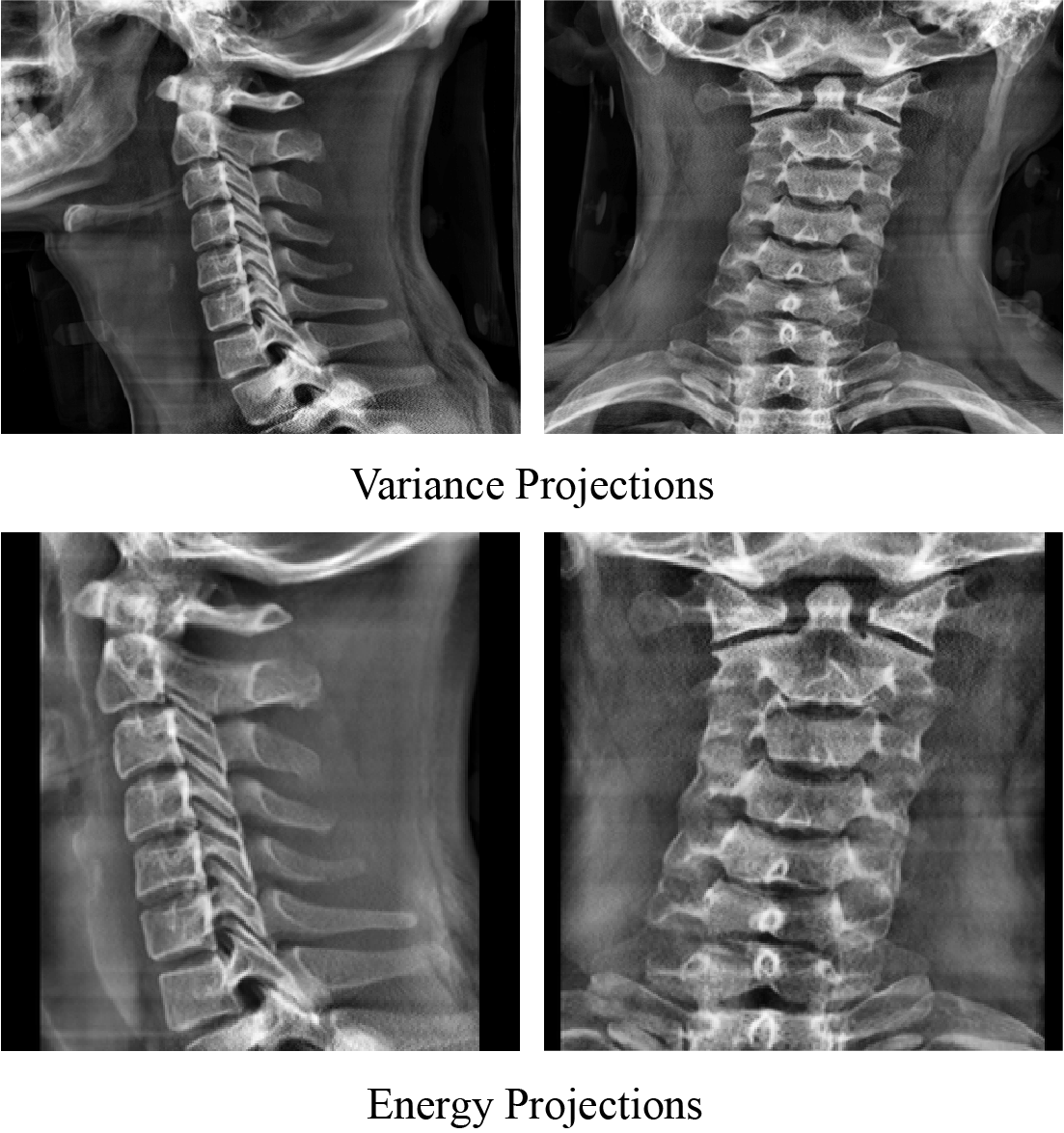}
  \caption{Selected projections for ROI detection and vertebra segmentation.}
  \label{fig:projections}
\end{figure}

\begin{figure*}[t]
  \centering
  \includegraphics[width=\textwidth]{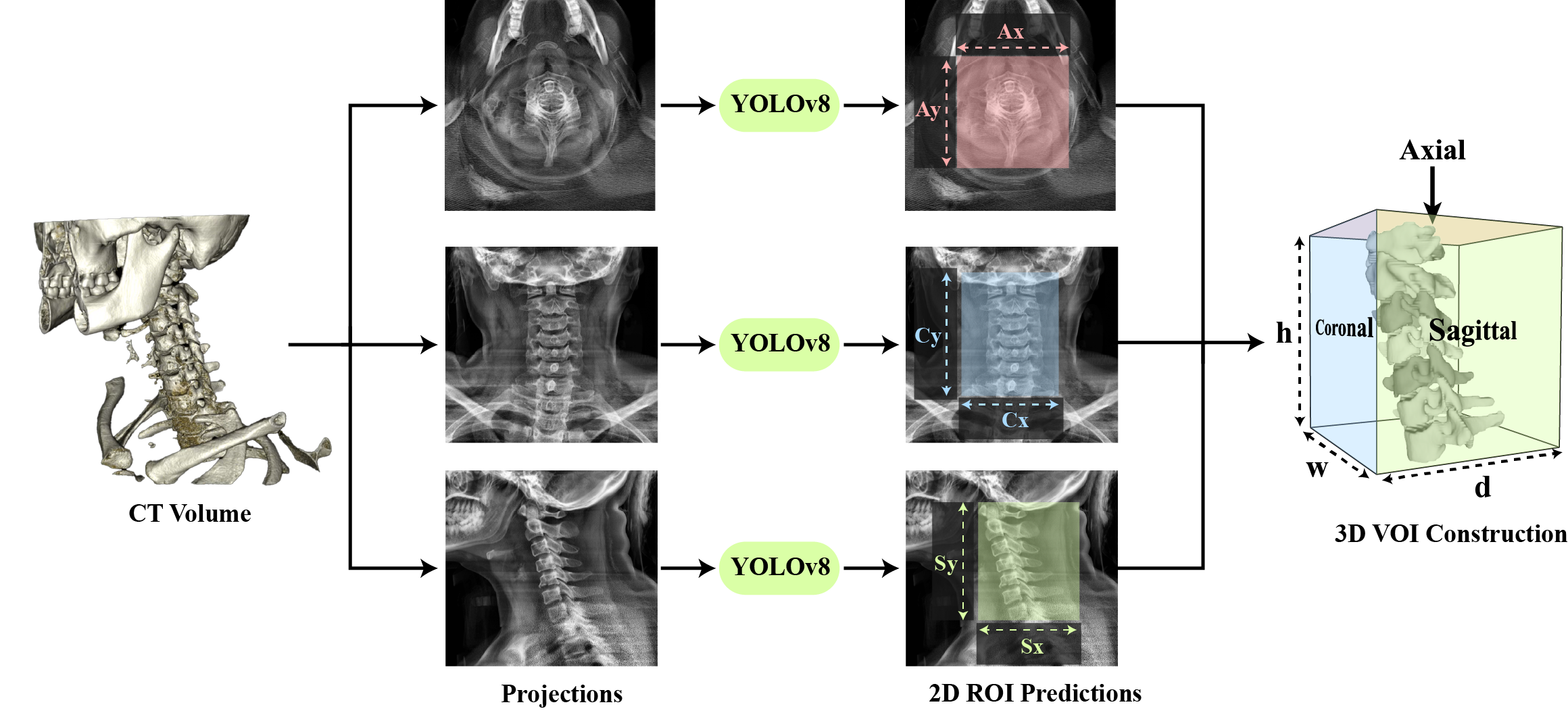}
  \caption{3D cervical spine VOI construction from 2D ROI predictions on the projections.}
  \label{fig:voi_construction}
\end{figure*}

\subsection{Theoretical Formulation: 2D Projections of 3D CT Volumes}\label{sec:projection-formulation}

Each CT volume was mapped into 2D by orthogonal projection, which collapses the voxel intensities along a viewing axis (axial, sagittal or coronal) onto a plane. The aggregating function applied along that axis determines which anatomical features are emphasized. Maximum Intensity Projection (MIP), which retains the brightest voxel along the viewing direction, is a widely used operator of this kind for CT analysis \citep{DiMuzio2011}. We compared four projection operators for cervical spine localization and 23 for vertebra segmentation, and adopted the variance projection for localization and the energy projection for segmentation. All operators are defined and compared visually in Supplementary Section S3.

Let $I(x,y,z)$ denote the voxel intensity at spatial location $(x,y)$ in slice $z$, and let $Z$ be the total number of slices along the projection axis. The variance projection is defined as:
\begin{equation}
    P_{\text{var}}(x,y) = \frac{1}{Z}\sum_{z = 1}^{Z}\left( I(x,y,z) - \mu(x,y) \right)^{2}
    \label{eq:variance_projection}
\end{equation}
\noindent where $\mu(x,y) = \frac{1}{Z}\sum_{z = 1}^{Z}{I(x,y,z)}$ represents the mean intensity along the projection axis. This projection emphasizes regions with high intensity variability across slices, giving strong contrast between vertebral structures and the surrounding soft tissue for localizing the cervical spine region of interest (ROI).

For vertebra segmentation, we employ the energy projection, whose sum-of-squares form follows the angular second moment texture feature \citep{Haralick2007}, defined as:
\begin{equation}
    P_{\text{eng}}(x,y) = \sum_{z = 1}^{Z}{I(x,y,z)^{2}}
    \label{eq:energy_projection}
\end{equation}

The energy projection amplifies high-intensity structures such as cortical bone and so strengthens vertebral boundaries. This benefited multi-label vertebra segmentation under anatomical overlap. Figure~\ref{fig:projections} shows both projections for the same volume. For both, each slice was first enhanced with contrast-limited adaptive histogram equalization (CLAHE) \citep{Zuiderveld1994}, and the resulting projection was enhanced with CLAHE again and rescaled.

\subsection{CT Volume Preprocessing}\label{sec:ct-preprocessing}

Raw CT intensities were converted to Hounsfield units (HU) with the rescale slope and intercept stored in each DICOM header. All three stages used a bone window with a level of 400 HU and a width of 1800 HU, clipping values to $[-500, 1300]$ HU and rescaling them linearly. The wide lower bound keeps lower-density osseous regions within the window while denser cortical bone retains stronger contrast.

The number of axial slices ranges from fewer than 200 to more than 500 across scans, and in sparsely sampled scans the vertebrae are underrepresented in projection space. Each volume was therefore interpolated along the axial axis to at least 400 slices, with cubic spline interpolation \citep{AjaFernandez2024} applied in HU before windowing and nearest-neighbor interpolation for the segmentation masks to preserve their labels.

\subsection{Stage 1: Cervical Spine VOI Detection}\label{sec:stage1-detection}

As each CT scan extends well beyond the cervical spine, the first stage isolates the cervical region, so that the projections used in the later stages show only the cervical vertebrae. Ground-truth regions were derived from the 87 patients with 3D vertebra masks. The C1--C7 labels were combined into a binary cervical mask and max-projected in the axial, sagittal and coronal views, and in each view the ground-truth box was the bounding box of the largest connected component, which ignores small isolated fragments (Supplementary Figure S8). Spine localization was then treated as object detection on these projections, and we compared real-time detectors from two families, the CNN-based You Only Look Once (YOLO) series \citep{RedmonYOLO}, including variants with anchor-free and NMS-free designs \citep{ZhangATSS, Wang2024YOLOv10}, and the transformer-based Real-Time Detection Transformer (RT-DETR) \citep{ZhaoDETR}, which builds on DETR \citep{CarionDETR}.

\subsubsection{Sequential Slice Selection}\label{sec:slice-selection}
Each projection should include only the slices that contain the cervical spine, and the slice ranges were set sequentially. The ground-truth boxes of the 87 patients were aggregated and, after a few outliers were excluded, gave a sagittal range of slices 100--420, which was used for the sagittal projection of every patient (Supplementary Figure S9). The remaining ranges were then derived for each patient. The width of the resulting sagittal box set the coronal slice range, and the axial range was the average of the coronal and sagittal box heights. Supplementary Figure S10 illustrates this procedure for one patient.

\subsubsection{VOI Localization from 2D Predictions}\label{sec:voi-localization}

The variance projection yielded the best results for predicting the 2D cervical spine ROI (Section~\ref{sec:select-detection}). Our goal was to localize the 3D cervical spine VOI within the full CT volume using these 2D projections. To achieve this, 2D ROIs were predicted on the projection of each view using our trained model, and the coordinates of these 2D bounding boxes were then used to calculate the 3D VOI bounding box for the cervical spine region. Figure~\ref{fig:voi_construction} demonstrates this reconstruction process. We added a 20-pixel tolerance ($t$) to each side of the predicted bounding boxes to mitigate potential underprediction. The final 3D VOI dimensions ($h, w, d$) were calculated by averaging the corresponding coordinates from the orthogonal views as follows:

\begin{align}
    h &= \frac{1}{2}\left( c_{y} + s_{y} \right) \pm 2t \label{eq:voi_h} \\
    w &= \frac{1}{2}\left( c_{x} + a_{x} \right) \pm 2t \label{eq:voi_w} \\
    d &= \frac{1}{2}\left( s_{x} + a_{y} \right) \pm 2t \label{eq:voi_d}
\end{align}

\noindent where $c$, $s$, and $a$ represent the coordinate sets from the coronal, sagittal, and axial projections, respectively, and subscripts $x$ and $y$ denote the horizontal and vertical axes within those 2D projections.

\subsection{Stage 2: Cervical Vertebra VOI Segmentation}\label{sec:stage2-segmentation}

The second stage approximates the 3D location of each cervical vertebra from 2D projections of the cervical spine VOI. We used the 87 patients with provided 3D vertebra masks to train a segmentation model that segments the C1--C7 vertebrae from the coronal and sagittal projections. The axial projection of the volume was not used at this stage, as the vertebrae are not separable from a top-down view. Because the goal was to approximate vertebra locations rather than to recover exact 3D surfaces, segmenting from these two orthogonal perspectives was sufficient.

\subsubsection{Multi-label Mask Generation}\label{sec:multilabel-mask-generation}

Because of cervical curvature and neck alignment, distinct vertebral structures can project onto the same location in the coronal and sagittal views, so a single projection pixel can belong to more than one vertebra, as C1 and C2 do at the dens in both views. In the projected ground-truth masks, 37.72\% of coronal pixels carried a single vertebral label, 12.69\% carried two, 0.33\% three and 0.02\% four, and 49.24\% were background. Overlap was less pronounced in the sagittal projection, where 32.17\% of pixels carried one label, 5.20\% carried two, none carried three or more, and 62.63\% were background. Further examples of this overlap are given in Supplementary Section S4.1.

\begin{figure}[htbp]
  \centering
  \includegraphics[width=\linewidth]{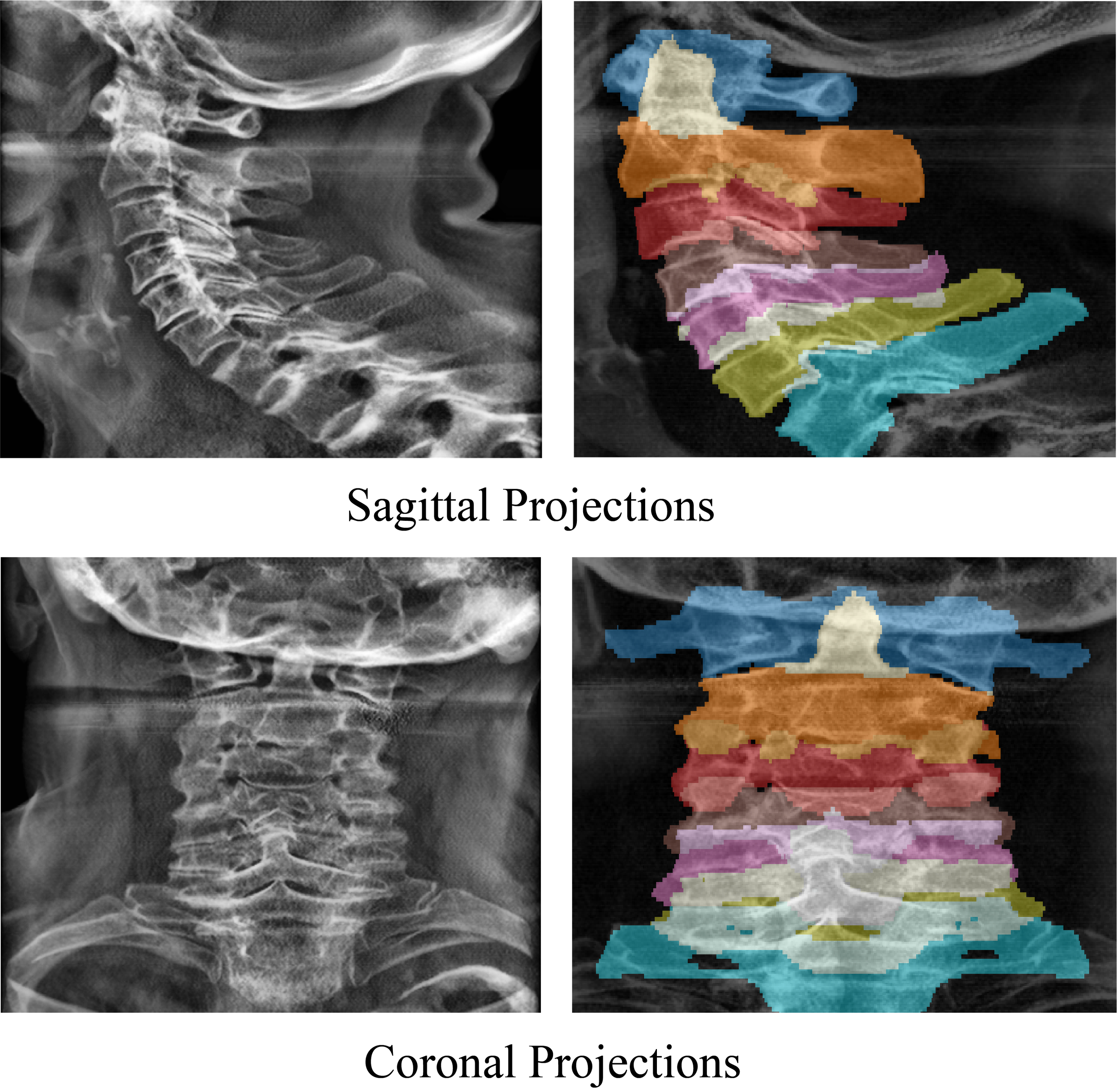}
  \caption{Multi-label vertebra mask shown over energy projections. Overlapping regions are assigned to multiple vertebral channels simultaneously.}
  \label{fig:multilabel_mask}
\end{figure}

Multi-class segmentation, which assigns each pixel to a single class, cannot represent this overlap, so we used a multi-label formulation in which a pixel can belong to several vertebrae. The C1--C7 labels of the 3D ground-truth mask were projected independently along the viewing axis and stacked as separate binary channels (Figure~\ref{fig:multilabel_mask}).

\begin{figure*}[t]
  \centering
  \includegraphics[width=\textwidth]{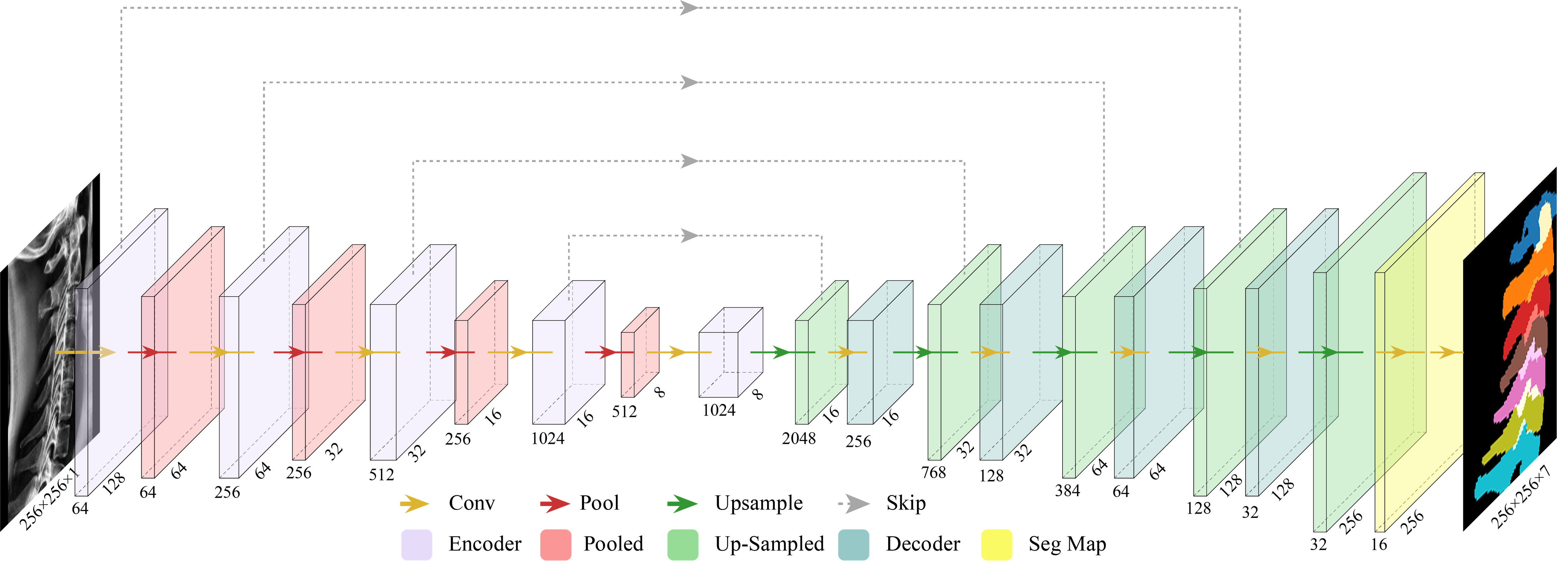}
\caption{Proposed segmentation network architecture. Each feature map is annotated with its spatial and channel dimensions: the value under each block's thickness denotes the number of channels, and the value along its slanted lower edge denotes the square spatial dimension.}
  \label{fig:proposed_seg_network}
\end{figure*}

\subsubsection{Proposed Segmentation Network}\label{sec:seg-network}

We used a modified U-Net \citep{RonnebergerUNet} with a DenseNet121 encoder \citep{HuangDenseNet}, initialized from ImageNet with its classification head removed. Its dense connectivity encourages feature reuse, which helps to discriminate between adjacent and overlapping vertebrae of similar appearance \citep{Cinar2022}. Skip connections forward the feature maps of the initial convolutional block and the dense blocks to the corresponding decoder stages, and average pooling replaces max pooling for a smoother reduction in resolution that better preserves context around thin or low-contrast vertebral boundaries.

The decoder emits a $C$-channel output in which each channel is an independent probability map for one of C1--C7, with a sigmoid applied per channel so that a pixel can be assigned to several vertebrae at once. Figure~\ref{fig:proposed_seg_network} shows the architecture with the tensor dimensions of each stage.

\subsubsection{Segmentation Loss}\label{sec:seg-loss}

The network was trained to minimize the average of the Dice loss, which is sensitive to small regions \citep{Fidon2017, MilletariVNet}, and the Jaccard loss, which penalizes false positives more strictly (Supplementary Section S4.6),
\begin{equation}
    L_{\text{Combined}} = \frac{L_{\text{Dice}} + L_{\text{Jaccard}}}{2}.
    \label{eq:combined_loss}
\end{equation}

\subsubsection{3D Vertebra Mask Estimation and Volume Extraction}\label{sec:3d-mask-estimation}

Segmentation masks were predicted on the coronal and sagittal projections of all 2019 patients and then reverse-resized and unpadded to the original dimensions. The 3D vertebra masks were reconstructed by back-projecting each predicted 2D mask along its viewing axis and fusing the two views on a per-voxel basis. Let a voxel be indexed by $(x, y, z)$, where $x$, $y$, and $z$ denote the left-right, anterior-posterior, and superior-inferior axes, respectively. The coronal projection collapses the anterior-posterior axis to yield a 2D mask $\hat{M}^{c}_{v}(x, z)$ for vertebra $v$, while the sagittal projection collapses the left-right axis to yield $\hat{M}^{s}_{v}(y, z)$. Each predicted 2D mask is back-projected along its collapsed axis, and the reconstructed 3D mask $V_{v}$ is obtained as their per-voxel logical AND:
\begin{equation}
    V_{v}(x, y, z) = \hat{M}^{c}_{v}(x, z) \wedge \hat{M}^{s}_{v}(y, z), \qquad v \in \{1, \dots, 7\}
    \label{eq:mask_fusion}
\end{equation}
\noindent so that a voxel is assigned label $v$ only when both back-projected masks are active at that location. This process is illustrated in Supplementary Figure S12. The resulting 3D approximations were used to crop individual vertebra volumes from each patient volume, yielding 14,129 extracted vertebrae, with 4 excluded due to noise-induced segmentation failure, all of which were from a single patient.

\begin{figure*}[t]
  \centering
  \includegraphics[width=\textwidth]{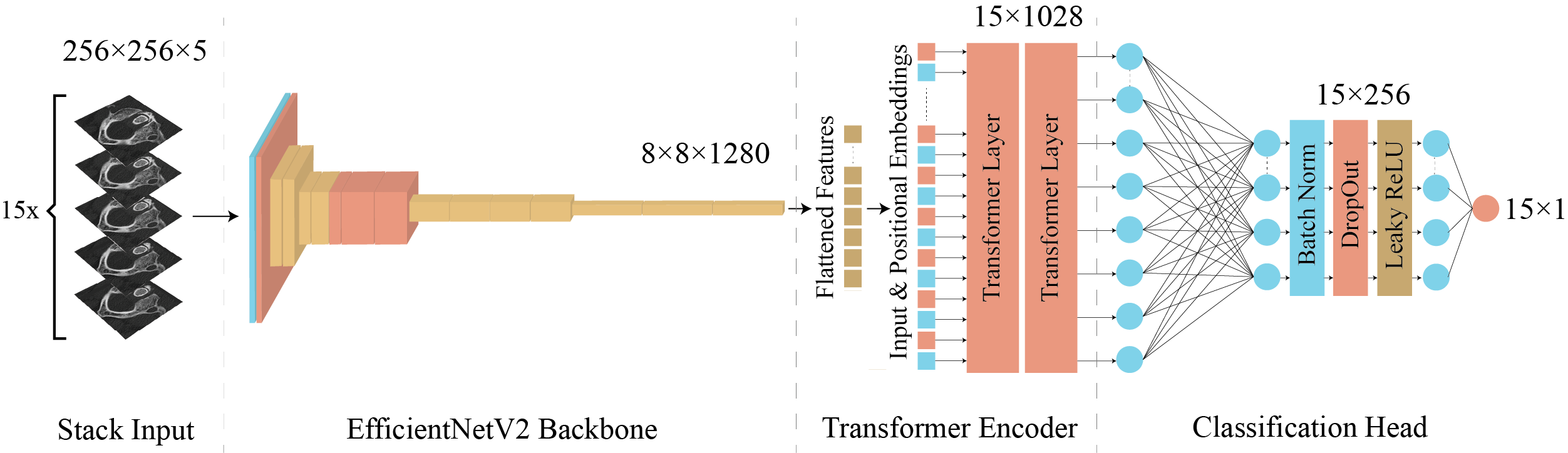}
  \caption{Proposed spatio-sequential fracture identification architecture. The output dimensions $(H, W, C)$ are annotated at the final step of each major stage; a complete layer-by-layer dimension list is provided in the Supplementary Material.}
  \label{fig:cnn_transformer}
\end{figure*}

\subsection{Stage 3: Cervical Vertebra Fracture Identification}\label{sec:stage3-fracture}

The final stage classifies each vertebra volume extracted in Stage 2 as fractured or not fractured, giving predictions at the vertebra and patient levels. We compared three classifier families, 2D projection-based, full 3D volumetric and 2.5D spatio-sequential models, and adopted the 2.5D framework, which gave the best balance between volumetric context and diagnostic performance. The 2D and 3D baselines are described in Supplementary Section S4.4. The 2.5D models read sequences of selected slices from each vertebra volume, in one variant as slice stacks that capture local 3D context and in the other as stacks of maximum intensity projections (MIPs) that widen the effective receptive field. Both predict a fracture probability at each sequence step, and the vertebra-level probability is the mean over steps. At the patient level, we evaluated an If-Any rule, which flags a patient when any vertebra is predicted fractured, and an adaptive threshold based on score-level fusion of the two variants.

\subsubsection{Data Preprocessing and Stacking}\label{sec:cls-preprocessing}

Each extracted vertebra volume was bone-windowed and normalized as in Section~\ref{sec:ct-preprocessing} and converted into a sequence of 15 inputs. For the slice stack, 15 evenly spaced slices were sampled, and each was stacked along the channel axis with its two preceding and two succeeding slices. For the MIP stack, 75 evenly spaced slices were sampled, each was reduced with its two neighbors on either side to a single maximum intensity projection, and the projections were grouped into 15 sequential sets, which broadens the volumetric coverage without increasing the input size. All inputs were resized to $256 \times 256 \times 5$, and Supplementary Figure S13 shows an example.

Following the augmentation strategy of the RSNA challenge-winning solution \citep{HaKaggle2022}, every training vertebra, whatever its label, received geometric, photometric, noise and blur, region-dropout and slice-order augmentation (Supplementary Table S15). MixUp \citep{Zhang2018Mixup} was applied to a batch with probability 0.55, combining each vertebra stack with a randomly paired stack from the same batch using a mixing coefficient drawn from $\mathrm{Beta}(1,1)$, and the loss was combined in the same proportions for the two labels.

\begin{figure*}[t]
  \centering
  \includegraphics[width=\textwidth]{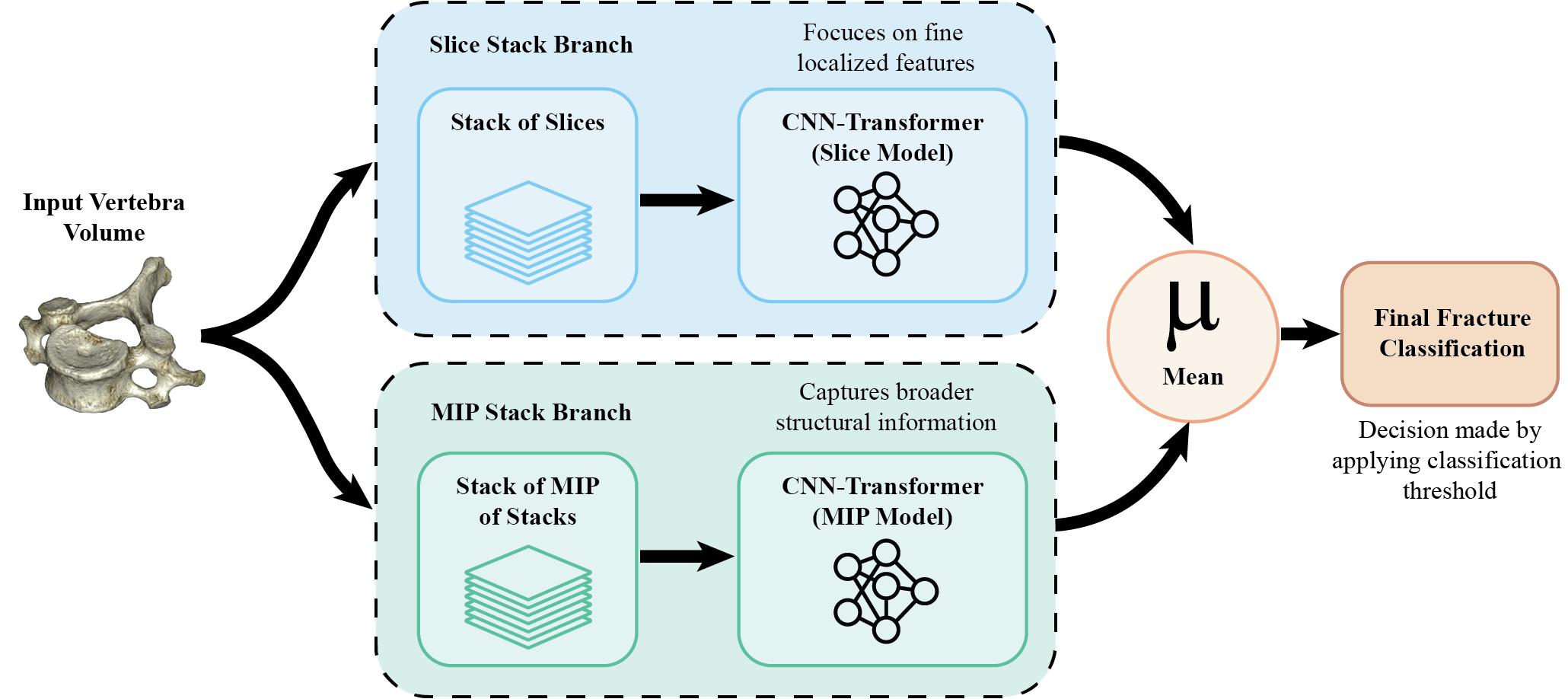}
  \caption{Proposed score-fusion ensemble combining the slice-stack and MIP-stack 2.5D variants.}
  \label{fig:ensemble}
\end{figure*}

\subsubsection{Spatio-Sequential Networks}\label{sec:spatio-seq-networks}

Fractures can span several slices so we modeled each vertebra as a sequence rather than classifying slices independently, pairing a CNN backbone for per-step spatial feature extraction with a sequential aggregator that captures inter-slice dependencies. We compared two aggregators, an LSTM \citep{Hochreiter1997} and a Transformer encoder \citep{Vaswani2017}, on top of the same backbone, and the CNN-Transformer gave the higher F1-score and ROC-AUC (Section~\ref{sec:select-classification}).

We therefore adopt the CNN-Transformer, illustrated in Figure~\ref{fig:cnn_transformer}, as our proposed network, using EfficientNetV2 \citep{TanEfficientNetV2} to extract a feature vector from each element of the input sequence. These per-step features, augmented with sinusoidal positional encoding to preserve slice order, are passed to a Transformer encoder that models dependencies across the sequence through multi-head self-attention \citep{Vaswani2017}. Two encoder layers were used, as this depth gave the highest F1-score and ROC-AUC among the depths compared (Section~\ref{sec:select-classification}). The encoder produces a fracture probability for each of the 15 sequence steps, and their mean is taken as the vertebra-level fracture probability.

\paragraph{Loss Function}

The network was trained with a weighted binary cross-entropy loss \citep{Goodfellow2016} that raises the penalty for misclassifying fractured cases, counteracting the strong majority of non-fractured vertebrae that would otherwise bias the model toward low sensitivity. Denoting by $y_i$ and $\widehat{y_i}$ the ground-truth label and predicted probability for sample $i$ and by $w_i$ a per-sample weight, the loss over $N$ samples is
\begin{equation}
    L = - \frac{1}{\sum_{i=1}^{N} w_{i}} \sum_{i=1}^{N} w_{i} \left[ y_{i} \log\left( \widehat{y_{i}} \right) + \left( 1 - y_{i} \right) \log\left( 1 - \widehat{y_{i}} \right) \right]
    \label{eq:weighted_bce_loss}
\end{equation}
\noindent where $w_i = 2$ for fracture-positive samples ($y_i = 1$) and $w_i = 1$ for fracture-negative samples ($y_i = 0$). Normalizing by $\sum_i w_i$ keeps the loss scale stable. Comparison with other imbalance mitigation techniques and the choice of weight values are described in Section~\ref{sec:imbalance-protocol}.

\paragraph{Score-Fusion Ensemble}

To strengthen vertebra-level performance, we ensembled the two CNN-Transformer variants, one trained on slice stacks and the other on MIP stacks. The two share an architecture but emphasize different information, the slice-stack model a detailed local view and the MIP-stack model broader structural context, so combining them draws on complementary evidence from the two representations. The fused fracture probability of each vertebra is the mean of the two variants' outputs, as illustrated in Figure~\ref{fig:ensemble}.

\paragraph{Patient-Level Adaptive Threshold}

For a definitive patient-level diagnosis, vertebra-level predictions were aggregated using a threshold that adapts to the agreement between the two 2.5D variants. For each cervical vertebra $v \in \{1, \dots, 7\}$, let $\mu^{(1)}_v$ and $\mu^{(2)}_v$ denote the mean fracture probabilities predicted by the slice-stack and MIP-stack variants across that vertebra's stacks. Each vertebra is assigned a unified score and an agreement score,
\begin{equation}
    s_v = \frac{1}{2}\left( \mu^{(1)}_v + \mu^{(2)}_v \right), \qquad a_v = 1 - \left| \mu^{(1)}_v - \mu^{(2)}_v \right|
    \label{eq:vertebra_scores}
\end{equation}
\noindent where $a_v = 1$ denotes perfect agreement between the two variants. The patient-level fracture signal is taken as the maximum unified score across the seven cervical vertebrae, and the overall inter-variant agreement as the mean of the per-vertebra agreement scores,
\begin{equation}
    S = \max_{v \in \{1, \dots, 7\}} s_v, \qquad \bar{A} = \frac{1}{7}\sum_{v=1}^{7} a_v.
    \label{eq:patient_signal}
\end{equation}
The decision threshold is then set adaptively from $\bar{A}$, so that stronger agreement permits a more sensitive (lower) threshold while weaker agreement enforces a more conservative (higher) one,
\begin{equation}
    \tau(\bar{A}) =
    \begin{cases}
        0.3, & \bar{A} > 0.8, \\
        0.5, & 0.6 < \bar{A} \le 0.8, \\
        0.7, & \bar{A} \le 0.6.
    \end{cases}
    \label{eq:adaptive_threshold}
\end{equation}
The agreement bands and their corresponding thresholds were set as a fixed rule of thumb. A patient is classified as fractured when $S > \tau(\bar{A})$.

\subsubsection{Explainability for Fracture Identification}
To examine the evidence behind the CNN-Transformer's predictions, we visualized it at two levels \citep{Holzinger2017, Miller2019}. Grad-CAM \citep{SelvarajuGradCAM}, applied to the final convolutional layers of the EfficientNetV2 backbone, gives heat maps of the regions within each of the 15 slice stacks that contribute most to a fracture prediction, which can be checked against the expected fracture location \citep{Simonyan2013}. The self-attention weights of the transformer encoder \citep{Vaswani2017}, averaged for each slice stack, show which of the 15 stacks the model weights most heavily, and whether these include the stacks that contain the fracture \citep{Vig2019, Wiegreffe2019}.

\subsubsection{Interobserver Variability Test}\label{sec:interobserver-protocol}

To assess the consistency of fracture labeling, three expert radiologists independently annotated a subset of 30 randomly sampled patients. The scans were anonymized and given to each reader under different names, and no reader had access to the original radiology reports. The annotations were compared with one another, with the reference labels of the dataset and with the out-of-fold predictions of the proposed model for the same patients, using pairwise Cohen's $\kappa$ \citep{Cohen1960} and, for agreement among more than two raters, Fleiss' $\kappa$ \citep{Fleiss1971}. Both rescale the observed agreement so that a value of 0 corresponds to the agreement expected by chance, given how often each label is used, and 1 to perfect agreement. Their definitions are given in Supplementary Section S11.

\subsection{Evaluation Metrics}\label{sec:metrics}

Each stage was evaluated with metrics suited to its output, and the formulas are given in Supplementary Section S4.2.

\subsubsection{Detection} Cervical spine ROI detection was evaluated using the mean Intersection over Union (mIoU), which quantifies the overlap between predicted and ground-truth regions and jointly reflects true positives, false positives, and false negatives \citep{RezatofighiGIOU}.

\subsubsection{Segmentation} Vertebra segmentation was evaluated using two spatial-overlap metrics, the Intersection over Union (IoU) and the Dice Similarity Coefficient (DSC), together with the 95th percentile Hausdorff Distance (HD95), which measures boundary agreement while excluding outlying points \citep{Taha2015}. To rank candidate models across folds using all three metrics jointly, we computed a normalized composite score,
\begin{equation}
\text{Score} = \frac{\text{IoU}}{\mu_{\text{IoU}}}
+ \frac{\text{DSC}}{\mu_{\text{DSC}}}
- \frac{\text{HD95}}{\mu_{\text{HD95}}}
\label{eq:score_metric}
\end{equation}
\noindent where $\mu_{\text{IoU}}$, $\mu_{\text{DSC}}$, and $\mu_{\text{HD95}}$ denote the mean values of IoU, DSC, and HD95 across the compared models, respectively. HD95 enters as a lower-is-better penalty.

\subsubsection{Fracture Identification} Fracture classification was evaluated using Accuracy, Precision, Sensitivity, Specificity, and the F1-score, computed at both the vertebra and patient levels. To assess threshold-independent performance, we additionally reported the area under the receiver operating characteristic curve (ROC-AUC) and the area under the precision-recall curve (AUPRC). The latter is particularly informative under the strong class imbalance of this task, as it focuses on performance for the minority fractured class rather than being dominated by the abundant non-fractured cases. A classifier that ranks cases at random has an expected AUPRC equal to the proportion of positive cases, so its chance level here is 10.2\% per vertebra and 47.6\% per patient, rather than the 50\% of ROC-AUC \citep{SaitoRehmsmeier2015}.

\subsubsection{Statistical Analysis}\label{sec:statistics}

All comparisons between pipelines are paired. The pipelines share the five cross-validation folds of Section~\ref{sec:experimental_setup} and are scored on the same 14,133 vertebrae and 2,019 patients (Section~\ref{sec:eval-rule}). AUPRC is the primary metric and ROC-AUC the secondary metric, each at the vertebra and patient levels, and differences are reported as $\Delta$, the proposed pipeline minus the comparator, in percentage points.

The primary analysis is a paired $t$-test on the five fold-wise differences, with four degrees of freedom \citep{Dietterich1998}. Each comparison is reported as $\Delta$ with its two-sided 95\% confidence interval (CI) and $p$-value. No equivalence margin was specified, so we report intervals rather than tests of equivalence, and each interval can be judged against whatever margin a reader considers acceptable. We refer to the lower limit of the 95\% CI as the loss bound, because losses larger than its absolute value lie outside the interval and are not supported by the data.

Because the fold-wise test rests on five differences, we also applied a patient-clustered paired bootstrap to the out-of-fold predictions \citep{EfronTibshirani1993}. Within each fold, patients were resampled with replacement, each keeping all of its vertebrae, and both pipelines were scored on the same resample. The metric was computed per fold, averaged over the five folds and then differenced, and 2,000 replicates (seed 42) gave 95\% percentile intervals around the $\Delta$ observed on the original data. The bootstrap holds the trained models fixed and captures variation across patients, whereas the fold-wise test also captures variation from retraining. The same procedure gives the per-level comparisons in Supplementary Section S8. The segmentation subgroup comparisons in Section~\ref{sec:seg-accuracy} resample the 87 patients in the same way, with differences taken between subgroups at the same vertebral level. All intervals in the text, tables and figures are two-sided 95\% CIs.

\subsection{Experimental Setup and Implementation Details}\label{sec:experimental_setup}

All models were trained and evaluated on a single workstation equipped with an NVIDIA GeForce RTX 3090 GPU (24GB VRAM), an Intel Core i9-12900K CPU (3.20 GHz), and 32GB of RAM, running Python 3.11.9 and PyTorch 2.2. The detection and segmentation stages were trained on the 87 patients for whom 3D vertebra masks were available, while the fracture identification stage was trained on all 2019 patients. The pipeline was assembled sequentially, with the best-performing model from each stage used to generate the inputs for the next.

Each stage was evaluated using 5-fold cross-validation. The detection and segmentation stages shared identical fold assignments over the 87-patient subset, while the fracture identification stage used its own folds over the full dataset. The masks from which the classifier's VOIs were extracted came from segmentation models that had not seen the patient, because each of the 87 annotated patients was segmented by the model of the fold in which it was held out, and the remaining 1,932 patients were never used to train the detection or segmentation stages. In every fold, 20\% of the training split was held out for validation, which alone drove hyperparameter tuning and early stopping, and the test set rotated across the five folds so that every patient was tested exactly once. Unless stated otherwise, the results in Section~\ref{results} are the means over the five test sets, which were held out from training, early stopping and hyperparameter tuning within each fold.

The detectors were implemented with the Ultralytics library (v8.2.5), initialized with COCO-pretrained weights and trained with mosaic, HSV-jitter, random-scaling and translation augmentation. The segmentation models were implemented with the Segmentation Models PyTorch library \citep{SegmentationModelsPyTorch} with ImageNet-pretrained encoders \citep{Deng2009}, trained with random rotation and horizontal and vertical flipping, and the learning rate was reduced by a fixed factor when the validation loss plateaued. They were trained on projections of the ground-truth ROI-cropped spine columns and applied at inference to the detector-predicted ROIs of all patients. The classifiers were implemented with TIMM, and its 3D extension TIMM-3D for the 3D baseline, and were trained and evaluated on all 2,019 patients, comprising the 14,129 extracted vertebrae (the remaining four are scored as described in Section~\ref{sec:eval-rule}). Table~\ref{tab:training_params} lists the training parameters of all three stages.

\begin{table}[htbp]
  \centering
  \caption{Training parameters of the detection, segmentation and fracture classification stages. All stages used five-fold cross-validation.}
  \label{tab:training_params}
  \resizebox{\linewidth}{!}{%
    \begin{tabular}{lccccc}
      \toprule
      \multirow{2}{*}{Training parameter} & \multirow{2}{*}{Detection} & \multirow{2}{*}{Segmentation} & \multicolumn{3}{c}{Classification} \\
      \cmidrule(lr){4-6}
       & & & 2D & 2.5D & 3D \\
      \midrule
      Input size & 640 & 256 & $224\times224$ & $15\times5\times256\times256$ & $64\times224\times224$ \\
      Batch size & 8 & 8 & 8 & 8 & 4 \\
      Optimizer & SGD & Adam & Adam & AdamW & AdamW \\
      Initial learning rate & 1e-2 & 1e-4 & 1e-4 & 6e-5 & 1e-4 \\
      Weight decay & 5e-4 & 1e-4 & 1e-4 & 1e-4 & 1e-4 \\
      Max epochs & 100 & 120 & 100 & 120 & 100 \\
      Early stopping patience & 20 & 20 & 14 & 20 & 14 \\
      LR drop patience & -- & 5 & 5 & -- & 10 \\
      LR reduction factor & -- & 0.2 & 0.1 & -- & 0.5 \\
      \bottomrule
    \end{tabular}%
  }
  \\[2pt]
  {\footnotesize Detection used a 3-epoch warm-up followed by linear decay of the learning rate to 0.01 times its initial value, and the 2.5D models used cosine annealing from $6\times10^{-5}$ to $2.3\times10^{-5}$, so neither reduced the learning rate on a plateau.\par}
\end{table}

\subsubsection{Computational Cost}

To clarify the computational requirements of the pipeline, we report per-stage figures measured at inference on a per-patient basis in Table~\ref{tab:compute}, namely peak GPU memory, and the floating-point operations (GFLOPs) and parameter count of each neural stage. A complete per-step runtime breakdown, including the CPU-bound DICOM loading, projection, and volume-extraction steps, is provided in Supplementary Table~S17. GFLOPs and parameter counts were computed with fvcore \citep{fvcore}.

The segmentation stage is where the projection-driven formulation most directly reduces the dimensionality of the task. By segmenting vertebrae from 2D coronal and sagittal projections rather than from the full 3D volume, this stage requires only 33.6 GFLOPs in total across both views, compared with 836.8 GFLOPs for the 3D U-Net segmentation stage of the RSNA challenge-winning pipeline operating on a $128^3$ volume \citep{HaKaggle2022}, an approximately twenty-five-fold reduction, and with fewer parameters as well (27.2M versus 39.7M).

\begin{table}[htbp]
  \centering
  \caption{Per-stage computational cost at inference, reported per patient.}
  \label{tab:compute}
  \resizebox{\linewidth}{!}{%
    \begin{tabular}{lccc}
      \hline
      Stage & Peak GPU (GB) & GFLOPs & Params (M) \\
      \hline
      Detection (3 views) & 0.676 & 773.40 & 204.45 \\
      Segmentation (2 views) & 0.376 & 33.63 & 27.20 \\
      Classification (per model) & 0.426 & 604.53  & 44.13 \\
      Classification (score fusion) & 0.426 & 1209.07 & 88.25 \\
      \hline
    \end{tabular}%
  }
\end{table}

\subsection{Comparison and Robustness Protocols}\label{sec:protocols}

This section describes the experimental protocols added to test the projection-derived masks against 3D segmentation, to handle missing predictions consistently, to examine class imbalance, and to trace the propagation of upstream errors. All protocols use the five-fold design of Section~\ref{sec:experimental_setup}, and all comparisons use the statistical analysis of Section~\ref{sec:statistics}.

\subsubsection{Mask-Source Comparison}\label{sec:mask-source-arm}

To compare projection-derived masks with full 3D segmentation under a fixed classifier, we built a second version of the pipeline in which the vertebra VOIs come from a 3D segmentation network. Its segmentation stage is the 3D ResNet18d-Unet of the RSNA 2022 winning solution \citep{HaKaggle2022}, trained from scratch with the publicly released code following the published description and run on all 2,019 patients, and its vertebra VOIs are the volumes cropped by that pipeline. We refer to its masks as the 3D-segmentation masks. These VOIs were passed to the same 2.5D classifiers as in the proposed pipeline (slice-stack, MIP-stack and score fusion), trained with the same settings on the same five folds, so the two pipelines differ only in how the vertebra VOIs are obtained.

As an independently engineered whole-pipeline comparator, we also reproduced the RSNA 2022 winning solution, which classifies each vertebra with its own CNN-LSTM \citep{HaKaggle2022}. Its classifier was trained and evaluated on the same five folds, using the per-vertebra input stacks released by the author, which also contain the segmentation mask of the middle slice, an input that the stacks in our pipeline do not use. These stacks were produced by the author's segmentation models, which were trained on the 87 voxel-annotated patients. Its final competition entry combined nearly 20 such models, a configuration assembled for a challenge ranked by weighted log loss on a private test set and too large for practical deployment. We therefore used its publicly released single-model implementation, which can be trained and evaluated under the same five-fold protocol as our pipeline and keeps the comparison on the pipeline design, 3D segmentation followed by vertebra classification.

\subsubsection{Evaluation Rule for Missing Predictions}\label{sec:eval-rule}

Every pipeline was scored on all 14,133 cervical vertebrae (1,444 fractured) and all 2,019 patients (961 fractured). A vertebra for which a pipeline produced no prediction was scored as non-fractured with a fracture probability of zero, which affected at most 27 vertebrae in any pipeline. As a sensitivity analysis, we repeated the paired comparisons after removing the union of these vertebrae (27 in total) from every pipeline in the vertebra-level analyses, and the five patients who own them in the patient-level analyses, and report the results in Supplementary Section S8.3. None of the five patients belongs to the 87-patient subset, so the analyses on that subset are unaffected.

\subsubsection{Class-Imbalance Experiments}\label{sec:imbalance-protocol}

To examine the training objective (Section~\ref{sec:imbalance}), the slice-stack model was trained with seven strategies, with all other settings held fixed. Four strategies use the weighted binary cross-entropy of Equation~\ref{eq:weighted_bce_loss} with a positive-class weight of 1 (unweighted), 2 (proposed), 3 or 8.78, the last approximating the ratio of non-fractured to fractured vertebrae in the cohort (prevalence weighting). The fifth is a focal loss \citep{Lin2017Focal}, in which the per-sample cross-entropy is multiplied by $(1-p_{t,i})^{\gamma}$, where $p_{t,i}$ is the predicted probability of the true class,
\begin{equation}
    L_{\text{focal}} = \frac{\sum_{i=1}^{N} w_{i}\,(1-p_{t,i})^{\gamma}\,\mathrm{BCE}_{i}}{\sum_{i=1}^{N} w_{i}},
    \label{eq:focal_loss}
\end{equation}
with $w_i = \alpha$ for fractured and $w_i = 1$ for non-fractured vertebrae, $\alpha = 2$ and $\gamma = 2$. In this form $\alpha$ acts as a positive-class weight, so the focal loss differs from the proposed loss only by the focusing term. The remaining two strategies resample the training split of each fold to a 1:1 ratio of fractured to non-fractured vertebrae and use unweighted binary cross-entropy, while the validation and test splits are left unchanged. Oversampling draws fractured vertebrae with replacement until they match the non-fractured count, so that each fractured vertebra appears about nine times on average and the training split grows from about 9,040 to about 16,200 vertebrae. Undersampling draws non-fractured vertebrae without replacement down to the fractured count, keeping about 11\% of them, so that the training split shrinks to between 1,750 and 1,870 vertebrae. Both draws use a fixed seed per fold.

\begin{figure*}[t]
  \centering
  \includegraphics[width=\textwidth]{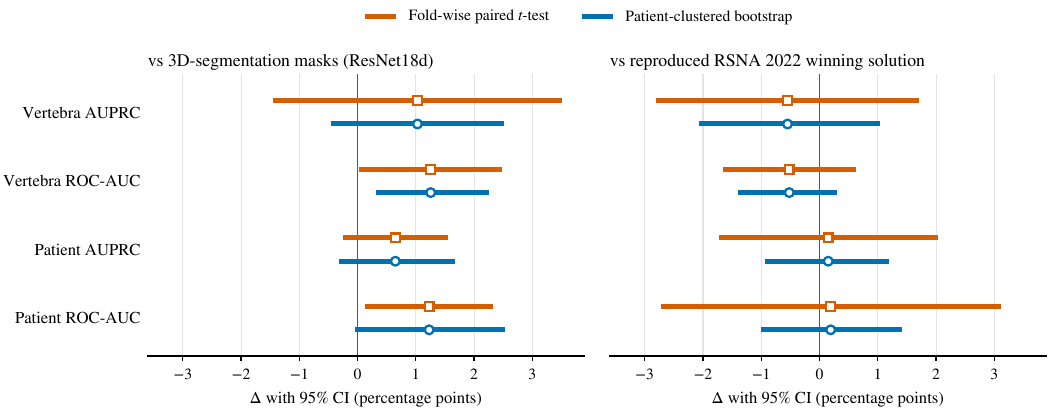}
  \caption{Paired differences for the proposed score-fusion ensemble against the 3D-segmentation masks (left) and the reproduced RSNA 2022 winning solution (right). Points are $\Delta$ (proposed minus comparator, percentage points), and bars are 95\% CIs from the fold-wise paired $t$-test and the patient-clustered bootstrap. The vertical line marks zero, and the lower end of each bar is the loss bound.}
  \label{fig:paired_forest}
\end{figure*}

\paragraph{Balanced-Data Protocol}\label{sec:balanced-protocol}
Section~\ref{sec:balanced} compares three settings. Setting A is the proposed model, trained and tested at the native prevalence of the dataset. Setting B is the undersampling model above, with its checkpoint selected on the imbalanced validation split and tested on the full cohort. Setting C scores the same Setting B models on a class-balanced test set, so B and C differ only in the test set. The balanced test set contains all 1,444 fractured vertebrae and 1,444 non-fractured vertebrae drawn at random within each test partition, 2,888 vertebrae from 1,571 patients in total, and the original patient-grouped folds were kept, so no patient appears in more than one split.

\subsubsection{Error Propagation and Ground-Truth Ceiling}\label{sec:gt-ceiling}

The error-propagation analysis (Section~\ref{sec:error-propagation}) perturbs the ground-truth spine ROI at levels $p = 0.10$, 0.20 and 0.30. Along each axis separately, the center of the box was shifted by a random amount drawn from $\mathcal{U}(-p, p)$ times the length of the box along that axis, and the length was rescaled by a factor drawn from $\mathcal{U}(1-p, 1+p)$. The box was then rebuilt from the new center and length, clipped to the volume and kept at least one voxel thick, and segmentation and classification were run on the perturbed ROI as in the full pipeline.

As a reference with ground-truth masks, we also built a training-matched ceiling, in which the score-fusion model is fine-tuned and tested on VOIs extracted with the ground-truth 3D masks of the 87 annotated patients. Because 87 patients are too few to train the classifier from scratch, both 2.5D variants were first pretrained on VOIs extracted with the 3D-segmentation masks of the remaining 1,932 patients. They were then fine-tuned and evaluated on ground-truth-mask VOIs by five-fold cross-validation over the 87 patients, using the folds of the detection and segmentation stages and the standard 2.5D training settings (Table~\ref{tab:training_params}). Patient-level predictions in the error-propagation analysis, including the ceiling, use the proposed adaptive-average aggregation.

\section{Results}\label{results}

 Section~\ref{sec:mask-source} compares fracture recognition from VOIs extracted with projection-derived masks and with 3D-segmentation masks. Section~\ref{sec:mask-quality} reports the accuracy of the projection-derived masks and their sensitivity to the projection operator, and Section~\ref{sec:recognition} presents vertebra- and patient-level fracture recognition, including the handling of class imbalance, the choice of operating point and the comparison with published methods. Section~\ref{sec:error-propagation} analyzes how localization and segmentation errors propagate to the diagnosis, Section~\ref{sec:decision-evidence} covers explainability and agreement with radiologists, and Section~\ref{sec:component-selection} summarizes how each component of the pipeline was selected.

\subsection{Projection-Derived versus 3D-Segmentation Masks}\label{sec:mask-source}

To isolate the effect of the mask source, the classifier, training protocol and cross-validation folds were held fixed, and only the masks used to extract the vertebra VOIs were changed (Section~\ref{sec:mask-source-arm}). Differences were assessed with the paired analyses of Section~\ref{sec:statistics}.

Table~\ref{tab:mask_source} reports AUPRC and ROC-AUC for both mask sources across the three classifier configurations. At both diagnostic levels, AUPRC differed between the two mask sources by about one point or less, and the sign of the difference varied between configurations. The paired comparisons of the score-fusion ensemble are listed in Table~\ref{tab:paired_stats}, and Figure~\ref{fig:paired_forest} shows their 95\% CIs. Against the 3D-segmentation masks, the AUPRC differences were $+1.03$ points per vertebra and $+0.65$ points per patient, and their 95\% CIs contained zero in both analyses (fold-wise $p = 0.315$ and $0.116$). The widest loss bound was for vertebra-level AUPRC, at 1.46 points in the fold-wise analysis and 0.46 points in the bootstrap, so losses larger than these were not supported by the data.

\begin{table}[htbp]
  \centering
  \caption{Fracture recognition with projection-derived masks (Proj.) and 3D-segmentation masks (3D). Both pipelines use the same classifier, training protocol and five cross-validation folds; only the masks used to extract the VOIs differ. Values are means over the five test folds (\%), and $\Delta$ is projection-derived minus 3D-segmentation masks, in percentage points. Full metric profiles and per-fold values are given in Supplementary Sections S8 and S9.}
  \label{tab:mask_source}
  \footnotesize
  \setlength{\tabcolsep}{3pt}
  \begin{tabular}{llcccccc}
    \toprule
    \multirow{2}{*}{Classifier} & \multirow{2}{*}{Level} & \multicolumn{3}{c}{AUPRC} & \multicolumn{3}{c}{ROC-AUC} \\
    \cmidrule(lr){3-5} \cmidrule(lr){6-8}
     & & Proj. & 3D & $\Delta$ & Proj. & 3D & $\Delta$ \\
    \midrule
    \multirow{2}{*}{Slice stack} & Vertebra & 72.14 & 71.11 & $+1.03$ & 90.20 & 88.94 & $+1.26$ \\
     & Patient & 90.42 & 89.67 & $+0.75$ & 89.09 & 87.93 & $+1.16$ \\
    \midrule
    \multirow{2}{*}{MIP stack} & Vertebra & 71.89 & 72.69 & $-0.80$ & 89.85 & 89.46 & $+0.39$ \\
     & Patient & 90.58 & 90.55 & $+0.03$ & 89.15 & 88.91 & $+0.24$ \\
    \midrule
    \multirow{2}{*}{Score fusion} & Vertebra & 75.64 & 74.61 & $+1.03$ & 91.63 & 90.37 & $+1.26$ \\
     & Patient & 92.00 & 91.36 & $+0.65$ & 90.95 & 89.72 & $+1.23$ \\
    \bottomrule
  \end{tabular}
\end{table}

\begin{table*}[t]
  \centering
  \caption{Paired comparisons of the proposed score-fusion ensemble with the 3D-segmentation masks and with the reproduced RSNA 2022 winning solution. Values are means over the five test folds (\%), and $\Delta$ is proposed minus comparator, in percentage points; because $\Delta$ is computed from unrounded per-fold values, it can differ by 0.01 from the difference of the rounded means. Every pipeline is scored on all 14,133 vertebrae and 2,019 patients. The fold-wise analysis is a paired $t$-test over the five test folds (df = 4); the bootstrap resamples whole patients within folds (2,000 replicates). All intervals are two-sided 95\% CIs. Results with the missed vertebrae excluded are given in Supplementary Section S8.}
  \label{tab:paired_stats}
  \footnotesize
  \setlength{\tabcolsep}{4pt}
  \begin{tabular*}{\textwidth}{@{\extracolsep{\fill}}lcccccc}
    \toprule
     & & & & \multicolumn{2}{c}{Fold-wise paired $t$-test} & Patient-clustered bootstrap \\
    \cmidrule(lr){5-6} \cmidrule(lr){7-7}
    Level and metric & Proposed & Comparator & $\Delta$ & 95\% CI & $p$ & 95\% CI \\
    \midrule
    \multicolumn{7}{l}{\textit{Against 3D-segmentation masks}} \\
    Vertebra AUPRC   & 75.64 & 74.61 & $+1.03$ & $[-1.46, +3.51]$ & 0.315 & $[-0.46, +2.52]$ \\
    Vertebra ROC-AUC & 91.63 & 90.37 & $+1.26$ & $[+0.02, +2.49]$ & 0.047 & $[+0.32, +2.26]$ \\
    Patient AUPRC    & 92.00 & 91.36 & $+0.65$ & $[-0.25, +1.55]$ & 0.116 & $[-0.32, +1.67]$ \\
    Patient ROC-AUC  & 90.95 & 89.72 & $+1.23$ & $[+0.13, +2.32]$ & 0.036 & $[-0.04, +2.53]$ \\
    \midrule
    \multicolumn{7}{l}{\textit{Against the reproduced RSNA 2022 winning solution}} \\
    Vertebra AUPRC   & 75.64 & 76.19 & $-0.55$ & $[-2.81, +1.71]$ & 0.537 & $[-2.07, +1.03]$ \\
    Vertebra ROC-AUC & 91.63 & 92.15 & $-0.52$ & $[-1.66, +0.62]$ & 0.275 & $[-1.40, +0.30]$ \\
    Patient AUPRC    & 92.00 & 91.86 & $+0.15$ & $[-1.73, +2.03]$ & 0.836 & $[-0.93, +1.20]$ \\
    Patient ROC-AUC  & 90.95 & 90.75 & $+0.19$ & $[-2.73, +3.11]$ & 0.864 & $[-1.01, +1.42]$ \\
    \bottomrule
  \end{tabular*}
\end{table*}

\begin{figure*}[t]
  \centering
  \includegraphics[width=\linewidth]{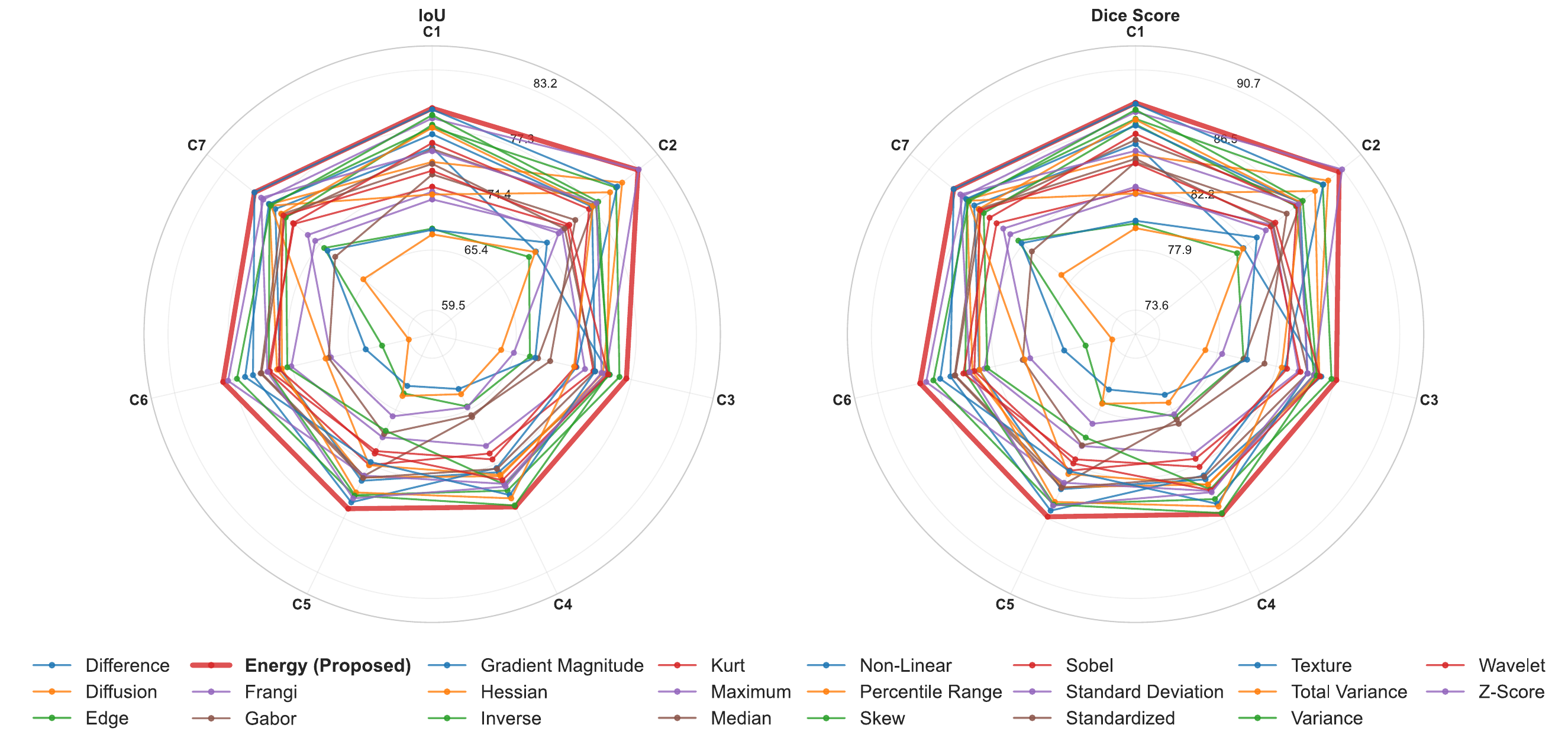}
    \caption{IoU (left) and Dice (right) per vertebra for the DenseNet121-Unet with each of the 23 projection operators (five-fold cross-validation, 87 patients). The energy projection is drawn as the thick red line.}
  \label{fig:projection_selection_radar}
\end{figure*}

\begin{table}[htbp]
  \centering
  \caption{Per-vertebra segmentation accuracy of the DenseNet121-Unet on the energy projection (five-fold cross-validation, 87 patients). The overall IoU and Dice reported in the text and in Table~\ref{tab:segmentation_comparison} pool intersections and mask areas over all patients and vertebral channels within each fold before the ratio is taken, which weights each vertebra by its size, so they differ slightly from the means of the seven rows (Dice 87.86\% against 87.73\%, IoU 78.45\% against 78.53\%).}
  \label{tab:vertebra_seg_performance}
  \resizebox{\linewidth}{!}{%
    \begin{tabular}{lccccccccc}
      \toprule
      \multirow{2}{*}{Level} & \multicolumn{3}{c}{Sagittal} & \multicolumn{3}{c}{Coronal} & \multicolumn{3}{c}{Average} \\
      \cmidrule(lr){2-4} \cmidrule(lr){5-7} \cmidrule(lr){8-10}
       & IoU$\uparrow$ & Dice$\uparrow$ & HD95$\downarrow$ & IoU$\uparrow$ & Dice$\uparrow$ & HD95$\downarrow$ & IoU$\uparrow$ & Dice$\uparrow$ & HD95$\downarrow$ \\
      \midrule
      C1 & 79.23 & 88.19 & 4.08 & 79.70 & 88.60 & 3.63 & 79.46 & 88.39 & 3.86 \\
      C2 & 83.16 & 90.22 & 2.50 & 83.15 & 90.72 & 3.35 & 83.15 & 90.47 & 2.93 \\
      C3 & 77.86 & 87.21 & 3.80 & 75.79 & 85.98 & 5.61 & 76.82 & 86.60 & 4.70 \\
      C4 & 77.34 & 86.97 & 4.27 & 74.81 & 85.33 & 5.80 & 76.07 & 86.15 & 5.04 \\
      C5 & 75.42 & 85.75 & 5.33 & 77.13 & 86.96 & 5.03 & 76.28 & 86.35 & 5.18 \\
      C6 & 77.26 & 86.94 & 4.77 & 79.40 & 88.38 & 4.32 & 78.33 & 87.66 & 4.55 \\
      C7 & 79.97 & 88.69 & 4.41 & 79.24 & 88.30 & 6.04 & 79.61 & 88.50 & 5.22 \\
      \bottomrule
    \end{tabular}%
  }
\end{table}

The ROC-AUC differences pointed in the same direction ($+1.26$ points per vertebra and $+1.23$ per patient). Three of their four 95\% CIs excluded zero, but the patient-level bootstrap interval did not, and the vertebra-level fold-wise interval included zero once the missed vertebrae were excluded. Because the classifier configuration was selected on projection-derived VOIs, we do not read these differences as an advantage of the projection-derived masks (Section~\ref{sec:disc-viability}).

Against the reproduced RSNA 2022 winning solution, the widest loss bound was again for vertebra-level AUPRC (75.64 versus 76.19, $\Delta = -0.55$), at 2.81 points in the fold-wise analysis and 2.07 points in the bootstrap. Patient-level AUPRC was 92.00 versus 91.86 ($\Delta = +0.15$). All four pooled 95\% CIs contained zero in both analyses. The ROC-AUC differences were $-0.52$ points per vertebra and $+0.19$ per patient. The complete metric profiles are given in Table~\ref{tab:literature_comparison}.

Per-level bootstrap comparisons are given in Supplementary Section S8. Against the 3D-segmentation masks, no per-level 95\% CI lay below zero. Projection-derived masks gave higher AUPRC and ROC-AUC at C4 and C6, with 95\% CIs excluding zero, whereas the AUPRC point estimates at C1--C3 favored the 3D-segmentation masks ($-2.30$, $-1.77$ and $-3.51$ points) with 95\% CIs that included zero. Against the reproduced winning solution, no per-level 95\% CI excluded zero, and the largest AUPRC differences in its favor were at C3 ($-6.18$) and C7 ($-2.68$). The pattern at the upper cervical levels is examined in Section~\ref{sec:disc-patterns}.

Excluding the 27 vertebrae missed by any pipeline, together with their five patients, changed no pooled $\Delta$ by more than 0.14 points and no lower 95\% limit by more than 0.16 points. The only change in significance was the vertebra-level ROC-AUC interval noted above (Supplementary Section S8).

\subsection{Quality of the Projection-Derived Masks}\label{sec:mask-quality}

\subsubsection{Choice of Projection Operator}\label{sec:projection-choice}

The sagittal and coronal projections are utilized to determine and localize the vertebrae. The projection operators therefore determine how well the vertebrae are delineated rather than which image detail reaches the classifier. We compared 23 operators (Supplementary Section S3) with the DenseNet121-Unet on both views, using five-fold cross-validation on the 87 patients with ground-truth masks, and ranked them by the composite score defined in Section~\ref{sec:metrics}. Figure~\ref{fig:projection_selection_radar} shows IoU and Dice per vertebra for every operator, and the full results are given in Supplementary Table S4.

The energy projection ranked first, with the highest IoU and Dice and the lowest HD95 in both views, a mean Dice of 87.86\% and an HD95 of 4.50~mm. Its margin over the next operators was small, since the standard deviation, variance, inversion and nonlinear projections all lay within 1.6 Dice points of it. Across all 23 operators, mean Dice ranged from 78.15\% to 87.86\% and HD95 from 4.50 to 10.55~mm. This 9.7-point spread compares with a 2.5-point spread across the ten segmentation architectures evaluated on the energy projection in Section~\ref{sec:component-selection}.

\begin{figure*}[t]
  \centering
  \includegraphics[width=\textwidth]{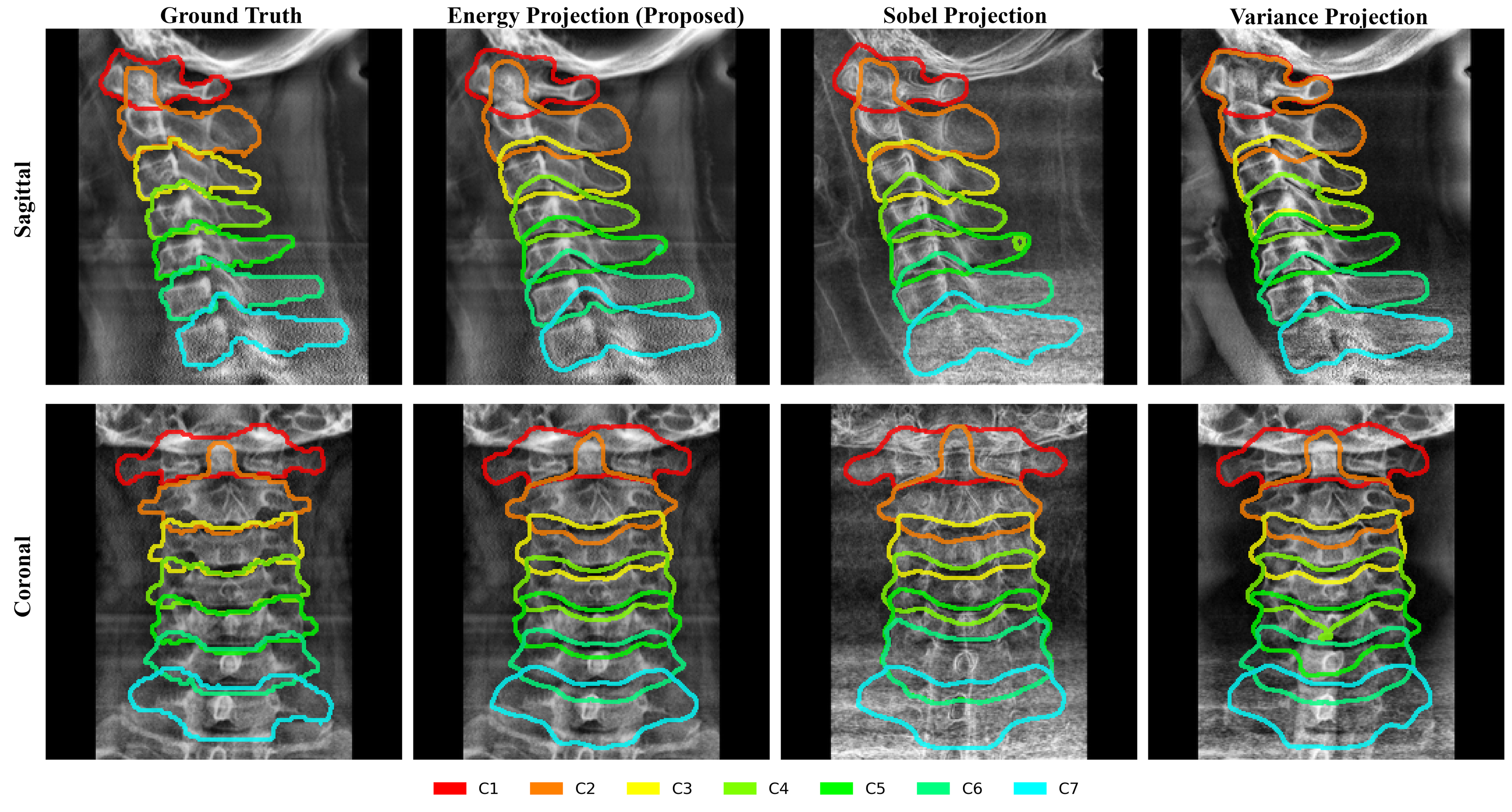}
  \caption{Multi-label segmentation of one patient by the DenseNet121-Unet with the energy (proposed), Sobel and variance projections, shown with the ground truth in the sagittal (top) and coronal (bottom) views. Contour colors identify C1--C7.}
  \label{fig:projection_comparison}
\end{figure*}

The four lowest-ranked operators, maximum gradient, edge, gradient and Hessian, are all edge- or derivative-based and trailed energy by 6.8 to 9.7 Dice points. The radar plots in Figure~\ref{fig:projection_selection_radar} show how these differences are distributed across levels. Energy lay on or near the outer boundary of both plots at every level, with other high-ranked operators close to it at C1, C2 and C7. Most operators followed a similar per-level profile, with the highest IoU and Dice at C2 and lower values at the mid-cervical levels, and the edge- and derivative-based operators formed the innermost polygons, falling furthest at C3--C5. This ordering is consistent with the motivation for the energy projection given in Section~\ref{sec:projection-formulation}, since energy amplifies high-intensity structures such as cortical bone and so strengthens vertebral boundaries, whereas edge-based operators retain mainly sparse edge responses. Figure~\ref{fig:projection_comparison} shows the masks obtained with the energy, Sobel and variance projections for one patient, in which the energy masks followed the ground-truth outlines most closely in both views.

Two preprocessing steps applied before projection also affected accuracy. Bone windowing raised mean Dice and IoU by 1.82 and 2.60 points and lowered HD95 from 6.09 to 4.50~mm, and interpolating each volume to at least 400 slices raised mean Dice by 5.51 points and lowered HD95 by 2.80~mm (Supplementary Section S6.3). All later experiments used the energy projection with both steps. Spine localization used the variance projection, selected among four operators in Section~\ref{sec:component-selection}.

\subsubsection{Per-Vertebra Segmentation Accuracy}\label{sec:seg-accuracy}

With the energy projection, the DenseNet121-Unet reached a mean Dice of 87.86\%, an IoU of 78.45\% and an HD95 of 4.50~mm averaged over the sagittal and coronal views. Table~\ref{tab:vertebra_seg_performance} breaks the results down by vertebra, and the pattern across levels closely follows the anatomy of each vertebra.

C2 (axis) was segmented most accurately, with the highest IoU and Dice and the lowest HD95 (2.93~mm), consistent with its distinctive dens, which gives the network a stable landmark. C1 (atlas) and C7 (vertebra prominens) also reached Dice close to 88.5\%, helped by the ring shape of C1 and the prominent, non-bifid spinous process of C7. C7 nevertheless had the largest HD95, mainly in the coronal view (6.04~mm), suggesting that its broad transverse processes and the C7--T1 junction make the boundary harder to place.

The mid-cervical vertebrae C3--C5 were the hardest to segment, with lower IoU and Dice than C1, C2 and C7 and higher HD95 than C1 and C2. These vertebrae are similar in shape, with comparable bodies and bifid spinous processes, and their spinous processes are often hidden behind the vertebral bodies in the coronal projection. C6 performed between the two groups. Even so, mean Dice stayed within 4.4 points across the seven levels, from 86.15\% at C4 to 90.47\% at C2. These per-level differences are revisited in Section~\ref{sec:disc-patterns}.

Because the masks define the volumes in which the classifier later looks for fractures, we also checked whether a fracture itself makes a vertebra harder to segment. Table~\ref{tab:seg_subgroups} compares fractured and non-fractured vertebrae at each level, and Supplementary Figure S19 plots the per-level differences. Averaged over the seven levels, Dice was 88.1\% for the 85 fractured vertebrae and 87.7\% for the 524 non-fractured vertebrae. In the patient-clustered bootstrap, the level-matched difference (fractured minus non-fractured) was $+0.41$ points and not significant, and its 95\% CI ranged from a loss of 0.53 points to a gain of 1.43 points for fractured vertebrae. No level showed a significant loss on fractured vertebrae, since no per-level 95\% CI lay below zero, and the one level whose interval excluded zero, C3, favored the fractured vertebrae. Ankylosed vertebrae, analyzed in Supplementary Section S8.5, were segmented 1.20 Dice points below non-ankylosed vertebrae at the same level (95\% CI $-2.18$ to $-0.21$). Fracture recognition did not differ significantly between the two groups, although only 30 ankylosed vertebrae were fractured. Because fracture prevalence differs between them, the comparison was made on ROC-AUC, which was 3.91 points lower for ankylosed vertebrae (95\% CI $-12.82$ to $+3.25$).

\begin{table}[htbp]
  \centering
  \caption{Dice (\%) of the projection-derived masks for fractured and non-fractured vertebrae, per level, with the number of vertebrae in parentheses, and the difference $\Delta$ (fractured minus non-fractured, Dice points) with its 95\% CI from the patient-clustered bootstrap. Dice is computed per vertebra and averaged over all patients at each level; Table~\ref{tab:vertebra_seg_performance} averages within folds first, which changes whole-level values by at most 0.04 points. The bottom row is the level-matched mean.}
  \label{tab:seg_subgroups}
  \footnotesize
  \setlength{\tabcolsep}{4pt}
  \begin{tabular}{lccc}
    \toprule
    Level & Fractured & Non-fractured & $\Delta$ [95\% CI] \\
    \midrule
    C1 & 89.44 (5)  & 88.34 (82) & $+1.10$ [$-1.17$, $+3.11$] \\
    C2 & 90.95 (13) & 90.41 (74) & $+0.53$ [$-1.34$, $+2.57$] \\
    C3 & 88.57 (11) & 86.29 (76) & $+2.28$ [$+0.64$, $+4.00$] \\
    C4 & 86.04 (9)  & 86.19 (78) & $-0.15$ [$-2.99$, $+2.53$] \\
    C5 & 86.50 (14) & 86.37 (73) & $+0.14$ [$-1.79$, $+1.89$] \\
    C6 & 87.09 (20) & 87.86 (67) & $-0.78$ [$-3.27$, $+1.32$] \\
    C7 & 88.29 (13) & 88.56 (74) & $-0.27$ [$-1.94$, $+1.29$] \\
    \midrule
    Mean of C1--C7 & 88.13 (85) & 87.72 (524) & $+0.41$ [$-0.53$, $+1.43$] \\
    \bottomrule
  \end{tabular}
\end{table}

\subsection{Fracture Recognition on the Full Imbalanced Cohort}\label{sec:recognition}

\subsubsection{Vertebra- and Patient-Level Recognition}\label{sec:recognition-levels}

We evaluated three families of classifiers to identify fractures in individual cervical vertebrae (C1--C7): projection-based 2D models, full-volume 3D models and 2.5D models. The 2D models were trained and evaluated on variance projections of each vertebra in each orthogonal view, and the 3D and 2.5D models on bone-windowed slices. Table~\ref{tab:vertebra_fracture_performance} reports the results.

The 2D classifiers, although computationally lightweight, performed poorly. The single-view models had F1-scores between 18.67\% and 21.72\% and AUPRC between 12.93\% and 21.38\%, and the coronal model reached a sensitivity of 96.81\% only at a specificity of 4.37\%, labeling nearly every vertebra as fractured. Fusing the three views into a three-channel input did not help (F1-score 19.93\%, AUPRC 20.25\%), plausibly because the views lack spatial correspondence. The 3D EfficientNet-B0 used the full volumetric context and reached a specificity of 97.99\% and an AUPRC of 32.51\%, but its sensitivity was 17.11\% and its F1-score 25.39\%, consistent with overfitting to the high-dimensional 3D input.

The proposed 2.5D CNN-Transformer gave the highest F1-scores and AUPRC among the evaluated classifiers, with F1-scores of 66.53\% on slice stacks and 64.61\% on MIP stacks and AUPRC of 72.14\% and 71.89\%. Score-level fusion of the two variants raised AUPRC to 75.64\% and ROC-AUC to 91.63\%, above both. With 3D-segmentation masks, the same fusion added the same 3.50 AUPRC points over the slice-stack model (Table~\ref{tab:mask_source}), so the gain does not depend on the source of the masks. At the default threshold of 0.50, fusion raised precision from 72.16\% to 84.16\% and the F1-score from 66.53\% to 68.15\%, while sensitivity fell from 62.33\% to 57.49\%. The effect of the threshold is examined in Section~\ref{sec:operating-point}.

\begin{table*}[htbp]
  \centering
  \caption{Vertebra-level fracture recognition by input type and model (\%, means over the five test folds). The AUPRC chance level equals the prevalence of fractured vertebrae, 10.2\%.}
  \label{tab:vertebra_fracture_performance}
  \resizebox{\textwidth}{!}{%
    \begin{tabular}{lllccccccc}
      \toprule
      Input & View & Model & Accuracy & Precision & Sensitivity & Specificity & F1-Score & ROC-AUC & AUPRC \\
      \midrule
      2D & Axial & DenseNet121 & 87.52 & 29.80 & 16.34 & 95.62 & 21.11 & 68.70 & 21.38 \\
      2D & Coronal & DenseNet121 & 13.82 & 10.33 & 96.81 & 4.37 & 18.67 & 56.11 & 12.93 \\
      2D & Sagittal & DenseNet121 & 50.05 & 12.93 & 67.80 & 48.03 & 21.72 & 60.57 & 14.05 \\
      2D & Fused & DenseNet121 & 87.20 & 27.64 & 15.58 & 95.36 & 19.93 & 68.14 & 20.25 \\
      2D & Fused & ResNet18 & 60.63 & 10.33 & 37.12 & 63.31 & 16.16 & 51.86 & 10.72 \\
      3D & - & EfficientNet-B0 & 89.72 & 49.20 & 17.11 & 97.99 & 25.39 & 73.42 & 32.51 \\
      \midrule
      2.5D (Slice Stacks) & Axial & Proposed Network & 93.55 & 72.16 & 62.33 & 97.09 & 66.53 & 90.20 & 72.14 \\
      2.5D (Max Proj.) & Axial & Proposed Network & 93.70 & 76.82 & 56.55 & 97.95 & 64.61 & 89.85 & 71.89 \\
      - & - & Proposed Network (Score Fusion) & 94.51 & 84.16 & 57.49 & 98.74 & 68.15 & 91.63 & 75.64 \\
      \bottomrule
    \end{tabular}%
  }
\end{table*}

\begin{table*}[htbp]
  \centering
  \caption{Patient-level fracture recognition by input type, model and aggregation rule (\%, means over the five test folds). The AUPRC chance level equals the prevalence of fractured patients, 47.6\%.}
  \label{tab:patient_performance}
  \resizebox{\textwidth}{!}{%
    \begin{tabular}{lllcccccccc}
      \toprule
      Input & View & Model & Aggregation & Accuracy & Precision & Sensitivity & Specificity & F1-Score & ROC-AUC & AUPRC \\
      \midrule
      2D (whole) & Coronal & DenseNet121 & - & 60.99 & 59.16 & 58.55 & 63.21 & 58.85 & 64.63 & 61.70 \\
      2D (whole) & Sagittal & ResNet50 & - & 59.26 & 64.58 & 32.12 & 83.96 & 42.90 & 64.46 & 58.78 \\
      2D (Fused) & - & ResNet50 (Fusion) & - & 66.67 & 66.86 & 59.59 & 73.11 & 63.01 & 69.76 & 65.25 \\
      \midrule
      2D & Axial & DenseNet121 & If Any & 61.42 & 65.64 & 39.75 & 81.10 & 49.51 & 65.96 & 62.57 \\
      2D & Coronal & DenseNet121 & If Any & 47.60 & 47.60 & 100.00 & 0.00 & 64.50 & 53.78 & 51.15 \\
      2D & Sagittal & DenseNet121 & If Any & 48.34 & 47.93 & 98.65 & 2.65 & 64.51 & 55.06 & 52.60 \\
      \midrule
      3D & - & EfficientNet-B0 & If Any & 62.00 & 74.14 & 30.56 & 90.43 & 41.38 & 65.79 & 67.19 \\
      \midrule
      2.5D (Slice Stacks) & Axial & Proposed Network & If Any & 82.67 & 86.15 & 76.18 & 88.38 & 80.66 & 89.09 & 90.42 \\
      2.5D (Max Proj.) & - & Proposed Network & If Any & 81.18 & 88.87 & 69.79 & 91.64 & 77.87 & 89.15 & 90.58 \\
      - & - & Proposed Fusion Network & If Any & 83.06 & 93.54 & 69.31 & 95.54 & 79.58 & 90.95 & 92.00 \\
      - & - & Proposed Fusion Network & Adaptive Avg & 83.06 & 82.12 & 82.51 & 83.56 & 82.28 & 90.95 & 92.00 \\
      \bottomrule
    \end{tabular}%
  }
\end{table*}

\begin{figure}[htbp]
  \centering
  \includegraphics[width=\linewidth]{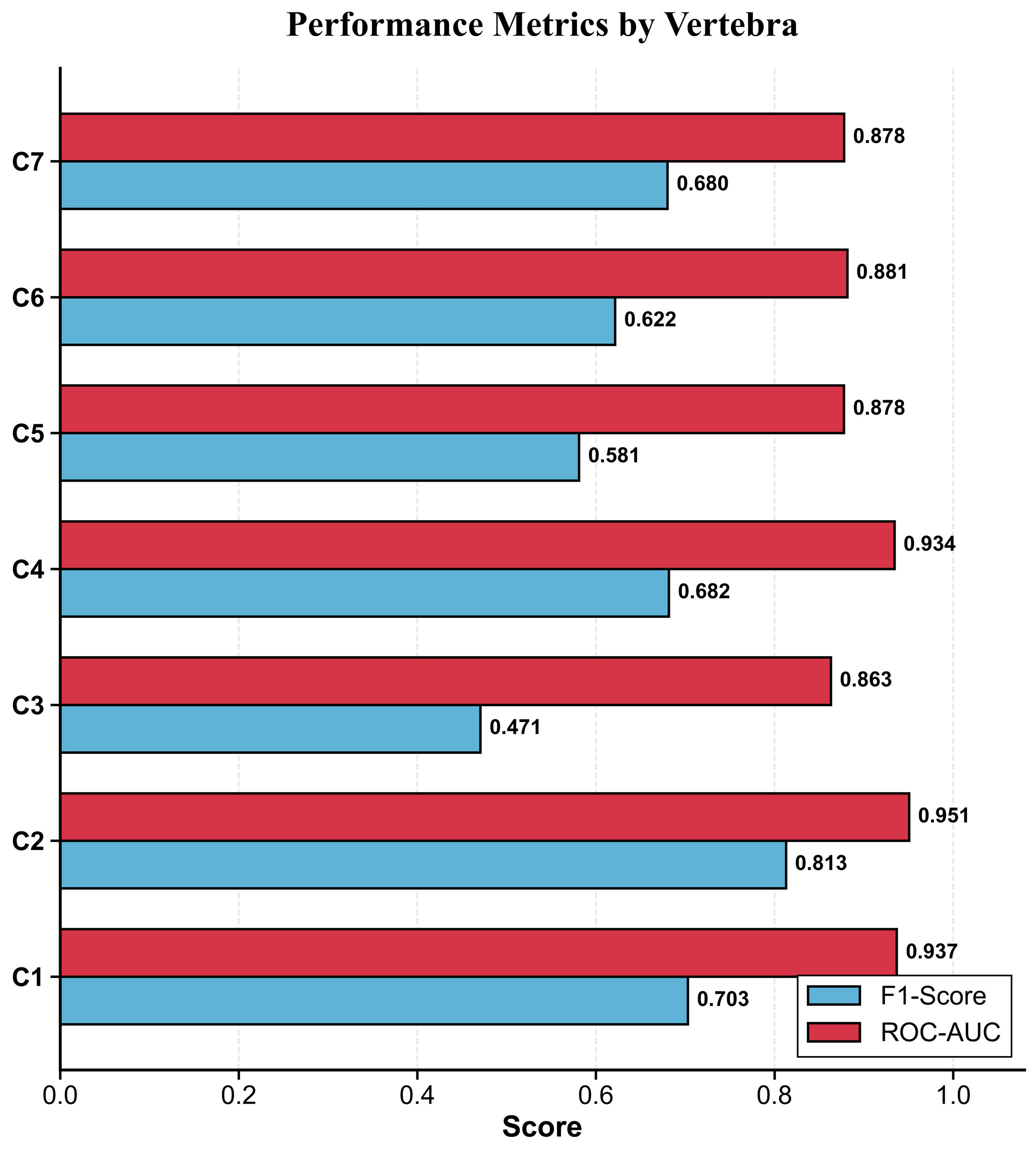}
  \caption{F1-score and ROC-AUC of the score-fusion model for each vertebra, C1--C7.}
  \label{fig:cls_results_vertebra_wise}
\end{figure}

Figure~\ref{fig:cls_results_vertebra_wise} breaks down the fusion model's performance by vertebra. The highest F1-scores were at C2 and C1, consistent with their distinctive anatomy and, for C2, its high fracture prevalence. The lowest F1-score was at C3, the least frequently fractured level (Supplementary Figure S1). C4 was recognized better than C6 despite having fewer fractured examples, and C7, the most frequently fractured level, was close to C4, so prevalence alone does not explain the pattern across levels. ROC-AUC exceeded 0.85 at every level. These per-level patterns are discussed in Section~\ref{sec:disc-patterns}.

A reliable patient-level diagnosis matters for clinical decision-making, so we compared two strategies, classifying whole-spine 2D projections directly and aggregating vertebra-level predictions into a patient-level outcome (Table~\ref{tab:patient_performance}). On whole-spine projections, DenseNet121 reached an F1-score of 58.85\% on the coronal view and ResNet50 an F1-score of 42.90\% on the sagittal view, and fusing the two views with ResNet50 reached 63.01\%, with AUPRC between 58.78\% and 65.25\% against a chance level of 47.6\%. Compressing the whole cervical volume into a single 2D view likely obscures the fine, localized patterns of fractures.

We then aggregated vertebra-level predictions with an If-Any rule, which classifies a patient as fractured when any of the seven vertebrae is predicted fractured. With the 2D vertebra classifiers this rule was unreliable. The axial model reached an F1-score of 49.51\%, while the coronal and sagittal models flagged almost every patient, with sensitivities of 100.00\% and 98.65\% at specificities of 0.00\% and 2.65\%, reflecting the excess false positives of their vertebra-level predictions. The 3D model under If-Any reached an F1-score of 41.38\% and an AUPRC of 67.19\%.

Applied to the 2.5D models, which already performed well at the vertebra level, the same rule gave a patient-level F1-score of 80.66\%, a ROC-AUC of 89.09\% and an AUPRC of 90.42\% for the slice-stack model. Fusing the two variants raised precision to 93.54\% and specificity to 95.54\%, but sensitivity fell to 69.31\%, which in a diagnostic setting means more fractured patients are missed. The adaptive-average aggregation, which sets the decision threshold from the agreement between the two variants, gave a more balanced result, with a sensitivity of 82.51\%, a specificity of 83.56\%, a precision of 82.12\% and an F1-score of 82.28\%. Because 47.6\% of patients have at least one fracture, its accuracy of 83.06\% is also informative at this level. The patient-level AUPRC of the fusion model was 92.00\%, against 90.42\% and 90.58\% for the slice-stack and MIP-stack models.

\subsubsection{Handling Class Imbalance}\label{sec:imbalance}

With fractures in 10.2\% of vertebrae, class imbalance affects how the classifier is trained, how it is evaluated and where its decision threshold is placed, and the study addresses it at each of these points. Every model is evaluated at the native prevalence, the decision threshold is examined in Section~\ref{sec:operating-point}, and Section~\ref{sec:balanced} shows how protocols built on class-balanced subsets change the reported results. For the training objective, Table~\ref{tab:imbalance} compares the seven strategies of Section~\ref{sec:imbalance-protocol}, which either reweight the loss or resample the training data while all other settings are held fixed.

At the vertebra level, a positive-class weight of 2 gave the highest mean AUPRC (72.14\%), ROC-AUC (90.20\%) and F1-score (66.53\%). A weight of 3 and focal loss were within about 2 AUPRC points of it, at 70.21\% and 70.03\%, while unweighted training and prevalence weighting were 2.84 and 3.48 points lower, at 69.30\% and 68.66\%. These differences are of the same order as the fold-to-fold standard deviations, so we describe the weight of 2 as giving the highest mean rather than as significantly better. Both sampling strategies scored lower. Oversampling the fractured vertebrae, which rebalances the training data without discarding any of it, reached an AUPRC of 65.34\%. Undersampling the non-fractured vertebrae, which discards most of them, reached 41.83\% with the largest variation across folds (standard deviation 15.20 points).

\begin{table*}[htbp]
  \centering
  \caption{Class-imbalance strategies for the slice-stack model (\%, mean $\pm$ standard deviation over the five test folds). All other training settings, including data augmentation and Mixup, are identical across rows. The undersampling row is Setting B of Table~\ref{tab:balanced}.}
  \label{tab:imbalance}
  \footnotesize
  \begin{tabular*}{\textwidth}{@{\extracolsep{\fill}}lccccccc}
    \toprule
    Strategy & AUPRC & ROC-AUC & F1-Score & Sensitivity & Specificity & Precision & Accuracy \\
    \midrule
    \multicolumn{8}{l}{\textit{Vertebra level}} \\
    BCE, unweighted & $69.30 \pm 2.60$ & $88.85 \pm 0.77$ & $61.46 \pm 2.97$ & $49.72 \pm 4.23$ & $98.62 \pm 0.62$ & $81.12 \pm 5.46$ & $93.63 \pm 0.74$ \\
    BCE, weight 2 (proposed) & $72.14 \pm 2.15$ & $90.20 \pm 1.06$ & $66.53 \pm 3.12$ & $62.33 \pm 4.31$ & $97.09 \pm 1.41$ & $72.16 \pm 8.51$ & $93.55 \pm 1.18$ \\
    BCE, weight 3 & $70.21 \pm 1.24$ & $89.08 \pm 0.81$ & $62.92 \pm 2.34$ & $64.85 \pm 2.76$ & $95.21 \pm 1.70$ & $61.57 \pm 6.23$ & $92.12 \pm 1.44$ \\
    BCE, prevalence weight (8.78) & $68.66 \pm 2.56$ & $88.52 \pm 1.64$ & $50.80 \pm 4.64$ & $74.08 \pm 7.62$ & $86.12 \pm 5.80$ & $39.57 \pm 7.43$ & $84.91 \pm 4.51$ \\
    Focal loss ($\alpha = 2$, $\gamma = 2$) & $70.03 \pm 2.67$ & $89.06 \pm 0.90$ & $63.33 \pm 3.08$ & $61.36 \pm 5.06$ & $96.28 \pm 1.42$ & $66.15 \pm 6.83$ & $92.71 \pm 1.27$ \\
    Oversampling & $65.34 \pm 3.20$ & $87.35 \pm 1.69$ & $59.68 \pm 2.82$ & $51.13 \pm 3.45$ & $97.72 \pm 0.84$ & $72.35 \pm 7.15$ & $92.94 \pm 0.75$ \\
    Undersampling & $41.83 \pm 15.20$ & $79.79 \pm 4.02$ & $41.30 \pm 9.29$ & $49.84 \pm 11.70$ & $89.62 \pm 2.00$ & $35.43 \pm 7.98$ & $85.59 \pm 2.22$ \\
    \midrule
    \multicolumn{8}{l}{\textit{Patient level (If-Any)}} \\
    BCE, unweighted & $88.97 \pm 2.27$ & $87.20 \pm 2.78$ & $74.71 \pm 3.80$ & $63.17 \pm 5.79$ & $94.76 \pm 2.89$ & $91.93 \pm 3.59$ & $79.74 \pm 2.68$ \\
    BCE, weight 2 (proposed) & $90.42 \pm 0.90$ & $89.09 \pm 1.09$ & $80.66 \pm 2.35$ & $76.18 \pm 4.95$ & $88.38 \pm 5.25$ & $86.15 \pm 5.19$ & $82.67 \pm 1.91$ \\
    BCE, weight 3 & $89.64 \pm 1.00$ & $87.82 \pm 1.66$ & $79.16 \pm 2.14$ & $79.31 \pm 2.74$ & $80.43 \pm 7.36$ & $79.22 \pm 4.79$ & $80.04 \pm 3.02$ \\
    BCE, prevalence weight (8.78) & $88.89 \pm 1.19$ & $86.63 \pm 1.08$ & $72.64 \pm 1.95$ & $90.97 \pm 4.66$ & $45.15 \pm 15.43$ & $60.75 \pm 4.45$ & $67.16 \pm 5.29$ \\
    Focal loss ($\alpha = 2$, $\gamma = 2$) & $89.57 \pm 1.63$ & $87.94 \pm 2.21$ & $79.36 \pm 2.11$ & $76.35 \pm 2.63$ & $84.95 \pm 7.04$ & $82.85 \pm 5.00$ & $80.98 \pm 3.27$ \\
    Oversampling & $86.52 \pm 1.58$ & $85.01 \pm 1.49$ & $73.75 \pm 3.18$ & $65.92 \pm 4.89$ & $88.44 \pm 3.60$ & $84.04 \pm 3.88$ & $77.66 \pm 3.29$ \\
    Undersampling & $71.24 \pm 12.45$ & $72.23 \pm 8.87$ & $67.40 \pm 6.87$ & $77.65 \pm 9.23$ & $52.16 \pm 11.13$ & $59.80 \pm 6.50$ & $64.39 \pm 7.30$ \\
    \bottomrule
  \end{tabular*}
\end{table*}

\begin{table*}[htbp]
  \centering
  \caption{Slice-stack model under three training and testing protocols, vertebra level (\%, means over the five test folds). A: proposed protocol, full cohort at native prevalence, with class weighting. B: balanced training split (undersampling of non-fractured vertebrae; the undersampling row of Table~\ref{tab:imbalance}), native-prevalence test. C: the same models as B, balanced test set. The balanced test set contains 2,888 vertebrae (1,444 fractured). Precision, F1-score and AUPRC depend on prevalence, so they are comparable only between settings with the same test set.}
  \label{tab:balanced}
  \footnotesize
  \begin{tabular*}{\textwidth}{@{\extracolsep{\fill}}llcccccc}
    \toprule
    Setting & Test set & AUPRC & ROC-AUC & Precision & Sensitivity & Specificity & F1-Score \\
    \midrule
    A & Native prevalence & 72.14 & 90.20 & 72.16 & 62.33 & 97.09 & 66.53 \\
    B & Native prevalence & 41.83 & 79.79 & 35.43 & 49.84 & 89.62 & 41.30 \\
    C & Balanced & 80.61 & 79.82 & 82.64 & 49.84 & 89.90 & 61.72 \\
    \bottomrule
  \end{tabular*}
\end{table*}

The strategies also moved the operating point at the default threshold of 0.50. Unweighted training gave the highest vertebra-level precision (81.12\%) and the lowest sensitivity (49.72\%). Weighting the fractured class shifted the balance toward sensitivity, which rose to 62.33\% with a weight of 2, 64.85\% with a weight of 3 and 74.08\% with prevalence weighting, while precision fell to 72.16\%, 61.57\% and 39.57\%. Of these settings, the weight of 2 gave the highest F1-score, the harmonic mean of precision and sensitivity.

The patient-level results followed the same ordering. A weight of 2 gave the highest mean AUPRC (90.42\%), ROC-AUC (89.09\%) and F1-score (80.66\%), with the smallest fold-to-fold variation in AUPRC (standard deviation 0.90 points). Unweighted training gave the highest patient-level specificity (94.76\%) and precision (91.93\%) but a sensitivity of 63.17\%, and undersampling reached a sensitivity of 77.65\% only at a specificity of 52.16\%. Prevalence weighting gave the highest patient-level sensitivity of all strategies (90.97\%), but at the lowest specificity (45.15\%). The weight of 2 was used for all other experiments.

Data augmentation and Mixup act on both classes alike, so they were not counted among the imbalance strategies. In a separate ablation of the proposed configuration, adding Mixup raised the vertebra-level F1-score from 61.61\% to 66.53\% and ROC-AUC from 88.33\% to 90.20\%. Most of the gain came from precision (63.00\% to 72.16\%), while sensitivity changed less (60.41\% to 62.33\%). Mixup is therefore treated here as a general regularizer rather than as a strategy for the minority class (Supplementary Table~S7).

\subsubsection{Balanced-Subset Evaluation versus Native Prevalence}\label{sec:balanced}

Many published cervical fracture classifiers are trained and tested on class-balanced subsets of the data (Section~\ref{sec:literature}). To measure how this protocol affects the reported results, Table~\ref{tab:balanced} compares the three settings of Section~\ref{sec:balanced-protocol} for the slice-stack model: the proposed protocol at native prevalence (A), balanced training tested at native prevalence (B), and the models of B tested on a balanced subset (C).

Comparing Settings B and C isolates the effect of the test set, because the models and their predictions are identical. Moving from the native prevalence to a balanced test set raised vertebra-level precision from 35.43\% to 82.64\% and AUPRC from 41.83\% to 80.61\%, while sensitivity was unchanged and ROC-AUC barely moved (79.79\% against 79.82\%). Because Setting C evaluates the same models as Setting B, these gains in Setting C come from the composition of the test set rather than from the model.

At the native prevalence, the balanced-training models of Setting B fell well below the proposed protocol, with a vertebra-level AUPRC of 41.83\% against 72.14\% and a precision of 35.43\% against 72.16\%, and a patient-level specificity of 52.16\% against 88.38\%. Models that appear strong on a balanced test set thus flag many more non-fractured vertebrae and patients as fractured once applied at the native prevalence, so results reported on balanced subsets overstate the performance such models reach on the full cohort.

\subsubsection{Operating Points}\label{sec:operating-point}

\begin{figure}[htbp]
  \centering
  \includegraphics[width=\linewidth]{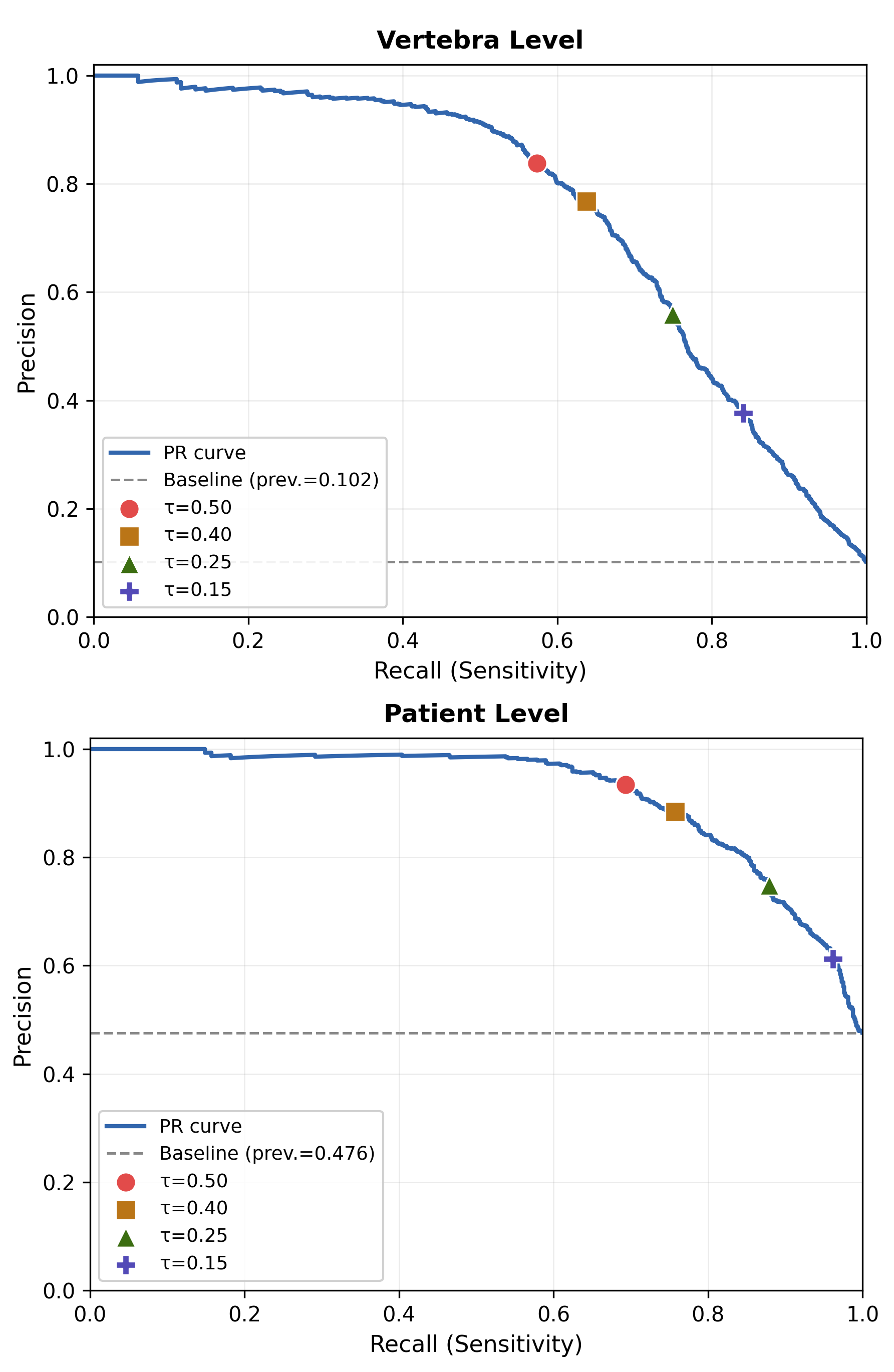}
  \caption{Precision-recall curves for the fusion model at the vertebra and patient levels. The dashed line in each panel is the random-chance baseline equal to the positive prevalence (0.102 vertebra, 0.476 patient). Marked points are the four thresholds of Table~\ref{tab:threshold_sweep}. Vertebra-level precision falls steeply as recall rises, whereas patient-level precision declines more gradually.}
  \label{fig:pr_curves}
\end{figure}

The vertebra-level and If-Any patient-level results above were obtained at a fixed decision threshold of 0.50. Under strong class imbalance, the threshold governs the trade-off between sensitivity and the other metrics, so we performed a threshold sweep on the fusion model. For each criterion, the threshold for each test fold was selected on the pooled predictions of the other four folds, using a grid with a step of 0.05. The selected threshold was identical in all five folds for every criterion, so each fold was evaluated at a threshold chosen without its own data. At the patient level, the If-Any rule was used throughout so that the effect of the threshold is separated from the aggregation rule. We considered the default threshold and the thresholds that maximize the F1, F2 and F3 scores, which weight sensitivity progressively more heavily than precision. Table~\ref{tab:threshold_sweep} reports the results, and Figure~\ref{fig:pr_curves} marks these operating points on the precision-recall curves.

Lowering the threshold from 0.50 to 0.15 raised sensitivity from 57.5\% to 84.0\% at the vertebra level and from 69.3\% to 96.1\% at the patient level, while precision and specificity declined. At the vertebra level, where prevalence is 10.2\%, precision fell from 84.2\% to 38.5\% over this range. At the patient level, where prevalence is 47.6\%, it fell less steeply, from 93.5\% to 61.6\%. The best-F1 threshold of 0.40 gave the highest F1-score at both levels (69.7\% per vertebra and 81.6\% per patient). The F2 threshold of 0.25 raised patient-level sensitivity to 87.9\% at a specificity of 73.1\%, and the F3 threshold of 0.15 raised it to 96.1\% at 44.6\%. The adaptive-average aggregation of Table~\ref{tab:patient_performance}, at 82.5\% sensitivity and 83.6\% specificity without a threshold search, lies between the F1 and F2 operating points.

Operating points also differ between models with similar discrimination. Between the two mask sources of Section~\ref{sec:mask-source}, threshold-dependent metrics at the 0.50 threshold moved in opposite directions (for example, vertebra-level sensitivity $-3.93$ and precision $+3.24$ points), which reflects where each model places its operating point rather than how well it discriminates. The clinical implications of the operating points are examined in Section~\ref{sec:disc-operating}.

\begin{table}[htbp]
  \centering
  \caption{Effect of the decision threshold on vertebra-level and patient-level (If-Any) performance of the fusion model (\%). Thresholds were selected by leave-one-fold-out; the selected threshold was identical in all five folds.}
  \label{tab:threshold_sweep}
  \resizebox{\linewidth}{!}{%
    \begin{tabular}{llccccc}
      \toprule
      Threshold & Level & Accuracy & Precision & Sensitivity & Specificity & F1-Score \\
      \midrule
      \multirow{2}{*}{Default (0.50)} & Vertebra & 94.5 & 84.2 & 57.5 & 98.7 & 68.1 \\
       & Patient & 83.1 & 93.5 & 69.3 & 95.5 & 79.6 \\
      \midrule
      \multirow{2}{*}{Best F1 (0.40)} & Vertebra & 94.3 & 76.9 & 63.9 & 97.8 & 69.7 \\
       & Patient & 83.8 & 88.5 & 75.8 & 91.0 & 81.6 \\
      \midrule
      \multirow{2}{*}{Best F2 (0.25)} & Vertebra & 91.4 & 56.3 & 74.9 & 93.3 & 64.1 \\
       & Patient & 80.2 & 75.0 & 87.9 & 73.1 & 80.8 \\
      \midrule
      \multirow{2}{*}{Best F3 (0.15)} & Vertebra & 84.2 & 38.5 & 84.0 & 84.2 & 52.5 \\
       & Patient & 69.2 & 61.6 & 96.1 & 44.6 & 75.0 \\
      \bottomrule
    \end{tabular}%
  }
\end{table}

\begin{table*}[!ht]
  \centering
  \small
  \caption{Fracture Recognition Performance Comparison with Existing Literature}
  \label{tab:literature_comparison}
  \setlength{\tabcolsep}{4pt}
  \begin{tabularx}{\textwidth}{@{} l >{\RaggedRight\hsize=1.2\hsize}X l l >{\RaggedRight\hsize=0.8\hsize}X @{}}
    \toprule
    \textbf{Study} & \textbf{Dataset} & \textbf{\makecell[l]{Vertebra Level\\Results}} & \textbf{\makecell[l]{Patient Level\\Results}} & \textbf{Major Drawback} \\
    \midrule
    \citep{Bhavya2022} & Test Set: 14 Patients & - & Acc: 83 & Very small test set \\
    \addlinespace[3pt]
    \citep{RasulAnalytics} & RSNA Subset (4200 images) \newline Test: 400 Images & Acc: 99.75 & - & Balanced before splitting \\
    \addlinespace[3pt]
    \citep{MahajanINCOFT} & Full RSNA Dataset & - & Acc: 76 & - \\
    \addlinespace[3pt]
    \citep{Chlad2023} & RSNA Subset \newline Test: 4330 slices & \makecell[l]{mAP: 98 \\ Frac Acc: 98} & - & Balanced before splitting \\
    \addlinespace[3pt]
    \citep{Yaseen2024} & RSNA Subset (14k slices) \newline Test: 2887 slices & \makecell[l]{Acc: 97.8 \\ F1: 97.8} & - & Balanced before splitting \\
    \addlinespace[3pt]
    \citep{Vel2026LightweightCNN} & RSNA Subset (balanced binary) \newline Test: 400 Images & \makecell[l]{Acc: 99.98 \\ Sen: 99.98} & - & Balanced before splitting \\
    \addlinespace[3pt]
    \citep{Golla2023} & Private (355 Patients) \newline Test: 127 Patients & Sensitivity: 87 & - & Dataset was small \\
    \addlinespace[3pt]
    \citep{Sutradhar2025} & RSNA Subset (235 Patients) \newline Test: 24 Patients & mAP 93 & - & Dataset was small \\
    \addlinespace[3pt]
    \citep{SalehinejadISBI} & Private (3666 Patients) \newline Test: 208 Patients & - & Acc: 70.92 & - \\
    \addlinespace[3pt]
    \citep{Kim2023PlatCon} & Full RSNA Public Dataset \newline Test: 404 Patients & Acc: 94.9 & - & - \\
    \addlinespace[3pt]
    \makecell[l]{RSNA 1st \citep{HaKaggle2022}\\(reproduced)} & Full RSNA Public Dataset \newline Test: 2019 Patients (CV) & \makecell[l]{Acc: 94.04, Pre: 73.01 \\ Sen: 66.85, Spe: 97.15 \\ F1: 69.68, ROC: 92.15 \\ AUPRC: 76.19} & \makecell[l]{Acc: 84.35, Pre: 88.06 \\ Sen: 77.75, Spe: 90.41 \\ F1: 82.52, ROC: 90.75 \\ AUPRC: 91.86} & - \\
    \midrule
    Proposed Method (slice-wise stack) & Full RSNA Public Dataset \newline Test: 2019 Patients (CV) & \makecell[l]{Acc: 93.55, Pre: 72.16 \\ Sen: 62.33, Spe: 97.09 \\ F1: 66.53, ROC: 90.20 \\ AUPRC: 72.14} & \makecell[l]{Acc: 82.67, Pre: 86.15 \\ Sen: 76.18, Spe: 88.38 \\ F1: 80.66, ROC: 89.09 \\ AUPRC: 90.42} & - \\
    \addlinespace[3pt]
    Proposed Method (Fusion) & Full RSNA Public Dataset \newline Test: 2019 Patients (CV) & \makecell[l]{Acc: 94.51, Pre: 84.16 \\ Sen: 57.49, Spe: 98.74 \\ F1: 68.15, ROC: 91.63 \\ AUPRC: 75.64} & \makecell[l]{Acc: 83.06, Pre: 82.12 \\ Sen: 82.51, Spe: 83.56 \\ F1: 82.28, ROC: 90.95 \\ AUPRC: 92.00} & - \\
    \bottomrule
  \end{tabularx}
\end{table*}

\begin{table*}[t]
  \centering
  \caption{Cross-stage error propagation on the 87-patient subset with ground-truth vertebra masks (means over the five folds). Spine localization is reported as VOI mIoU, segmentation as average Dice, and classification as F1-score and AUPRC at the vertebra and patient levels, with patient-level predictions from the adaptive-average aggregation. The training-matched ceiling uses a classifier pretrained on VOIs from the 3D-segmentation masks and fine-tuned on ground-truth-mask VOIs. All other rows use the pipeline classifier, trained on VOIs from projection-derived masks.}
  \label{tab:error_prop}
  \footnotesize
  \setlength{\tabcolsep}{6pt}
  \begin{tabular}{lcccccc}
    \toprule
    \multirow{2}{*}{Pipeline configuration} & VOI & Seg. & \multicolumn{2}{c}{Vertebra level} & \multicolumn{2}{c}{Patient level} \\
    \cmidrule(lr){4-5} \cmidrule(lr){6-7}
     & mIoU & Avg Dice & F1 & AUPRC & F1 & AUPRC \\
    \midrule
    GT ROI + GT 3D mask (training-matched ceiling) & 100 & 100.00 & 70.2 & 78.5 & 77.3 & 90.0 \\
    \midrule
    GT ROI + projection-derived mask & 100 & 87.86 & 58.2 & 77.1 & 75.8 & 93.4 \\
    Predicted ROI + projection-derived mask & 94.45 & 86.21 & 54.3 & 74.3 & 76.3 & 91.7 \\
    \midrule
    10\% perturbed GT ROI + projection-derived mask & 82.59 & 66.24 & 49.2 & 63.3 & 65.0 & 89.8 \\
    20\% perturbed GT ROI + projection-derived mask & 68.92 & 50.70 & 38.5 & 46.0 & 60.8 & 85.6 \\
    30\% perturbed GT ROI + projection-derived mask & 57.90 & 41.41 & 25.9 & 37.6 & 57.4 & 84.6 \\
    \bottomrule
  \end{tabular}
\end{table*}

\subsubsection{Comparison with Published Methods}\label{sec:literature}

Table~\ref{tab:literature_comparison} places the proposed pipeline among published cervical fracture classifiers. Direct comparison is limited by differences in data and evaluation, since several studies used small private datasets and others trained and tested on class-balanced subsets of the public RSNA dataset, which inflate precision and AUPRC relative to the native prevalence (Section~\ref{sec:balanced}). The most comparable results are therefore those obtained on the full RSNA 2022 dataset at its native prevalence.

The first-place solution of the RSNA challenge segments the vertebrae voxel-wise with a 3D ResNet18d-Unet, classifies each vertebra with a CNN-LSTM, and combined nearly 20 such models in its final submission \citep{HaKaggle2022}. For a comparison under an identical protocol, we reproduced its publicly released single-model implementation and evaluated it with the same five-fold cross-validation on the full dataset. At the 0.50 threshold, the fusion model reached a vertebra-level F1-score of 68.15\% and accuracy of 94.51\%, against 69.68\% and 94.04\% for the reproduced solution, and a patient-level F1-score of 82.28\% and accuracy of 83.06\%, against 82.52\% and 84.35\%. These values depend on each model's operating point. The fusion model favors precision at this threshold, with a vertebra-level precision of 84.16\% at a sensitivity of 57.49\%, against 73.01\% at 66.85\% for the reproduced solution. AUPRC, which does not depend on the threshold, was 75.64\% against 76.19\% per vertebra and 92.00\% against 91.86\% per patient, and the paired statistics for this comparison are reported in Section~\ref{sec:mask-source}.

The challenge organizers later reported aggregate results for the highest-scoring submissions on the private test set \citep{lee2024performance}. For the top eight algorithms, they reported a patient-level AUC of 0.96 and F1-score of 90\%, with a sensitivity of 88\% and a specificity of 94\%. Vertebra-level identification remained harder, and from the organizers' vertebra-level F1 box plots we estimate a median F1-score of about 67.57\% across the top submissions. These results were obtained with level-specific and patient-level threshold optimization and large ensembles, on a test set that is not publicly released and with ranking by weighted log loss rather than F1, so they serve as an approximate reference rather than a directly comparable baseline (Section~\ref{sec:disc-comparison}).

\subsection{Error Sources and Propagation}\label{sec:error-propagation}

As the pipeline is sequential, localization and segmentation errors can propagate into the final fracture decision. To quantify this, we traced how degradation at the detection stage affects segmentation and, in turn, classification, on the 87-patient subset for which ground-truth voxel-level masks are available. For each patient, the detection, segmentation and classification weights used at inference were taken from the fold in which that patient was held out, so that no patient was evaluated by a model that had seen it during training. The conditions in Table~\ref{tab:error_prop} differ in two controlled variables, the source of the spine ROI and the source of the segmentation mask. In all rows except the first, the mask is predicted by the proposed segmentation model from the cut ROI, so that moving down the table changes only ROI quality, first by replacing the ground-truth ROI with the predicted YOLOv8x ROI and then by perturbing the ground-truth ROI by 10\%, 20\% and 30\% before segmentation. The first row uses the ground-truth 3D masks directly, which fixes segmentation at a Dice of 100\%, and serves as a training-matched ceiling.

The training-matched ceiling (Section~\ref{sec:gt-ceiling}) reached a vertebra-level AUPRC of 78.5\%, the highest of any configuration, against 77.1\% with the ground-truth ROI and projection-derived masks and 74.3\% for the full pipeline. On a subset of 87 patients these differences are point estimates, interpreted in Section~\ref{sec:disc-viability}. At the patient level, the ceiling reached an AUPRC of 90.0\%, below both of these configurations. Replacing the ground-truth ROI with the predicted ROI lowered VOI mIoU to 94.45\% and segmentation Dice from 87.86\% to 86.21\%. The downstream effect was correspondingly small. Vertebra-level AUPRC moved from 77.1\% to 74.3\% and patient-level AUPRC from 93.4\% to 91.7\%, with patient-level F1 essentially unchanged (75.8\% and 76.3\%) and vertebra-level F1 lower by 3.9 points. The learned detector therefore operates close to the perfect-localization configuration.

\begin{figure}[htbp]
  \centering
  \includegraphics[width=0.8\linewidth]{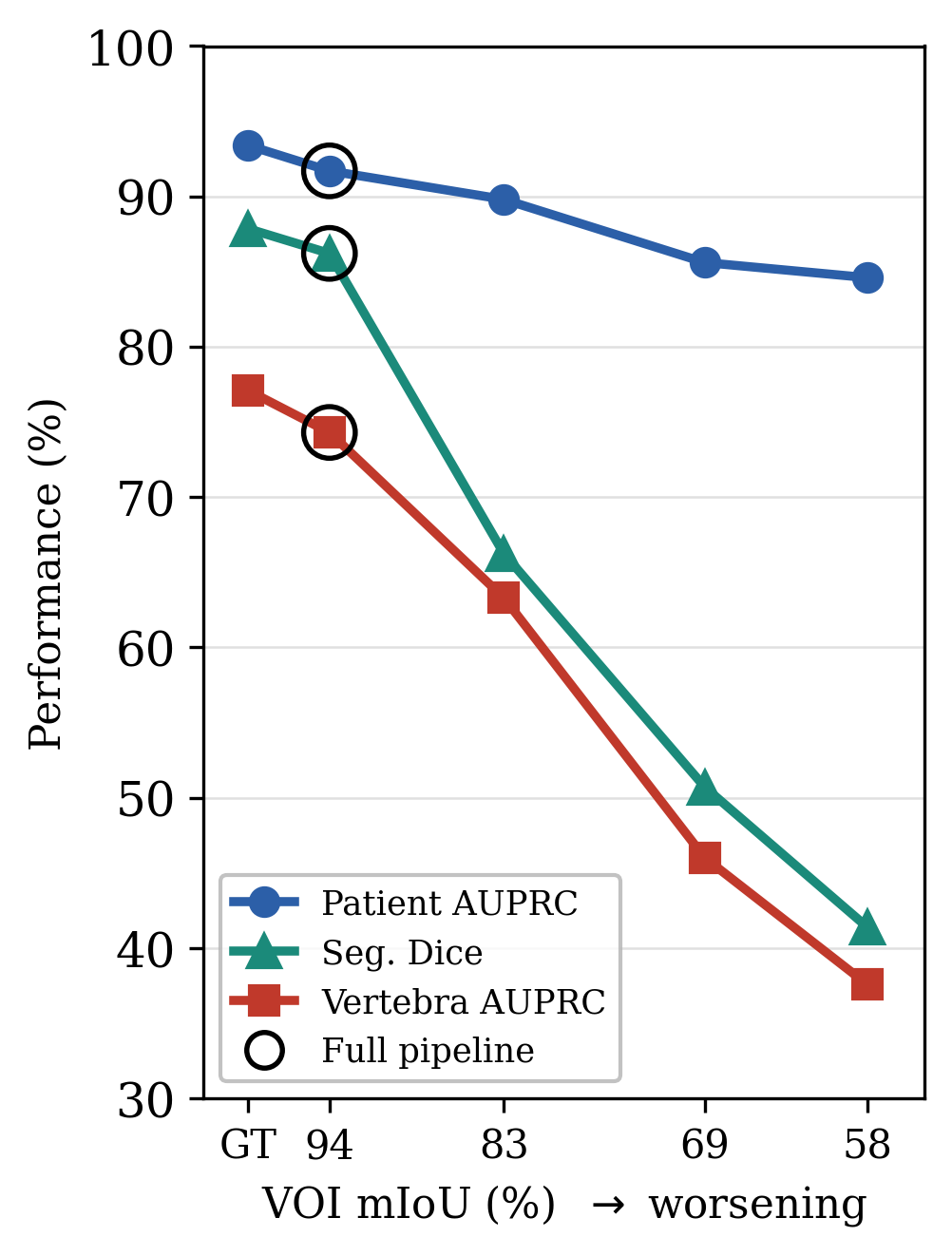}
  \caption{Cross-stage error propagation on the 87-patient subset as the spine ROI moves from ground truth to the predicted ROI and to 10\%, 20\% and 30\% perturbation.}
  \label{fig:error_prop_cascade}
\end{figure}

\begin{figure*}[t]
  \centering
  \includegraphics[width=0.8\textwidth]{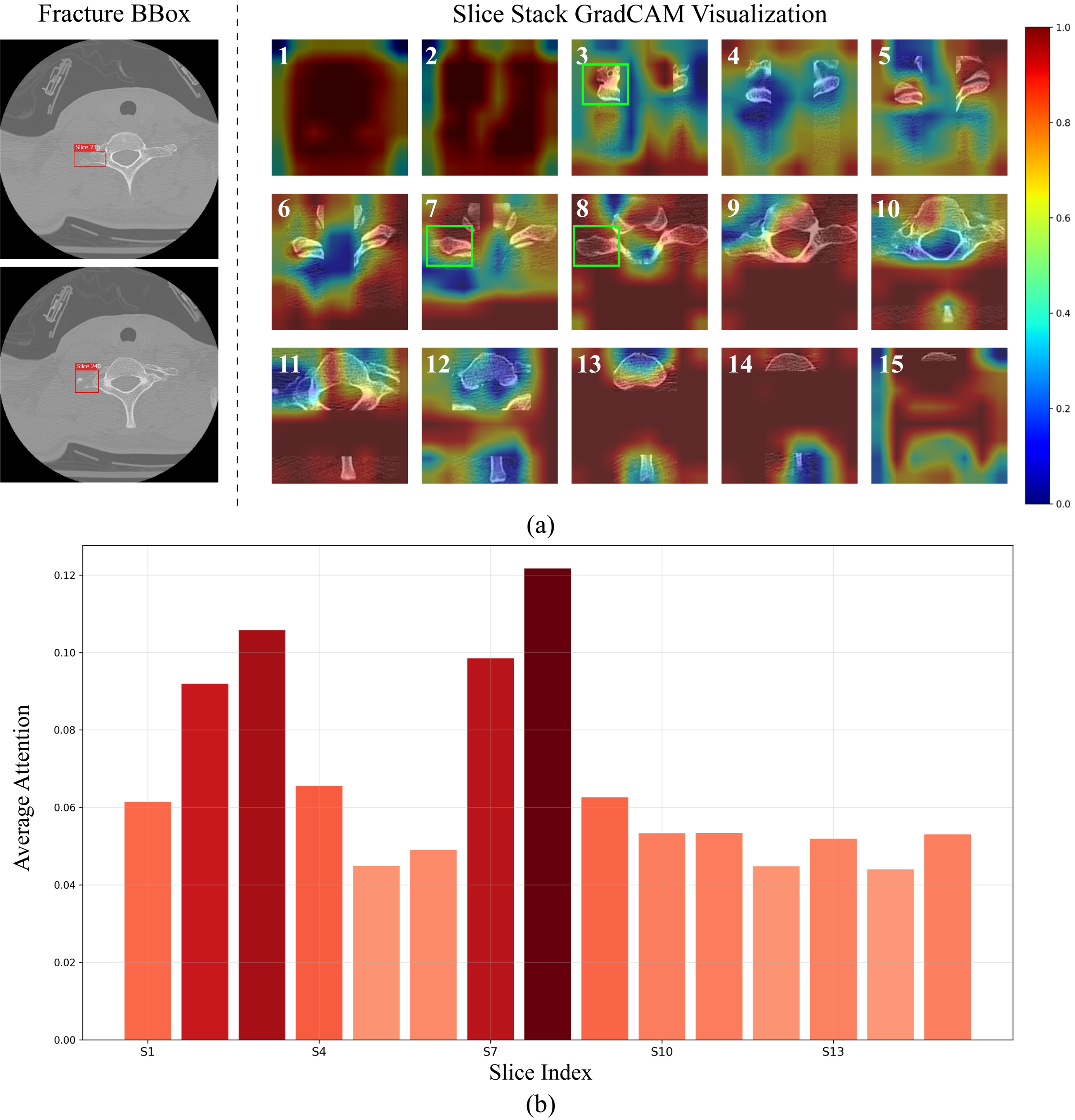}
  \caption{Explainability for a correctly classified fractured vertebra. (a) Grad-CAM maps over the 15 slice stacks, with the annotated fracture outlined in green. (b) Average transformer attention per slice stack.}
  \label{fig:explainability}
\end{figure*}

Figure~\ref{fig:error_prop_cascade} traces how ROI perturbation propagates as localization quality decreases. Segmentation Dice fell steeply, to 66.24\%, 50.70\% and 41.41\% at the three perturbation levels. Vertebra-level AUPRC fell roughly in step, from 74.3\% to 63.3\%, 46.0\% and 37.6\%, whereas patient-level AUPRC fell only from 91.7\% to 84.6\%, even when VOI mIoU dropped to 57.90\%. Upstream quality therefore matters, and its effect on fracture recognition is gradual rather than abrupt.

Threshold-dependent metrics can move differently from AUPRC. Under the predicted ROI, patient-level F1 rose slightly, from 75.8\% to 76.3\%, while patient-level AUPRC fell from 93.4\% to 91.7\%, because F1 is read at a fixed operating point whose precision and sensitivity can shift in opposite directions. We therefore treat AUPRC as the indicator of propagation. Within the limits of the 87-patient subset, the pipeline tolerates the localization error of the real detector with little downstream loss, and it degrades gradually under larger ROI perturbation, more steeply at the vertebra level than at the patient level. The implications for the viability of projection-driven approximation are discussed in Section~\ref{sec:disc-viability}.

\subsection{Explainability and Reader Agreement}\label{sec:decision-evidence}

This section examines whether the classifier's decisions rest on fracture evidence within the projection-derived VOIs, and how its agreement with the reference standard compares with that of radiologists.

\subsubsection{Explainability}\label{sec:explainability}

\begin{figure}[htbp]
  \centering
  \includegraphics[width=\linewidth]{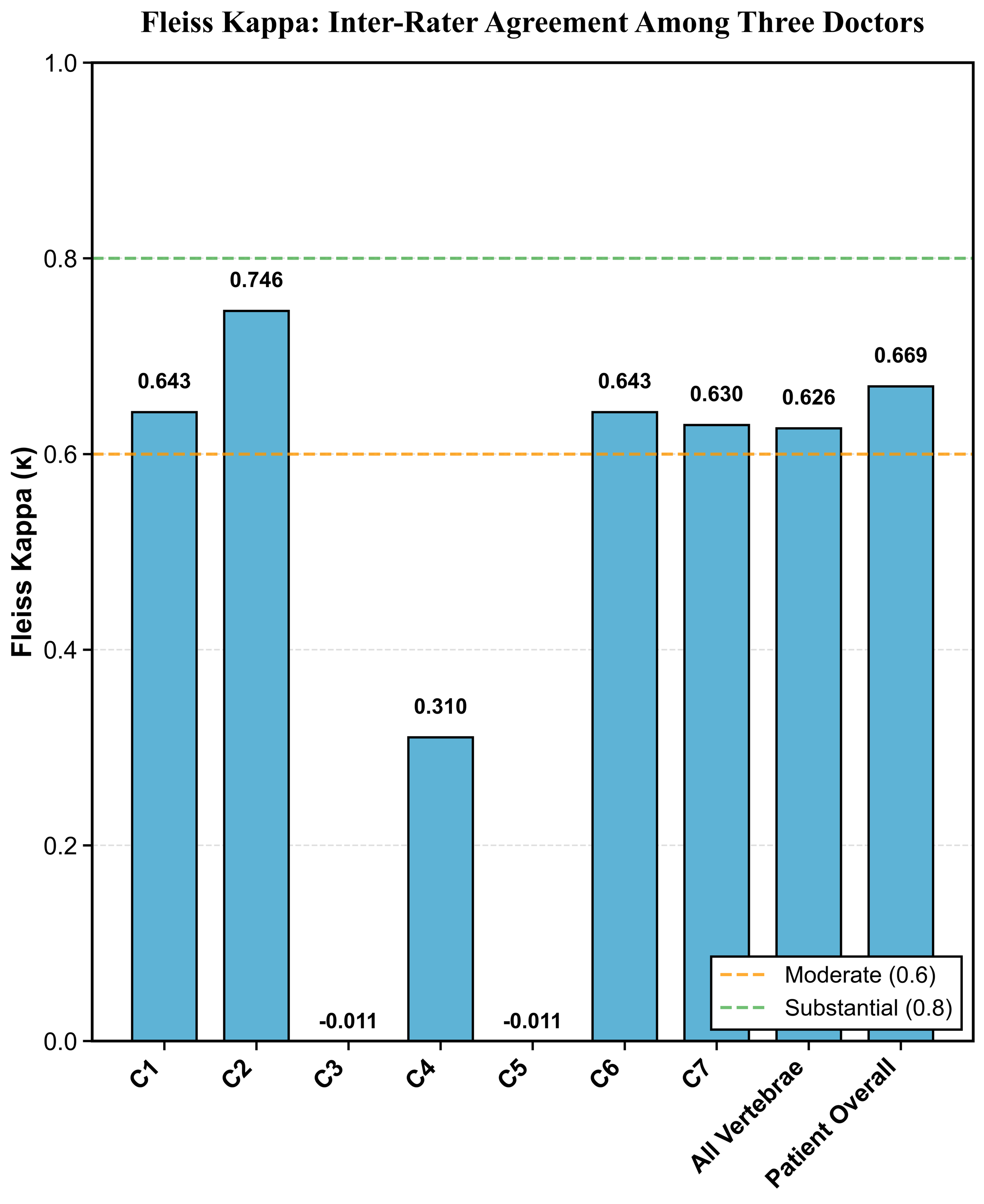}
  \caption{Fleiss' $\kappa$ among the three radiologists and the reference standard at the patient level, across all vertebrae and for each vertebra (30 patients).}
  \label{fig:interobserver_fleiss}
\end{figure}

Figure~\ref{fig:explainability} demonstrates the explainability analysis for a representative vertebra that was correctly classified as fractured by the proposed 2.5D CNN-Transformer model. The visualization illustrates both the spatial and the sequential aspects of the model's decision-making process. In Figure~\ref{fig:explainability}(a), the Grad-CAM analysis shows the spatial attention patterns across all 15 slice stacks, with the heat maps overlaid on the original CT slices using a color scale in which red indicates the regions of highest model attention. The annotated fracture location is marked by green bounding boxes on the reference slices. Figure~\ref{fig:explainability}(b) presents the corresponding attention weights extracted from the transformer component, quantifying how much each slice stack contributes to the final fracture prediction. The model predominantly focused on slice stacks 3, 7 and 8, which received the three highest average attention scores in the sequence. Notably, the spatial attention patterns in these high-weighted stacks include the anatomical region where the annotated fracture is located. This alignment between the model's spatial focus and the annotated fracture location suggests that the CNN-Transformer attends to anatomically relevant regions within the projection-derived VOI.

\subsubsection{Agreement with Radiologists}\label{sec:interobserver}

On the 30-patient subset of Section~\ref{sec:interobserver-protocol}, we compared the three radiologists with one another, and the radiologists and the model with the reference standard of the dataset.

\paragraph{Inter-Rater Agreement Among Experts}

The analysis of consensus among the four raters, the three radiologists and the reference standard, revealed a substantial but imperfect level of agreement. As illustrated in Figure~\ref{fig:interobserver_fleiss}, the overall agreement at the patient level was substantial, with a Fleiss' $\kappa$ ($\kappa_F$) of 0.669, and a similar level of consensus was found when all vertebrae were considered collectively ($\kappa_F = 0.626$) \citep{LandisKoch1977}. The vertebra-specific analysis showed clear differences in consistency across the cervical spine. The highest consensus was observed for the C2 vertebra ($\kappa_F = 0.746$), likely attributable to distinct anatomical landmarks such as the dens, which make fractures more consistently identifiable. The C1, C6 and C7 vertebrae also showed substantial agreement, with $\kappa_F$ values of 0.643, 0.643 and 0.630, respectively. Agreement was much lower for the mid-cervical vertebrae. C4 showed only fair agreement ($\kappa_F = 0.310$), while C3 and C5 yielded $\kappa_F$ values of $-0.011$, indicating no agreement beyond chance. This suggests that these levels are harder to assess consistently, although the small number of fractures at these levels in the chosen subset also contributes.

To further examine the sources of this variability, we analyzed pairwise agreement using Cohen's $\kappa$ (Supplementary Section S11). The radiologists often had higher agreement with each other than with the reference standard, and there were clear differences in individual diagnostic patterns. At the patient level, Radiologist 1 showed the highest concordance with the reference standard, with substantial agreement ($\kappa = 0.67$), whereas Radiologist 3 showed the lowest, with a $\kappa$ of 0.40 indicating only fair agreement. The agreement between Radiologists 1 and 2 was substantial ($\kappa = 0.80$), as was the agreement between Radiologists 2 and 3 ($\kappa = 0.70$). The agreement between Radiologists 1 and 3 was weaker ($\kappa = 0.52$), suggesting that Radiologist 3 diverged more from the diagnostic pattern of Radiologist 1. This trend was more pronounced at the vertebra level, where agreement with the reference standard dropped for all radiologists. Radiologist 2 had the highest agreement with the reference standard at this level, though it was only moderate ($\kappa = 0.46$), and Radiologist 1 the lowest ($\kappa = 0.37$). Radiologist 3 again showed only moderate agreement with Radiologist 1 ($\kappa = 0.51$) and Radiologist 2 ($\kappa = 0.59$), while the substantial agreement between Radiologists 1 and 2 persisted ($\kappa = 0.76$). This pattern across both levels suggests differences between the expert annotations and the dataset reference standard, as well as reader-specific variability in diagnostic criteria.

\paragraph{Model Agreement Relative to Expert Consensus}

\begin{figure}[htbp]
  \centering
  \includegraphics[width=0.9\linewidth]{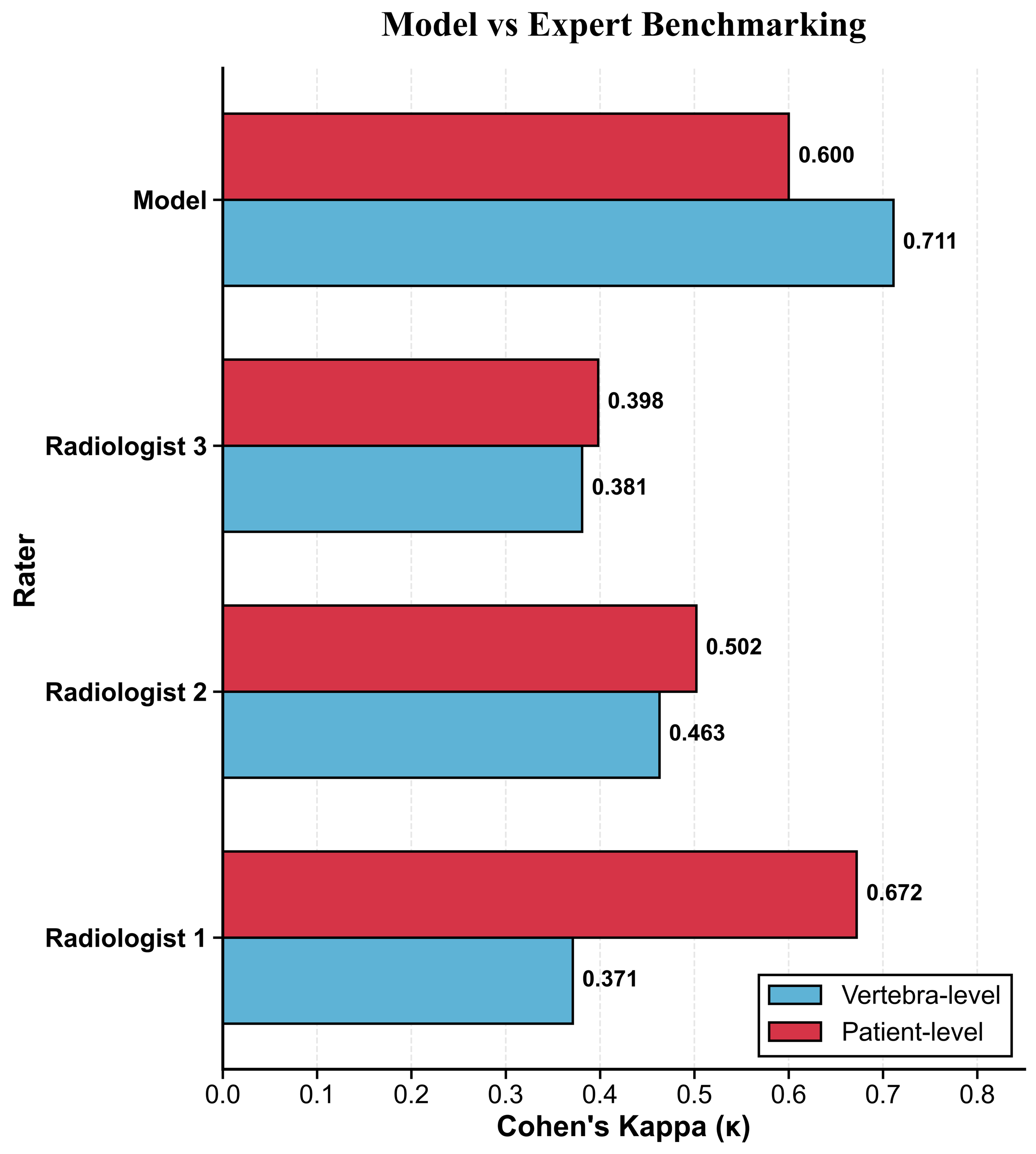}
  \caption{Cohen's $\kappa$ with the reference standard for the model and for each radiologist at the patient and vertebra levels (30 patients).}
  \label{fig:model_doctor_bar_chart}
\end{figure}

We evaluated the model against the reference standard alongside the three radiologists. As summarized in Figure~\ref{fig:model_doctor_bar_chart}, the model's agreement with the reference standard at the vertebra level was substantial ($\kappa = 0.711$) and exceeded that of each radiologist on this subset, whose agreement ranged from 0.371 to 0.463. The model also identified 19 of the 29 fractures in the subset, compared with 10, 11 and 8 for the three radiologists. At the patient level, the model's agreement ($\kappa = 0.600$) fell between those of the least- and best-agreeing radiologists ($\kappa = 0.40$ and $0.67$), and it identified 14 of the 19 fractured patients, as many as Radiologist 1 and more than the other two. Detailed confusion matrices are given in Supplementary Section S11. This comparison favors the model to some extent. The reference labels were derived from the radiology reports written at the contributing institutions (Section~\ref{dataset-description}) \citep{Lin2023RSNA}, and the model was trained on these labels, whereas the three radiologists annotated the scans independently of those reports. The model was therefore trained to match the reference standard, which partly explains why its agreement with the reference exceeded that of the radiologists, who in turn agreed more closely with one another than with the reference. The implications are discussed in Section~\ref{sec:disc-patterns}.

\begin{table*}[t]
  \centering
  \caption{Comparison of segmentation architectures on the energy projection (five-fold cross-validation, 87 patients). Score is the composite of Equation~\ref{eq:score_metric}.}
  \label{tab:segmentation_comparison}
  \resizebox{\linewidth}{!}{%
    \begin{tabular}{lcccccccccc}
    \toprule
    \multirow{2}{*}{Model} & \multicolumn{3}{c}{Sagittal} & \multicolumn{3}{c}{Coronal} & \multicolumn{4}{c}{Average} \\
    \cmidrule(lr){2-4} \cmidrule(lr){5-7} \cmidrule(lr){8-11}
     & IoU$\uparrow$ & DSC$\uparrow$ & HD95$\downarrow$ & IoU$\uparrow$ & DSC$\uparrow$ & HD95$\downarrow$ & IoU$\uparrow$ & DSC$\uparrow$ & HD95$\downarrow$ & Score$\uparrow$ \\
    \midrule
    SE\_ResNext101\_UperNet & 74.87 & 85.60 & 4.42 & 77.22 & 87.12 & 4.28 & 76.05 & 86.36 & 4.35 & 1.044 \\
    EfficientNetv2\_Segformer & 74.06 & 85.05 & 4.55 & 77.84 & 87.52 & 4.09 & 75.95 & 86.29 & 4.32 & 1.048 \\
    EfficientNetv2\_DeepLabV3+ & 73.96 & 85.00 & 4.61 & 77.99 & 87.61 & 4.06 & 75.97 & 86.30 & 4.33 & 1.046 \\
    DenseNet169\_Unet++ & 73.65 & 84.79 & 4.70 & 77.50 & 87.30 & 4.20 & 75.58 & 86.05 & 4.45 & 1.011 \\
    EfficientNet-b2\_MAnet & 73.86 & 84.93 & 4.52 & 76.95 & 86.95 & 4.49 & 75.41 & 85.94 & 4.50 & 0.997 \\
    MobileNetv2\_Linknet & 73.61 & 84.76 & 4.82 & 77.27 & 87.15 & 4.29 & 75.44 & 85.96 & 4.56 & 0.984 \\
    MiT\_b0\_SegFormer & 74.10 & 85.10 & 4.56 & 76.40 & 86.60 & 4.56 & 75.25 & 85.85 & 4.56 & 0.981 \\
    EfficientNet-b2\_FPN & 72.65 & 84.11 & 4.96 & 77.32 & 87.19 & 4.30 & 74.98 & 85.65 & 4.63 & 0.959 \\
    DenseNet169\_Unet & 73.22 & 84.50 & 4.79 & 75.71 & 86.15 & 4.87 & 74.46 & 85.33 & 4.83 & 0.904 \\
    \textbf{DenseNet121\_Unet (Proposed)} & \textbf{78.53} & \textbf{87.93} & \textbf{4.17} & \textbf{78.36} & \textbf{87.79} & 4.83 & \textbf{78.45} & \textbf{87.86} & 4.50 & \textbf{1.060} \\
    \bottomrule
    \end{tabular}%
  }
\end{table*}

\subsection{Component Selection}\label{sec:component-selection}

This section reports the comparisons used to select the spine detector, the segmentation network and the configuration of the classification network. The projection operators for segmentation are compared in Section~\ref{sec:projection-choice}, and the 2D, 3D and 2.5D classifier families in Section~\ref{sec:recognition-levels}.

\subsubsection{Spine Localization}\label{sec:select-detection}

ROI detection of the cervical spine was evaluated using five-fold cross-validation across the 87 patients with 3D vertebra masks, with 3D mIoU as the primary metric for localization accuracy. We assessed multiple size variants of the YOLO (v5, v8, v9, v10) and RT-DETR architectures on the maximum, mean, gradient and variance projections.

\begin{figure}[htbp]
  \centering
  \includegraphics[width=\linewidth]{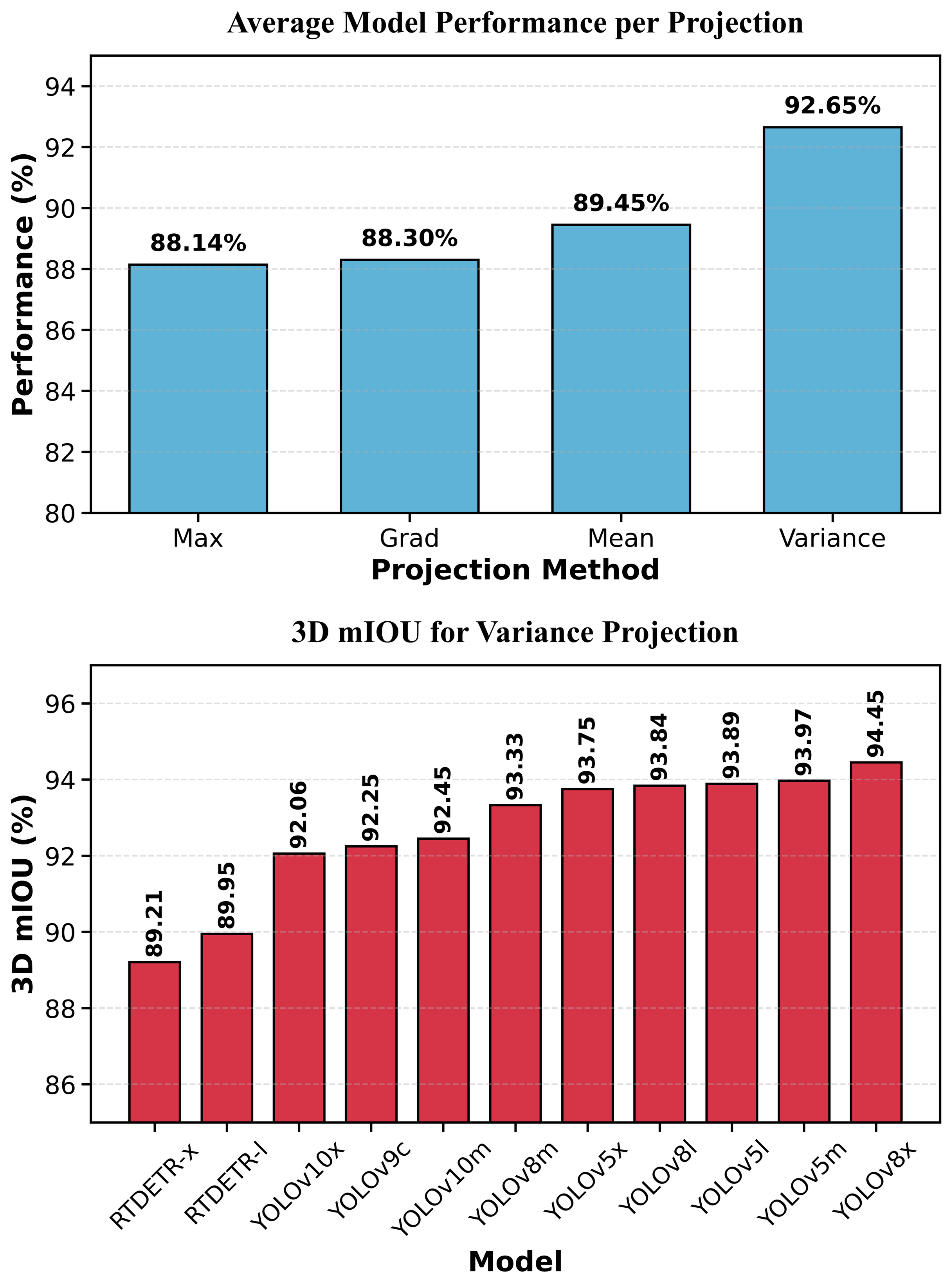}
  \caption{Spine localization performance (3D mIoU) of the evaluated detectors. Top, average over all detectors for each projection. Bottom, each detector on the variance projection.}
  \label{fig:roi_results}
\end{figure}

Among the projection strategies, the variance projection was the most effective. As illustrated in Figure~\ref{fig:roi_results} (top), it achieved the highest average performance across all models (92.65\%), compared with the mean (89.45\%), gradient (88.30\%) and maximum (88.14\%) projections. The variance projection highlights inter-slice intensity fluctuations, providing a more stable representation of the vertebral column than the maximum intensity projection, which can obscure fine details by emphasizing high-intensity regions. On the variance projection, YOLOv8x achieved the highest 3D mIoU, 94.45\% (Figure~\ref{fig:roi_results}, bottom). The preprocessing ablation showed that volume interpolation increased the 3D mIoU of YOLOv8x from 93.46\% to 94.45\%. Supplementary Figure S15 shows consistent spine localization by the top-performing v8 and v5 detectors. Based on these findings, the predictions of YOLOv8x on the variance projection were used to extract the cervical spine VOI from all 2019 patient scans for the subsequent stages. Detailed results for all projections and models are provided in Supplementary Section S5.

\subsubsection{Segmentation Network}\label{sec:select-segmentation}

\begin{table*}[htbp]
  \centering
  \caption{Ablation of the sequential modeling component of the slice-stack model (vertebra level, \%).}
  \label{tab:ablation_sequential}
  \footnotesize
  \begin{tabular*}{\textwidth}{@{\extracolsep{\fill}}lccccccc}
    \toprule
    Sequential Component & Layers & Accuracy & Precision & Sensitivity & Specificity & F1-Score & ROC-AUC \\
    \midrule
    LSTM & 2 & \textbf{93.70} & \textbf{75.36} & 58.67 & \textbf{97.73} & 65.67 & 89.83 \\
    Transformer & 1 & 93.11 & 68.92 & 60.35 & 96.85 & 64.00 & 89.12 \\
    Transformer & 2 & 93.55 & 72.16 & \textbf{62.33} & 97.09 & \textbf{66.53} & \textbf{90.20} \\
    Transformer & 3 & 93.22 & 71.37 & 58.86 & 97.17 & 63.90 & 88.95 \\
    \bottomrule
  \end{tabular*}
\end{table*}

Table~\ref{tab:segmentation_comparison} compares the segmentation architectures on the sagittal and coronal projections, together with their average performance. All models were trained on the energy projection selected in Section~\ref{sec:projection-choice}. DenseNet121-Unet achieved the highest average IoU (78.45\%) and DSC (87.86\%) and the highest overall score (1.060). In the sagittal view, it achieved an IoU of 78.53\%, a DSC of 87.93\% and an HD95 of 4.17\,mm, the best values of all models in this view. In the coronal view, it obtained the highest IoU (78.36\%) and DSC (87.79\%), although its coronal HD95 of 4.83\,mm was not the lowest, and EfficientNetv2-DeepLabV3+ reached 4.06\,mm in this view. EfficientNetv2-Segformer and EfficientNetv2-DeepLabV3+ also performed well, with scores of 1.048 and 1.046, particularly on coronal HD95. DenseNet121-Unet provided the highest IoU and Dice values in both views and a moderate HD95, which together gave the best overall average performance.

In the qualitative examples of Supplementary Figure S17, DenseNet121-Unet produces more compact and anatomically plausible masks than the comparison models, especially around overlapping vertebral structures. Because the downstream task relies on accurate vertebra-centered cropping rather than exact boundary placement, DenseNet121-Unet was used for vertebra mask estimation in all subsequent experiments.

\subsubsection{Classification Network}\label{sec:select-classification}

To configure the classification network, we conducted ablation studies on the sequential modeling component and on two training choices. As summarized in Table~\ref{tab:ablation_sequential}, the CNN-Transformer with two encoder layers achieved the highest F1-score (66.53\%) and ROC-AUC (90.20\%) among the LSTM and the Transformer depths compared, while the LSTM gave the highest accuracy, precision and specificity. Increasing the depth to three layers lowered both F1-score and ROC-AUC, likely reflecting overfitting, so the two-layer configuration was adopted. We also examined square padding of the extracted vertebra VOIs before classification. The configuration without square padding performed better, raising the F1-score from 61.52\% to 66.53\% and the ROC-AUC from 89.10\% to 90.20\%, with a corresponding gain in sensitivity. A likely explanation is that square padding reduces the effective resolution of the vertebra within the fixed-size input, since the padded border consumes pixels that would otherwise carry anatomical detail. The proposed configuration therefore omits square padding (Supplementary Table~S8).

\section{Discussion}\label{sec:discussion}

\subsection{Viability of Projection-Driven Approximation}\label{sec:disc-viability}

The central question of this study was whether vertebra masks approximated from a small number of optimized 2D projections retain enough anatomical and fracture-relevant information to support downstream cervical fracture recognition. We addressed it with two comparisons, one against 3D-segmentation masks under the same classifier and one against the reproduced RSNA 2022 winning solution. In the matched comparison of Section~\ref{sec:mask-source}, replacing the 3D-segmentation masks with projection-derived masks under the same classifier, training protocol and folds changed AUPRC by about one point or less in every configuration, and the fold-wise 95\% CI did not support a vertebra-level AUPRC loss larger than 1.46 points. Against the reproduced RSNA 2022 winning solution \citep{HaKaggle2022}, the corresponding bound for the whole pipeline was 2.81 points. The main finding of this study is therefore that projection-domain mask approximation, in which each vertebral mask is fused from two orthogonal projection segmentations rather than delineated voxel by voxel, is a viable substitute for full-resolution 3D segmentation in this pipeline.

Neither comparison is sufficient on its own, but each covers the main weakness of the other. The matched comparison isolates the mask source, but our classifier was developed and selected on projection-derived VOIs, so the 3D-segmentation masks are evaluated with a classifier tuned for a different input. This asymmetry favors the projection-derived masks and is the most likely explanation for the positive ROC-AUC differences in Section~\ref{sec:mask-source}. The comparison with the reproduced winning solution removes the asymmetry, since each pipeline uses a classifier developed for its own VOIs, but it changes architecture and training together with the masks. Its ROC-AUC differences are small and opposite in sign ($-0.52$ points at the vertebra level and $+0.19$ at the patient level), with 95\% CIs that include zero, so the positive shift seen in the matched comparison does not recur once the selection asymmetry is removed. Because the vertebra-level AUPRC loss stays bounded in both settings, at 1.46 points in the matched comparison and 2.81 points against the winning solution (lower limits of the fold-wise 95\% CIs), the small cost of the projection-derived masks is unlikely to be an artifact of the classifier having been tuned to them.

The reason such approximate masks suffice lies in how the segmentation stage is used. The estimated masks are not themselves the diagnostic output. They delineate the vertebra-centered volume that the classifier then examines, so the downstream requirement is accurate localization of each vertebra rather than precise delineation of its surface. On the 87 patients with voxel-level masks, the training-matched ceiling reached a vertebra-level AUPRC of 78.5\%, against 77.1\% with ground-truth ROIs and projection-derived masks and 74.3\% for the full pipeline (Section~\ref{sec:error-propagation}). The ceiling was thus 1.4 points above the projection-derived masks, and the learned detector lowered AUPRC by a further 2.8 points. The 1.4-point difference reflects the pretraining of the ceiling classifier on VOIs from 3D-segmentation masks as well as mask approximation, and both differences are small for a subset of this size. At the patient level the ceiling was not the highest of the three configurations, so the subset cannot rank the two sources of error. The perturbation experiment is more informative. A 30\% ROI perturbation, which lowers segmentation Dice to 41.41\%, reduces vertebra-level AUPRC from 77.1\% to 37.6\% and patient-level AUPRC only from 93.4\% to 84.6\%. Upstream errors therefore matter in proportion to their size, and those of the actual pipeline are small by comparison. The patient-level decision is more tolerant because it aggregates over all seven vertebrae, whereas a vertebra-level decision depends on where each volume is cropped. This graded dependence is why the localization and segmentation stages were refined rather than left to the classifier to absorb.

Projection-domain approximation is lossy at the level of the mask, since a voxel is assigned to a vertebra only where both back-projected views agree, and the crop boundaries it defines are therefore less precise than those of voxel-level segmentation. The image content inside each crop is unaffected, because the classifier reads the full-resolution CT within the approximated volume. The two models of the score-fusion ensemble read each volume differently, from conventional slice stacks and from maximum-projection stacks, and fusing them raised vertebra-level AUPRC from 72.14\% and 71.89\% to 75.64\%, and patient-level AUPRC from 90.42\% and 90.58\% to 92.00\%. The fused AUPRC lies above both members, and fusion added the same 3.50 vertebra-level AUPRC points when the 3D-segmentation masks were used (Section~\ref{sec:recognition-levels}), so the gain comes from combining the two representations, whichever masks define the volume. The fusion also raises vertebra-level precision from 72.16\% to 84.16\%, although under the fixed default threshold it lowers sensitivity from 62.33\% to 57.49\%, a threshold-dependent behavior examined in Section~\ref{sec:disc-operating}.

Accurate vertebra-level and patient-level fracture recognition is therefore achievable from projection-approximated vertebra volumes without full-resolution 3D segmentation. With the classifier architecture fixed, vertebra-level AUPRC changed by about one point or less between mask sources, and losses larger than 1.46 points were not supported by the data. On the 87 voxel-annotated patients, the training-matched ceiling was about 1.4 AUPRC points higher.

\subsection{Positioning Against 3D Baselines and the Broader Field}\label{sec:disc-comparison}

The first-place RSNA 2022 solution \citep{HaKaggle2022} is the most informative external reference for this work, because it shares our task and dataset and differs most visibly in the component under study. It localizes and segments the vertebrae voxel-wise with a 3D ResNet18d-Unet and classifies each vertebra with a CNN-LSTM, whereas our pipeline approximates the vertebra masks in the projection domain and classifies with a two-model 2.5D fusion. Reproducing its single-model implementation under our five-fold protocol allowed the two designs to be compared on the full imbalanced cohort rather than through balanced-subset results. This comparison complements the matched experiment of Section~\ref{sec:mask-source}, as discussed in Section~\ref{sec:disc-viability}. It shows that a pipeline built on projection-derived masks reaches discrimination close to that of a well-engineered 3D-segmentation pipeline, with no significant difference in AUPRC or ROC-AUC at either level.

The same full-cohort protocol explains why our results are not directly comparable with the accuracies above 90\% reported on balanced subsets of this dataset (Section~\ref{sec:literature}). When the same trained models were scored on a balanced test set instead of the full cohort, vertebra-level precision rose from 35.43\% to 82.64\% and AUPRC from 41.83\% to 80.61\%, while sensitivity was unchanged and ROC-AUC barely moved (Section~\ref{sec:balanced}). Much of the gap in precision, F1-score and AUPRC between balanced-subset and full-cohort results therefore reflects the composition of the test set rather than the classifier. Many of those studies also classify individual slices or images rather than every vertebra of every patient (Supplementary Table~S1), which is a different unit of prediction from the vertebra-level decisions reported here.

A second, looser reference is the aggregate performance that the challenge organizers reported for the highest-scoring submissions \citep{lee2024performance}. These figures were obtained on a private test set with large ensembles and level-specific threshold optimization, so they are best read as an approximate ceiling rather than a comparable baseline (Section~\ref{sec:literature}). Their most informative aspect is the gap between diagnostic levels. The top submissions reached a patient-level AUC of 0.96, whereas their vertebra-level F1-score had an estimated median of about 67.57\%, which suggests that vertebra-level identification is intrinsically harder than patient-level diagnosis rather than a problem left unsolved only by smaller models. The vertebra-level F1-score of 68.15\% reached here under five-fold cross-validation on the full public cohort places the projection-based pipeline in the same performance band as the strongest challenge ensembles, despite operating in a reduced-dimensionality projection domain and without their scale or threshold optimization. The comparison is not head-to-head, but it indicates that the projection-driven approximation does not forfeit discriminative capability relative to the level of the literature, reinforcing the viability argument of Section~\ref{sec:disc-viability} against the broader field rather than a single reproduced baseline.

\subsection{Operating Points and Clinical Use}\label{sec:disc-operating}
The fixed 0.50 decision threshold used above places vertebra-level and patient-level sensitivity at 57.5\% and 69.3\%, but these reflect a conservative operating point rather than a limit of the model, since under this imbalance a default threshold favors specificity while discrimination itself remains high, with a vertebra-level AUPRC of 75.6\%. We do not present these operating points as evidence of clinical readiness. The question relevant to this study is narrower, namely whether masks approximated from 2D projections retain enough fracture-relevant signal to be tuned toward a deployment-appropriate operating point at all, and the operating-point analysis (Section~\ref{sec:operating-point}) indicates that they do. Sensitivity can be raised substantially by relocating the threshold without retraining, which suggests that the approximation preserves discriminative information well beyond what the default threshold exposes. For a screening-oriented role, where a study marked fracture-free receives only a brief read and sensitivity is paramount, operating at the F2 or F3 thresholds reaches patient-level sensitivities of 87.9\% and 96.1\% at specificities of 73.1\% and 44.6\%, while the adaptive-average fusion gives a more balanced 82.5\% sensitivity at 83.6\% specificity for supplementary use. Even at its most sensitive setting the model still misses close to 4\% of fractured patients and about one in six fractured vertebrae, so the appropriate reading is not that the projection-based pipeline is ready to substitute for radiologist review, but that it shows potential in a screening-oriented direction. This caution is not specific to the projection-driven approximation. Independent external validations of the RSNA-2022 challenge winners, including the first-place architecture reproduced as our baseline, report substantial performance loss when the models move from the curated competition test set to naturally imbalanced clinical cohorts \citep{Hu2024RSNAClinicalValidation, Harper2025ExternalValidation}, indicating that the gap between benchmark and deployment performance reflects the evaluation regime rather than any single method. The configurability demonstrated here is itself further evidence for the viability of the projection-driven approximation, though a deployment-grade operating point would still require prospective validation on clinically representative data.

\subsection{Explainability and Clinical Translation}\label{sec:disc-explainability}

Aggregate metrics show that the projection-based pipeline performs well but do not reveal whether the model is reading true fracture evidence from the approximated volumes or exploiting incidental correlations. The explainability analysis offers a qualitative check on this question. The 2.5D classifier exposes two views of its reasoning, the transformer's attention over the slice-stack sequence and the spatial Grad-CAM response within each stack, and in the representative case of Section~\ref{sec:explainability} the two coincide on the fracture. The transformer concentrated its weight on slice stacks 3, 7, and 8, and the Grad-CAM maps within those stacks localized to the annotated fracture site (Figure~\ref{fig:explainability}). The volume in this case was cropped with projection-approximated masks rather than full 3D segmentation, yet the model selected the slices containing the fracture and attended to the fractured region within them. This suggests that the approximated volume still carries the fracture-relevant morphology and that the model uses it, complementing the population-level comparisons of Section~\ref{sec:disc-viability} at the level of an individual decision.

\subsection{Per-Vertebra Difficulty and Interobserver Agreement}\label{sec:disc-patterns}

A consistent anatomical pattern runs through fracture recognition and the expert annotations. C2 has both the highest fracture F1-score and the strongest expert agreement (Fleiss $\kappa = 0.746$), and C1, C6 and C7 follow it in expert agreement ($\kappa = 0.630$ to $0.643$). The morphologically homogeneous mid-cervical vertebrae C3 to C5 divide the experts most sharply, with C4 reaching only fair agreement and C3 and C5 falling to chance. Segmentation Dice is also lowest at C3 to C5. Fracture recognition follows this pattern only in part. C3, the least frequently fractured level, has the lowest F1-score, whereas C4 is recognized about as well as C7 (68.17\% against 68.02\%) and better than C6, despite having fewer fractured examples than either. How recognizable a level's fractures are therefore matters alongside how often they occur. The two levels at which recognition is weakest, C3 and C5, are also two of the three on which experts disagree most, which suggests that much of the difficulty lies in the anatomy rather than in the pipeline.

The projection-derived masks are unlikely to explain the mid-cervical difficulty, since the per-level comparison with the 3D-segmentation masks showed no level at which the projection-derived masks were significantly worse and favored them at C4 and C6 (Section~\ref{sec:mask-source}). The AUPRC estimates at C1 to C3 nevertheless leaned toward the 3D-segmentation masks ($-2.30$, $-1.77$ and $-3.51$ points), with intervals too wide to show a difference. A plausible explanation is the overlap of C1 and C2 at the dens in both projections, which the multi-label formulation of Section~\ref{sec:multilabel-mask-generation} labels but cannot resolve along the projection axis, so the crops of these vertebrae may include more of their neighbors. Against the reproduced winning solution, the leaning levels were C3 and C7 ($-6.18$ and $-2.68$ points). At C7 the comparison with the 3D-segmentation masks leaned the other way ($+1.22$ points), so the C7 difference is more likely to reflect the rest of the pipeline than the masks.

Set against the radiologists on the 30-patient subset, the model's agreement with the reference standard was 0.600 at the patient level, within the range the radiologists themselves reached with the reference (0.40 to 0.67), and 0.711 at the vertebra level, above their range of 0.37 to 0.46. It also recovered 19 of the 29 reference fractures against a maximum of 11 for any single reader. The radiologists agreed more closely with one another, at 0.80 and 0.76 between the two most concordant readers, than with the reference. There is one caveat, because the reference labels originate in the institutional radiology reports rather than in the radiologists' annotations (Section~\ref{dataset-description}) \citep{Lin2023RSNA} and the model was trained to reproduce them, so the model and the reference share an origin that the independent readers do not, which may account for the model's closer alignment with the reference and the readers' closer alignment with one another.

\subsection{Computational Cost and Dimensionality}\label{sec:disc-dimensionality}

The viability the preceding sections establish has a quantitative counterpart in the cost of the segmentation stage, which is where the projection approach changes the dimensionality of the computation. Estimating each vertebral mask from two orthogonal 2D projections rather than from the full 3D volume reduces the segmentation stage from the 836.8 GFLOPs of the RSNA winner's 3D U-Net to 33.6 GFLOPs, a reduction of roughly twenty-five-fold, and uses fewer parameters as well, 27.2M against 39.7M (Section~\ref{sec:experimental_setup}) \citep{HaKaggle2022}. This comparison concerns the segmentation stage only. The detection and classification stages, including the two-model score-fusion classifier, carry their own costs (Table~\ref{tab:compute}), and we do not compare the pipelines as a whole.

The contribution here is therefore a demonstration that the localization-by-segmentation step, which conventionally relies on 3D segmentation, can be carried out in a lower-dimensional projection representation at little cost to diagnostic performance, rather than a reduction in the overall cost of the pipeline. Having established this viability, a natural next step is to pursue a more efficient classification stage, so that the saving realized at segmentation can extend to the pipeline as a whole.

\section{Limitations and Future Scopes}\label{sec:limitations}
Despite the promising results of our projection-based cervical spine fracture detection pipeline, several limitations must be acknowledged regarding its clinical applicability and generalizability. All results are obtained through internal cross-validation on the single public RSNA 2022 Cervical Spine Fracture dataset, and although this cohort aggregates studies from many institutions, the pipeline has not been assessed on an independent external test set or in a prospective clinical setting, so its behavior under unseen scanners, populations, and acquisition protocols remains to be established. The dataset's strict inclusion criteria are a related constraint. Its requirement of axial non-contrast CT with 1mm slice thickness is not restrictive in current practice, since thin-slice acquisition is standard on modern scanners, but our pipeline has been validated only on such thin-slice data and is not expected to transfer to thicker-slice studies, for which automated fracture analysis is generally less appropriate. A further deployment constraint arises at the localization stage, whose sequential slice selection is initialized from a fixed heuristic sagittal slice range of 100--420 (Section~\ref{sec:slice-selection}). This range was calibrated to the RSNA cohort, whose scans share a broadly consistent field of view. On unselected clinical scans, where acquisitions may begin at varying positions within the head or neck, this fixed initialization is unlikely to be appropriate and an additional head or neck localization step may be required before cervical VOI detection. The dataset's exclusion of post-surgical cases due to streak artifacts and altered anatomy also represents a significant limitation, as post-operative monitoring is a substantial part of clinical imaging, and its treatment of acute and chronic fractures as equivalent entities may not reflect clinical reality where differentiation is crucial for treatment decisions. Finally, because the reference labels are derived from the contributing institutions' radiology reports, the model trained to reproduce them inherits any systematic biases of that reference standard, a factor that also shapes the interobserver comparison, which was itself conducted on a limited 30-patient subset.

Another primary limitation stems from approximating each 3D vertebral mask from two 2D projections. The classifier still reads the original CT within each extracted volume, but intersecting two back-projected masks bounds a vertebra only by its two silhouettes, so the crop boundaries are imprecise and can take in parts of neighboring vertebrae. This is most pronounced at C1 (atlas) and C2 (axis), where the dens and the ring of the atlas overlap in both projections. Because segmentation is scored against ground-truth masks projected into the same views, Dice does not register this overlap. The per-level AUPRC estimates at C1 to C3 lean toward the 3D-segmentation masks, although their 95\% intervals include zero (Supplementary Section~S8.2). Bony connections between adjacent vertebrae, ranging from bridging osteophytes to complete fusion, may also impede per-vertebra separation in the projection domain. Supplementary Section~S8.5 shows a deficit of about one Dice point on ankylosed vertebrae with no significant difference in fracture recognition, although the latter rests on only 30 fractured ankylosed vertebrae from the 87 voxel-annotated patients. Furthermore, our model's reliance on 15 equally spaced slices may be suboptimal, since fracture information is not uniformly distributed and this rigid sampling strategy might not fully capture patient-specific anatomical diversity or varying injury severity. On a functional level, our pipeline identifies which vertebrae are fractured but does not localize the exact fracture site within the bone. From an implementation standpoint, the multi-stage design creates a risk of error propagation, a sensitivity heightened by the limited ground truth segmentation data available for only 87 patients. Moreover, the computational saving of the projection method is confined to the segmentation stage and does not extend to the pipeline as a whole, which also includes detection and a two-model score-fusion classifier.

We aim to overcome these limitations and further enhance the clinical utility and robustness of our approach in future work. We will explore the combination of multiple projection techniques as input to improve information retention from the 3D CT volumes, and we will replace the fixed equally spaced sampling with a content-aware slice-selection strategy that adapts to where fracture-relevant information is concentrated. To address the scarcity of ground truth masks for segmentation, we plan to leverage self-supervised pre-training methods or pre-training on other datasets, thereby improving segmentation accuracy and generalizability. Semi-supervised 3D segmentation combined with parameter-efficient fine-tuning, such as low-rank adaptation \citep{Hu2022LoRA}, is another route to working with few annotated volumes, and comparing the masks it produces with projection-derived masks under the same downstream classifier would extend the comparison made here. We will also investigate a computational method to precisely separate the anatomical information of the C2 (axis) from the C1 (atlas) vertebra within the extracted volumes to prevent information blending, and we will evaluate ankylosed vertebrae on a larger cohort. Further investigation into a wider range of projections for fracture recognition, together with a more in-depth study of efficient fracture-recognition networks, will be conducted to improve both the accuracy and the efficiency of the system. To support clinical reading workflows, we will add localized fracture bounding-box predictions that pinpoint the fracture site within an identified vertebra and consolidate them into a single visualization per suspected fracture. Finally, we will pursue external multi-institutional and prospective validation across diverse scanners and acquisition protocols, including a field-of-view normalization step that enables cervical localization on unselected clinical scans, to establish the pipeline's robustness beyond the curated dataset used here.

Beyond the cervical spine, the viability shown here raises a broader question about the cost of supervision. Voxel-level 3D annotation is among the most expensive and least scalable forms of labeling in volumetric medical imaging, which is why dense 3D masks remain scarce even in large public datasets \citep{Tajbakhsh2020}; in the RSNA cohort, 3D vertebra masks were available for only 87 of the 2019 patients. Because localizing anatomy from 2D projections is viable in this setting, a natural direction for future work is to ask whether projection-domain annotation can stand in for full 3D annotation more generally, supervising the localization of anatomy for downstream structural-anomaly detection in other regions and volumetric modalities at a fraction of the labeling effort. Establishing whether projection-level supervision generalizes in this way would extend the present contribution from a single anatomy to a more annotation-efficient approach to localized anomaly detection in CT and other volumetric data.

\section{Conclusion}
This study investigated the viability of a projection-based pipeline for automated cervical spine fracture detection, in which 3D vertebral masks are approximated from a small number of optimized 2D projections rather than recovered by full volumetric segmentation. The three-stage pipeline localized the cervical spine and approximated per-vertebra 3D masks from orthogonal 2D views with strong accuracy, and the ensemble 2.5D spatio-sequential classifier improved fracture classification over the evaluated 2D projection-based and full-3D volumetric baselines, which were limited by poor discrimination and by overfitting on the full volumetric input, respectively. Task-specific projection selection favored variance projections for localization and energy projections for segmentation. The score-fusion ensemble combined complementary evidence from conventional and maximum-projection slice stacks, improving discrimination and precision, and the adaptive-average fusion provided a balanced sensitivity-specificity operating point, while the operating-point analysis showed that the model can be tuned toward the higher-sensitivity regime appropriate for screening.

In the matched comparison, with the classifier architecture fixed, replacing 3D-segmentation masks with projection-derived masks changed AUPRC by $+1.03$ and $+0.65$ points at the vertebra and patient levels, and the pipeline differed from the reproduced RSNA 2022 winning solution by $-0.55$ and $+0.15$ points. None of these AUPRC differences was statistically significant, and fold-wise 95\% confidence intervals excluded vertebra-level losses larger than 1.46 and 2.81 points in the two comparisons. On the 87 voxel-annotated patients, the training-matched ceiling was about 1.4 vertebra-level AUPRC points above the projection-derived masks. On the 30-patient reader subset, the model's agreement with the reference standard was within the radiologists' range at the patient level and above it at the vertebra level. These results indicate that approximating 3D vertebral masks from 2D projections is a viable substitute for full volumetric segmentation in this task, retaining enough anatomical and fracture-relevant context for fracture recognition while reducing the dimensionality of the segmentation stage.


\section*{CRediT authorship contribution statement}

\textbf{Fabi Nahian Madhurja:} Conceptualization, Methodology, Software, Formal analysis, Investigation, Data curation, Visualization, Writing – original draft. 
\textbf{Rusab Sarmun:} Methodology, Software, Formal analysis, Investigation, Visualization, Writing – original draft. 
\textbf{Muhammad E. H. Chowdhury:} Conceptualization, Resources, Writing – review \& editing, Supervision, Project administration. 
\textbf{Adam Mushtak:} Methodology, Validation, Data curation. 
\textbf{Israa Al-Hashimi:} Methodology, Validation, Data curation. 
\textbf{Sohaib Bassam Zoghoul:} Methodology, Validation, Data curation.

\section*{Declaration of Competing Interest}
The authors declare that they have no known competing financial
interests or personal relationships that could have appeared to
influence the work reported in this paper.

\section*{Data and Code Availability Statement}
The study utilizes the public training set from the \textit{RSNA Cervical Spine Fracture Challenge 2022}. Derived preprocessed data can be shared upon reasonable request, subject to the dataset's terms of use. The preprocessing and inference code, along with a demo interface for visualization, are available at \url{https://github.com/fabinahian/cervical_spine_fracture_identification}.

\section*{Declaration of Generative AI Use}
During the preparation of this work the author(s) used ChatGPT and
Claude in order to refine the language and improve readability. After
using these tools, the author(s) reviewed and edited the content as
needed and take(s) full responsibility for the content of the published
article. All technical content and claims were reviewed and approved by the authors.


\bibliographystyle{elsarticle-harv} 
\bibliography{refs}

\clearpage

\newgeometry{top=2cm, bottom=2.5cm, left=2.5cm, right=2.5cm}

\onecolumn 

\setcounter{section}{0}
\setcounter{figure}{0}
\setcounter{table}{0}
\setcounter{equation}{0}
\renewcommand{\thesection}{S\arabic{section}}
\renewcommand{\thefigure}{S\arabic{figure}}
\renewcommand{\thetable}{S\arabic{table}}
\renewcommand{\theequation}{S\arabic{equation}}
\renewcommand{\theHsection}{S\arabic{section}}
\renewcommand{\theHfigure}{S\arabic{figure}}
\renewcommand{\theHtable}{S\arabic{table}}
\renewcommand{\theHequation}{S\arabic{equation}}

\titleformat{\paragraph}[block]
  {\normalfont\normalsize\itshape}
  {\theparagraph}
  {1em}
  {}
\titlespacing*{\paragraph}{0pt}{2.0ex plus 1ex minus .2ex}{0pt}

\begin{center}
    \vspace*{-1.5cm}
    {\Large \textbf{Supplementary Material}}
    \vspace{0.5em}
    
    {\Large Tracing 3D Anatomy in 2D Strokes: Can Projection-Derived Masks Replace 3D Segmentation for Cervical Spine Fracture Recognition?}
    
    \vspace{1.5em}
    
\end{center}

\vspace{0.5em}
\hrule 
\vspace{2em}



\section{Literature Review Summary}
Table \ref{tab:study_summary} compares published methods for cervical spine fracture detection on vertebra identification, fracture localization, detection level, explainability, interobserver variability, and cross-validation. Many of these studies rely on small, non-public datasets, and few use cross-validation (5 of 22) or report interobserver variability (2 of 22).

\begin{table}[htbp]
  \centering
  \small
  \setlength{\tabcolsep}{5pt} 
  \renewcommand{\arraystretch}{1.2}

  \begin{threeparttable}
    \caption{ Summary of spine fracture detection studies.}
    \label{tab:study_summary}

    \newcolumntype{L}[1]{>{\raggedright\arraybackslash}m{#1}}
    \newcolumntype{C}[1]{>{\centering\arraybackslash}m{#1}}

    \begin{tabular}{@{} L{3.6cm} C{0.7cm} C{0.7cm} C{1.2cm} C{2.8cm} C{0.7cm} C{0.7cm} C{0.7cm} @{}}
      \toprule
      \textbf{Study} & \textbf{VI} & \textbf{FL} & \textbf{View} & \textbf{DL} & \textbf{EX} & \textbf{IOV} & \textbf{CV} \\
      \midrule
      \citet{Small2021}       & \xmark & \xmark & Axial     & Slice              & \xmark & \xmark & \xmark \\
      \citet{Esfahani2023}    & \xmark & \xmark & Axial     & Slice              & \xmark & \xmark & \xmark \\
      \citet{Bhavya2022}      & \cmark & \xmark & 3D        & Vertebra \& Patient& \xmark & \xmark & \xmark \\
      \citet{NicolaesCSI}     & \cmark & \xmark & 3D        & Vertebra \& Patient& \xmark & \xmark & \cmark \\
      \citet{RasulAnalytics}  & \xmark & \xmark & Axial     & Slice              & \xmark & \xmark & \xmark \\
      \citet{MahajanINCOFT}   & \xmark & \xmark & 3 Views   & Patient            & \xmark & \xmark & \xmark \\
      \citet{Chen2024}        & \cmark & \xmark & 3D        & Vertebra           & \xmark & \cmark & \xmark \\
      \citet{Gaikwad2024}     & \xmark & \xmark & Axial     & Slice              & \cmark & \xmark & \xmark \\
      \citet{Chlad2023}       & \xmark & \xmark & Axial     & Slice              & \cmark & \xmark & \xmark \\
      \citet{Nejad2023}       & \cmark & \cmark & Axial     & Slice              & \xmark & \xmark & \xmark \\
      \citet{Boonrod2022}     & \xmark & \cmark & Lateral X-ray & Image              & \xmark & \cmark & \xmark \\
      \citet{Yaseen2024}      & \cmark & \xmark & Axial     & Slice              & \cmark & \xmark & \xmark \\
      \citet{Golla2023}       & \cmark & \cmark & 3D        & Vertebra           & \cmark & \xmark & \cmark \\
      \citet{Sutradhar2025}   & \cmark & \cmark & 3 Views   & Slice              & \cmark & \xmark & \xmark \\
      \citet{Tomita2018}      & \xmark & \xmark & Sagittal  & Patient            & \cmark & \xmark & \xmark \\
      \citet{SalehinejadISBI} & \xmark & \xmark & 3D, Axial & Slice, Patient     & \cmark & \xmark & \cmark \\
      \citet{HungCervical}    & \cmark & \cmark & 2.5D      & Vertebra, Patient  & \xmark & \xmark & \cmark \\
      \citet{Kim2023PlatCon}  & \cmark & \xmark & 3D        & -                  & \xmark & \xmark & \xmark \\
      \citet{Singh2025HybridUNet}        & \cmark & \xmark & Axial   & Vertebra          & \cmark & \xmark & \xmark \\
      \citet{Vel2026LightweightCNN}      & \xmark & \xmark & Axial   & Slice             & \xmark & \xmark & \xmark \\
      \citet{Kanwal2024SpineEDLNet}      & \xmark & \xmark & Axial   & Slice             & \xmark & \xmark & \xmark \\
      \citet{Pandey2025BayesianEnsemble} & \xmark & \xmark & Axial   & Slice             & \cmark & \xmark & \cmark \\

      \bottomrule
    \end{tabular}

    \begin{tablenotes}[flushleft]
      \small
      \item VI: Vertebra Identification; FL: Fracture Localization; DL: Detection Level; EX: Explainability; IOV: Interobserver Variability; CV: Cross-Validation. \cmark~indicates presence/usage, and \xmark~indicates absence.
    \end{tablenotes}
  \end{threeparttable}
\end{table}

\section{Cervical Spine Anatomy and Dataset Characteristics}
\begin{figure}[htbp]
    \centering
    \includegraphics[width=0.95\linewidth]{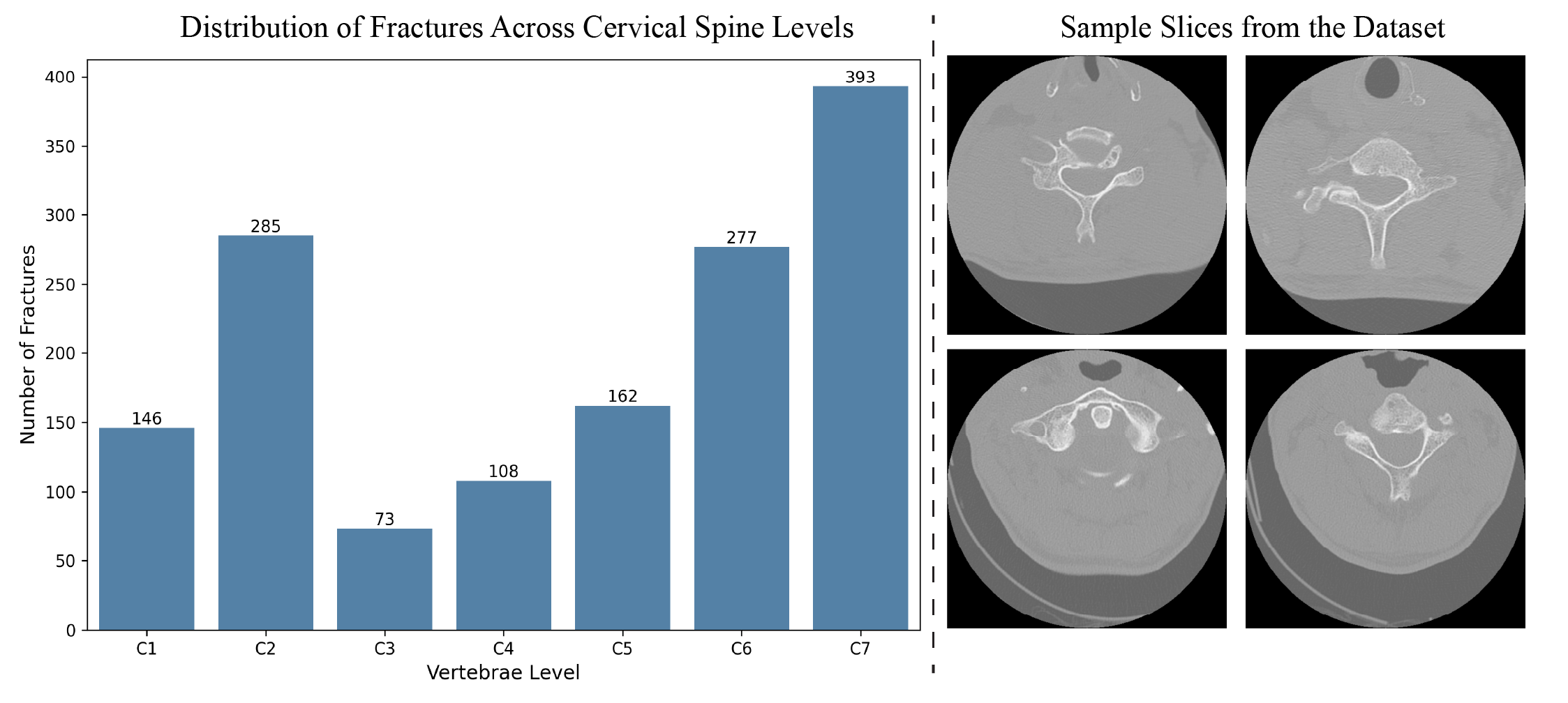}
    \caption{Distribution of fractures per vertebra and example CT data from the dataset.}
    \label{fig:data_distribution}
\end{figure}

\begin{figure}[htbp]
    \centering
    \includegraphics[width=0.95\linewidth]{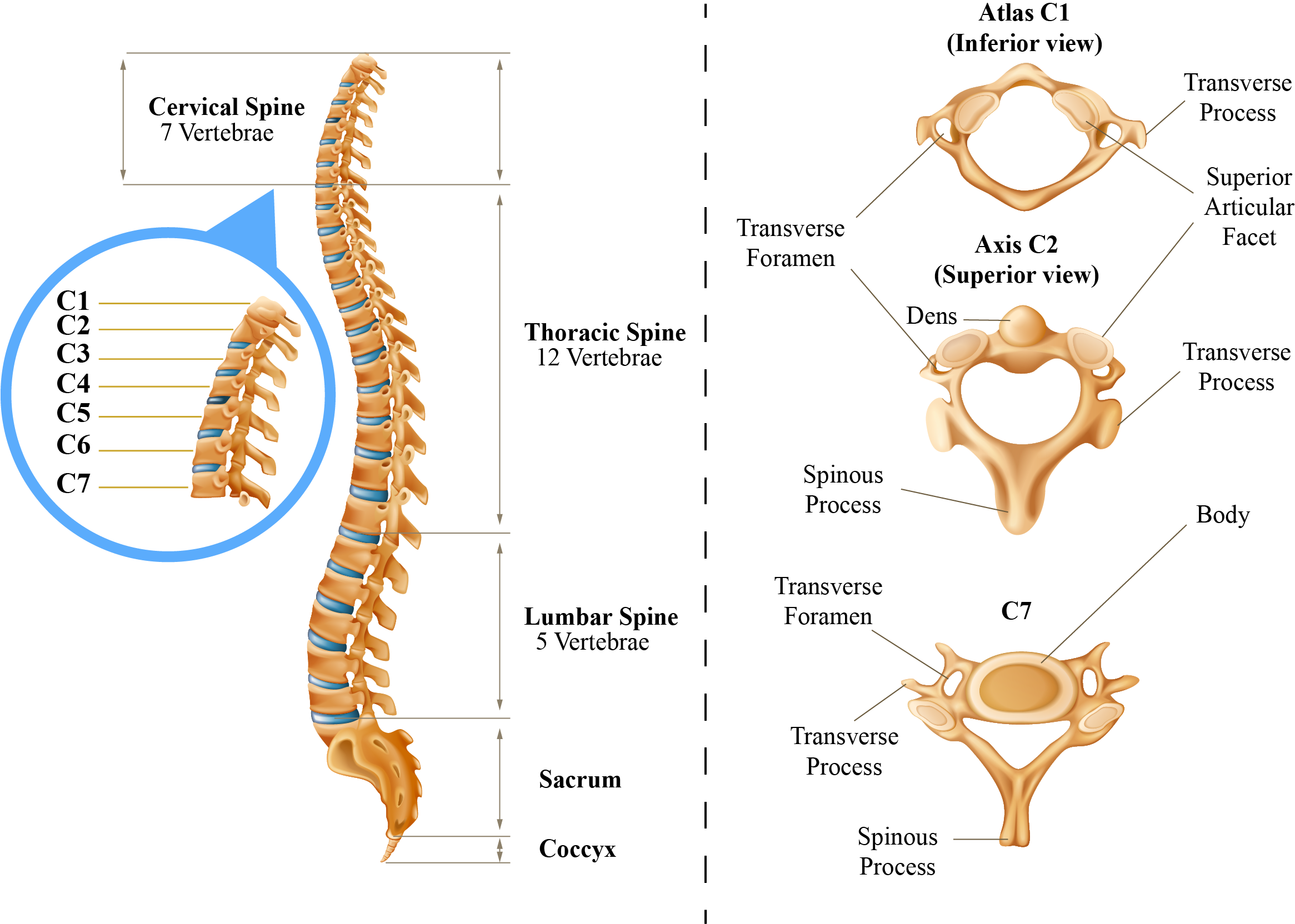}
    \caption{Anatomical structure of the cervical spine and the atypical cervical vertebrae. The illustration was adapted by the authors from a vector design by macrovector, licensed through Magnific (a Freepik Company).}
    \label{fig:S_anatomy}
\end{figure}

This section presents a visual overview of the RSNA 2022 Cervical Spine Fracture dataset distribution and image characteristics, and the anatomy of the cervical spine referred to in the Introduction of the main manuscript. Figure \ref{fig:data_distribution} illustrates the frequency of fractures across the cervical vertebrae (C1--C7). The data show noticeable class imbalance. Of the 14,133 vertebrae, 1,444 (10.2\%) are fractured, with fractures most frequent at C2, C6, and C7 and least frequent at C3 and C4. Representative CT slices illustrate the range of scan quality in the cohort. Figure~\ref{fig:S_anatomy} shows the anatomy of the cervical spine, including the atypical C1 (atlas) and C2 (axis) vertebrae.

\section{Mathematical Formulations of 2D Projections}\label{s3.-mathematical-formulations-of-2d-projections}
This section gives the projection operators compared in the pipeline, grouped by the stage in which they were used. Throughout, $I(x,y,z)$ is the intensity of the bone-windowed volume, $z = 1, \dots, N$ indexes the slices along the projection axis, $\mu(x,y)$ and $\sigma(x,y)$ are the mean and standard deviation of $I(x,y,\cdot)$ along that axis, and $I_z$ denotes slice $z$. After aggregation, every projection was min-max normalized to $[0, 1]$, so constant offsets and scale factors do not change the projected image.

\subsection{ROI Detection Projections}\label{s3.1.-roi-detection-projections}
\begin{figure}[htbp]
\centering
\includegraphics[width=\linewidth]{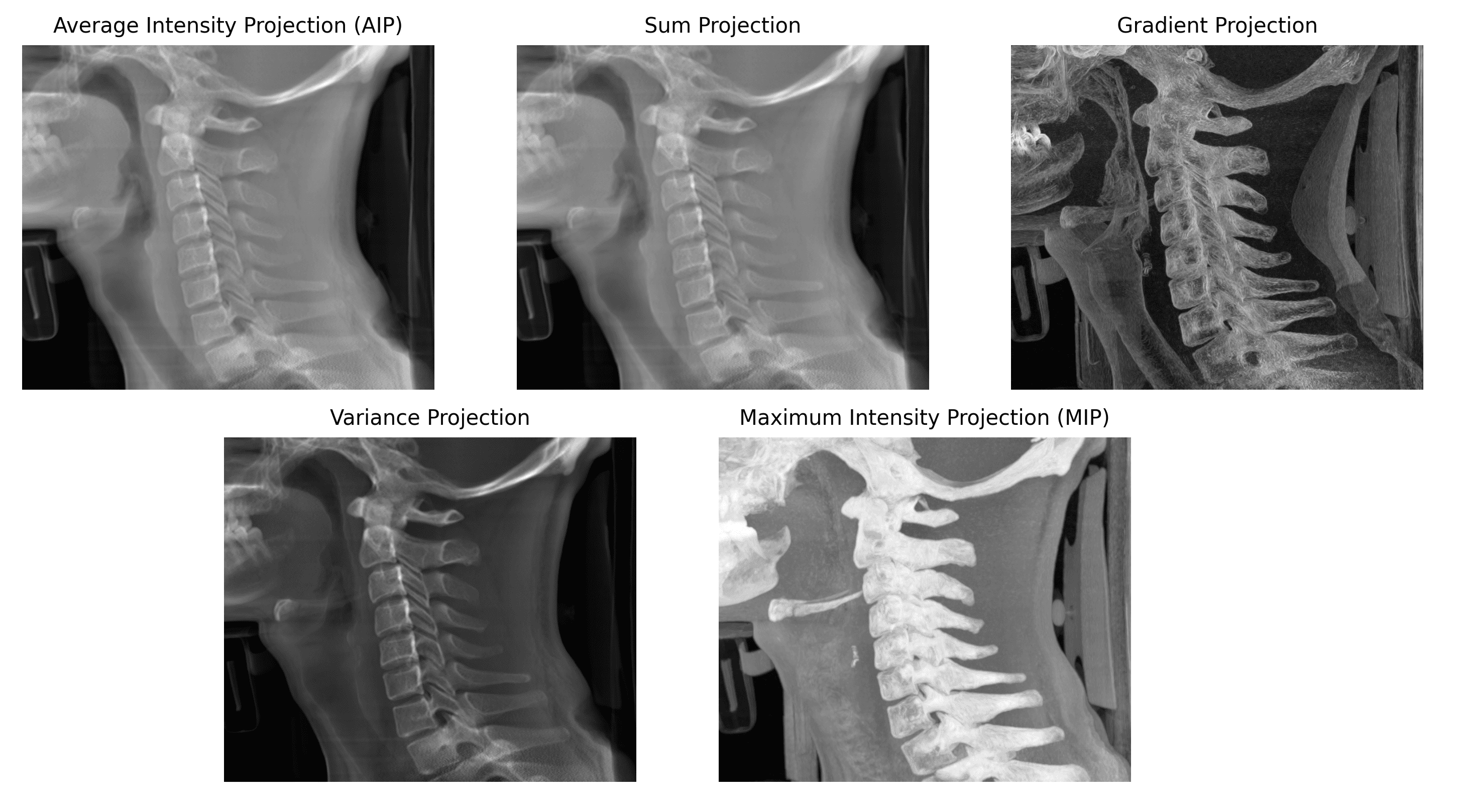}
\caption{Projections analyzed for cervical spine ROI detection.}
\label{fig:roi_projections}
\end{figure}

For cervical spine localization, we compared four projections suitable for capturing global anatomical structure. These correspond to the visual comparisons in Figure~\ref{fig:roi_projections}.

\begin{itemize}
  \item \textbf{Average Intensity Projection (AIP):} the mean intensity along the viewing axis.
\end{itemize}
\begin{equation}
I_{\mathrm{AIP}}(x,y) = \mu(x,y) = \frac{1}{N}\sum_{z = 1}^{N} I(x,y,z).
\tag{S1}
\end{equation}

\begin{itemize}
  \item \textbf{Gradient Projection:} the maximum over slices of the in-plane gradient magnitude, where $G(I_z)$ is computed with the Sobel operator \citep{Tian2021}.
\end{itemize}
\begin{equation}
I_{\mathrm{Gradient}}(x,y) = \max_{z}\, G(I_z)(x,y),
\tag{S2}
\end{equation}
\begin{equation}
G(I_z) = \sqrt{\left( \frac{\partial I_z}{\partial x} \right)^{2} + \left( \frac{\partial I_z}{\partial y} \right)^{2}}.
\tag{S3}
\end{equation}

\begin{itemize}
  \item \textbf{Variance Projection:} the variance of intensity along the viewing axis.
\end{itemize}
\begin{equation}
I_{\mathrm{Variance}}(x,y) = \frac{1}{N}\sum_{z = 1}^{N}\left( I(x,y,z) - \mu(x,y) \right)^{2}.
\tag{S4}
\end{equation}

\begin{itemize}
  \item \textbf{Maximum Intensity Projection (MIP):} the maximum intensity along the viewing axis.
\end{itemize}
\begin{equation}
I_{\mathrm{MIP}}(x,y) = \max_{z} I(x,y,z).
\tag{S5}
\end{equation}

\subsection{Segmentation Projections}\label{s3.2.-vertebra-segmentation-projections}
\begin{figure}[htbp]
\centering
\includegraphics[width=\textwidth]{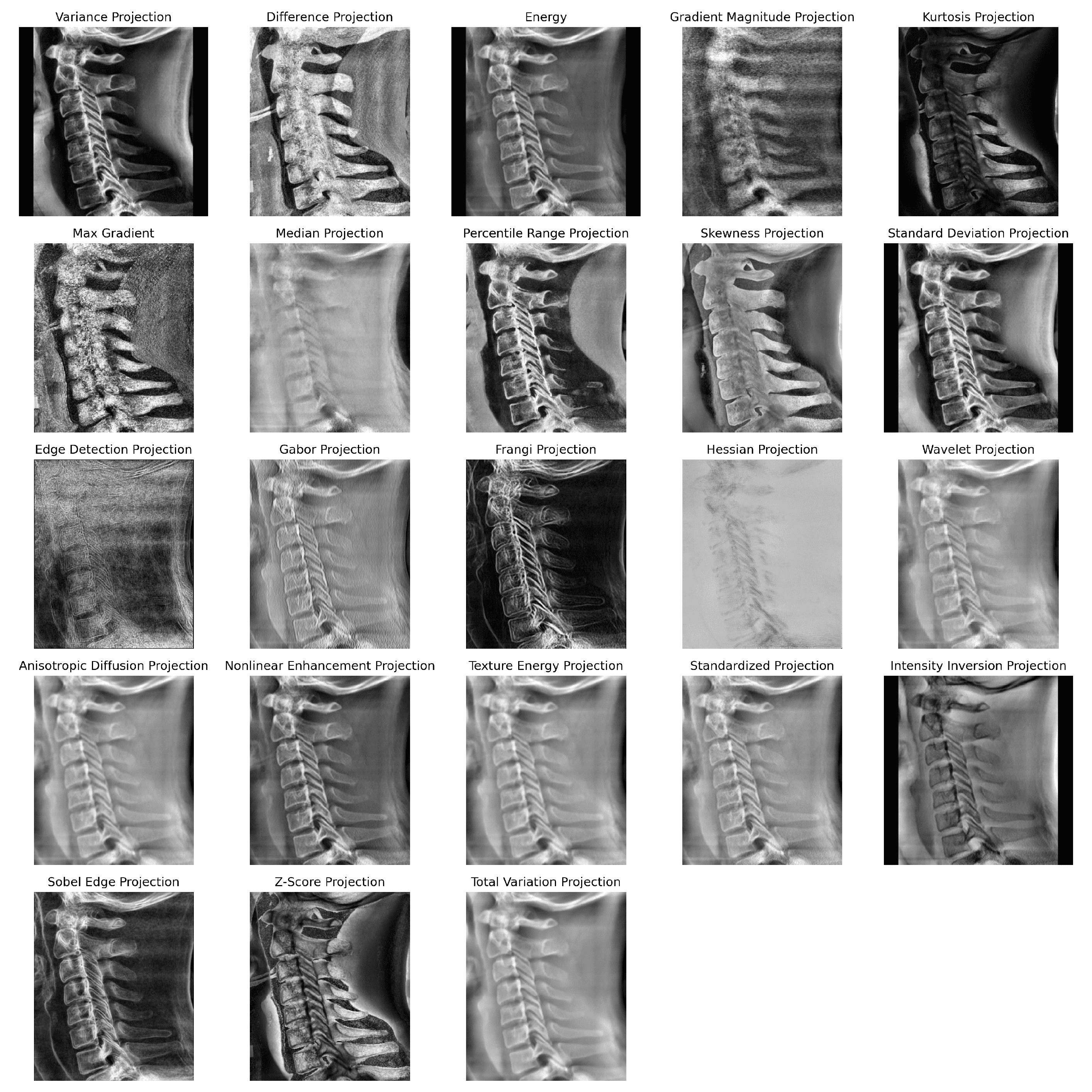}
\caption{Projections analyzed for vertebra segmentation (sagittal).}
\label{fig:seg_projections_sagittal}
\end{figure}

\begin{figure}[htbp]
\centering
\includegraphics[width=\textwidth]{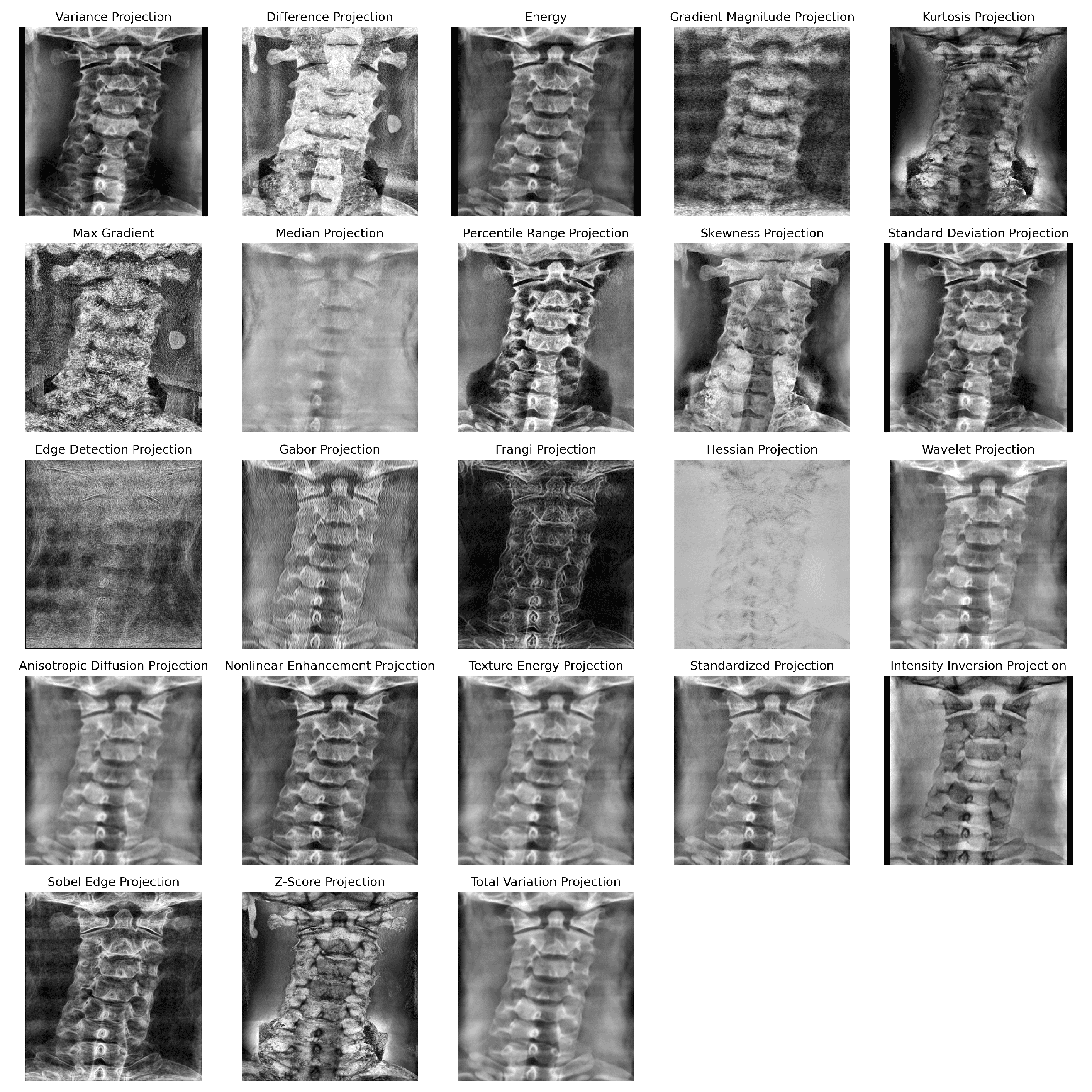}
\caption{Projections analyzed for vertebra segmentation (coronal).}
\label{fig:seg_projections_coronal}
\end{figure}

For vertebra segmentation, we compared 23 operators: the variance projection of Equation~S4 and the 22 operators below, which add statistical, derivative, filter-based and texture projections. Filter-based operators are applied to each slice and then aggregated over $z$. These definitions correspond to the visual comparisons in Figure~\ref{fig:seg_projections_sagittal} (sagittal) and Figure~\ref{fig:seg_projections_coronal} (coronal).

\begin{itemize}
  \item \textbf{Difference Projection:} the summed absolute change between adjacent slices.
\end{itemize}
\begin{equation}
I_{\mathrm{Difference}}(x,y) = \sum_{z = 1}^{N - 1}\left| I(x,y,z + 1) - I(x,y,z) \right|.
\tag{S6}
\end{equation}

\begin{itemize}
  \item \textbf{Energy Projection:} the sum of squared intensities across slices, emphasizing high-intensity structures such as cortical bone. Its sum-of-squares form follows the angular second moment texture feature \citep{Haralick2007}.
\end{itemize}
\begin{equation}
I_{\mathrm{Energy}}(x,y) = \sum_{z = 1}^{N} I(x,y,z)^{2}.
\tag{S7}
\end{equation}

\begin{itemize}
  \item \textbf{Gradient Magnitude Projection:} the summed magnitude of the intensity derivative along the projection axis, computed by central differences (one-sided at the first and last slice).
\end{itemize}
\begin{equation}
I_{\mathrm{Grad}}(x,y) = \sum_{z = 1}^{N}\left| \frac{\partial I(x,y,z)}{\partial z} \right|.
\tag{S8}
\end{equation}

\begin{itemize}
  \item \textbf{Max Gradient Projection:} the maximum magnitude of the same derivative along the projection axis.
\end{itemize}
\begin{equation}
I_{\mathrm{MaxGrad}}(x,y) = \max_{z}\left| \frac{\partial I(x,y,z)}{\partial z} \right|.
\tag{S9}
\end{equation}

\begin{itemize}
  \item \textbf{Kurtosis Projection:} the excess kurtosis of the intensity distribution across slices.
\end{itemize}
\begin{equation}
I_{\mathrm{Kurtosis}}(x,y) = \frac{\frac{1}{N}\sum_{z = 1}^{N}\left( I(x,y,z) - \mu(x,y) \right)^{4}}{\sigma(x,y)^{4}} - 3.
\tag{S10}
\end{equation}

\begin{itemize}
  \item \textbf{Median Projection:} the median intensity across slices.
\end{itemize}
\begin{equation}
I_{\mathrm{Median}}(x,y) = \mathrm{median}_{z}\, I(x,y,z).
\tag{S11}
\end{equation}

\begin{itemize}
  \item \textbf{Percentile Range Projection:} the difference between the 95\textsuperscript{th} and 5\textsuperscript{th} percentiles of intensity across slices, where $P_{q}(x,y)$ is the $q$\textsuperscript{th} percentile.
\end{itemize}
\begin{equation}
I_{\mathrm{PercentileRange}}(x,y) = P_{95}(x,y) - P_{5}(x,y).
\tag{S12}
\end{equation}

\begin{itemize}
  \item \textbf{Skewness Projection:} the asymmetry of the intensity distribution across slices.
\end{itemize}
\begin{equation}
I_{\mathrm{Skewness}}(x,y) = \frac{\frac{1}{N}\sum_{z = 1}^{N}\left( I(x,y,z) - \mu(x,y) \right)^{3}}{\sigma(x,y)^{3}}.
\tag{S13}
\end{equation}

\begin{itemize}
  \item \textbf{Standard Deviation Projection:} the spread of intensity values across slices.
\end{itemize}
\begin{equation}
I_{\mathrm{StdDev}}(x,y) = \sigma(x,y) = \sqrt{\frac{1}{N}\sum_{z = 1}^{N}\left( I(x,y,z) - \mu(x,y) \right)^{2}}.
\tag{S14}
\end{equation}

\begin{itemize}
  \item \textbf{Edge Detection Projection:} the number of slices in which each pixel is a Canny edge, where $\mathrm{C}(I_z)$ is the binary edge map of slice $z$ \citep{Canny2009}.
\end{itemize}
\begin{equation}
I_{\mathrm{Edge}}(x,y) = \sum_{z = 1}^{N} \mathrm{C}(I_z)(x,y).
\tag{S15}
\end{equation}

\begin{itemize}
  \item \textbf{Gabor Projection:} the summed real part of the response to a Gabor filter $g$ with spatial frequency $0.6$ \citep{Jain1991}, where $*$ denotes 2D convolution.
\end{itemize}
\begin{equation}
I_{\mathrm{Gabor}}(x,y) = \sum_{z = 1}^{N}\mathrm{Re}\left[ (I_z * g)(x,y) \right].
\tag{S16}
\end{equation}

\begin{itemize}
  \item \textbf{Frangi Projection:} the summed Frangi vesselness \citep{FrangiVessel}, which enhances tubular structures. At scale $s$, with Hessian eigenvalues $|\lambda_{1}| \le |\lambda_{2}|$, $R_{B} = \lambda_{1}/\lambda_{2}$ and $S = \sqrt{\lambda_{1}^{2} + \lambda_{2}^{2}}$, the vesselness is $V_{s} = \exp\left(-R_{B}^{2}/2\beta^{2}\right)\left(1 - \exp\left(-S^{2}/2c^{2}\right)\right)$. The maximum is taken over scales $s \in \{1, 3, 5, 7, 9\}$, with $\beta = 0.5$ and $c$ set from the Hessian norm (library defaults).
\end{itemize}
\begin{equation}
I_{\mathrm{Frangi}}(x,y) = \sum_{z = 1}^{N}\max_{s} V_{s}(I_z)(x,y).
\tag{S17}
\end{equation}

\begin{itemize}
  \item \textbf{Hessian Projection:} the summed determinant of the Hessian of each slice at scale $1$, which highlights blob-like and curvilinear structures \citep{FrangiVessel}.
\end{itemize}
\begin{equation}
I_{\mathrm{Hessian}}(x,y) = \sum_{z = 1}^{N} \det H(I_z)(x,y).
\tag{S18}
\end{equation}

\begin{itemize}
  \item \textbf{Wavelet Projection:} the summed approximation band of a single-level 2D Haar wavelet transform of each slice \citep{Mallat2002}, which has half the in-plane resolution.
\end{itemize}
\begin{equation}
I_{\mathrm{Wavelet}}(x,y) = \sum_{z = 1}^{N} A_{\mathrm{Haar}}(I_z)(x,y).
\tag{S19}
\end{equation}

\begin{itemize}
  \item \textbf{Anisotropic Diffusion Projection:} the mean of slices smoothed by five iterations of Perona--Malik diffusion, $\mathcal{D}_{5}$, which smooths while preserving edges \citep{Perona2002}.
\end{itemize}
\begin{equation}
I_{\mathrm{Diffusion}}(x,y) = \frac{1}{N}\sum_{z = 1}^{N}\mathcal{D}_{5}(I_z)(x,y).
\tag{S20}
\end{equation}

\begin{itemize}
    \item \textbf{Nonlinear Enhancement Projection:} enhances high-intensity areas using a nonlinear power function with exponent $p$.
\end{itemize}
\begin{equation}
I_{\mathrm{Nonlinear}}(x,y) = \frac{1}{N}\sum_{z = 1}^{N} I(x,y,z)^{p}.
\tag{S21}
\end{equation}

\begin{itemize}
  \item \textbf{Texture Energy Projection:} the summed absolute response to three $5 \times 5$ texture masks, $M \in \{L_{5}^{\top}L_{5},\, E_{5}^{\top}E_{5},\, S_{5}^{\top}S_{5}\}$, built from $L_{5} = [1, 4, 6, 4, 1]$, $E_{5} = [-1, -2, 0, 2, 1]$ and $S_{5} = [-1, 0, 2, 0, -1]$ \citep{Laws1980}.
\end{itemize}
\begin{equation}
I_{\mathrm{Texture}}(x,y) = \sum_{z = 1}^{N}\sum_{M}\left| (I_z * M)(x,y) \right|.
\tag{S22}
\end{equation}

\begin{itemize}
  \item \textbf{Standardized Projection:} the mean over slices after each slice is standardized by its own in-plane mean $m_{z}$ and standard deviation $s_{z}$.
\end{itemize}
\begin{equation}
I_{\mathrm{Standardized}}(x,y) = \frac{1}{N}\sum_{z = 1}^{N}\frac{I(x,y,z) - m_{z}}{s_{z}}.
\tag{S23}
\end{equation}

\begin{itemize}
  \item \textbf{Intensity Inversion Projection:} the mean over slices of the intensities inverted about the maximum of the whole volume, $I_{\max}$, which highlights low-intensity regions.
\end{itemize}
\begin{equation}
I_{\mathrm{Inverted}}(x,y) = \frac{1}{N}\sum_{z = 1}^{N}\left( I_{\max} - I(x,y,z) \right) = I_{\max} - \mu(x,y).
\tag{S24}
\end{equation}

\begin{itemize}
  \item \textbf{Sobel Edge Projection:} the summed in-plane Sobel gradient magnitude of each slice, $G(I_z)$ of Equation~S3 \citep{Sobel1990}.
\end{itemize}
\begin{equation}
I_{\mathrm{Sobel}}(x,y) = \sum_{z = 1}^{N} G(I_z)(x,y).
\tag{S25}
\end{equation}

\begin{itemize}
  \item \textbf{Z-Score Projection:} the largest absolute deviation from the mean along the projection axis, in units of the standard deviation along that axis.
\end{itemize}
\begin{equation}
I_{\mathrm{ZScore}}(x,y) = \max_{z}\frac{\left| I(x,y,z) - \mu(x,y) \right|}{\sigma(x,y)}.
\tag{S26}
\end{equation}

\begin{itemize}
  \item \textbf{Total Variation Projection:} the mean of slices denoised by total-variation minimization, $\mathcal{T}$, with weight $0.1$ \citep{Rudin1992}.
\end{itemize}
\begin{equation}
I_{\mathrm{TotalVariation}}(x,y) = \frac{1}{N}\sum_{z = 1}^{N}\mathcal{T}(I_z)(x,y).
\tag{S27}
\end{equation}

\section{Methodological Elaborations}
\label{sec:S4}
\setcounter{equation}{27}

This section gives the rationale for the multi-label segmentation masks, the definitions of the evaluation metrics, the slice-selection heuristic, the 2D and 3D baseline classifiers, the construction of the 2.5D inputs, and the segmentation loss.

\subsection{Justification for Multi-label Mask Generation}
\label{sec:S4_1}

\begin{figure}[htbp]
    \centering
    \includegraphics[width=\textwidth]{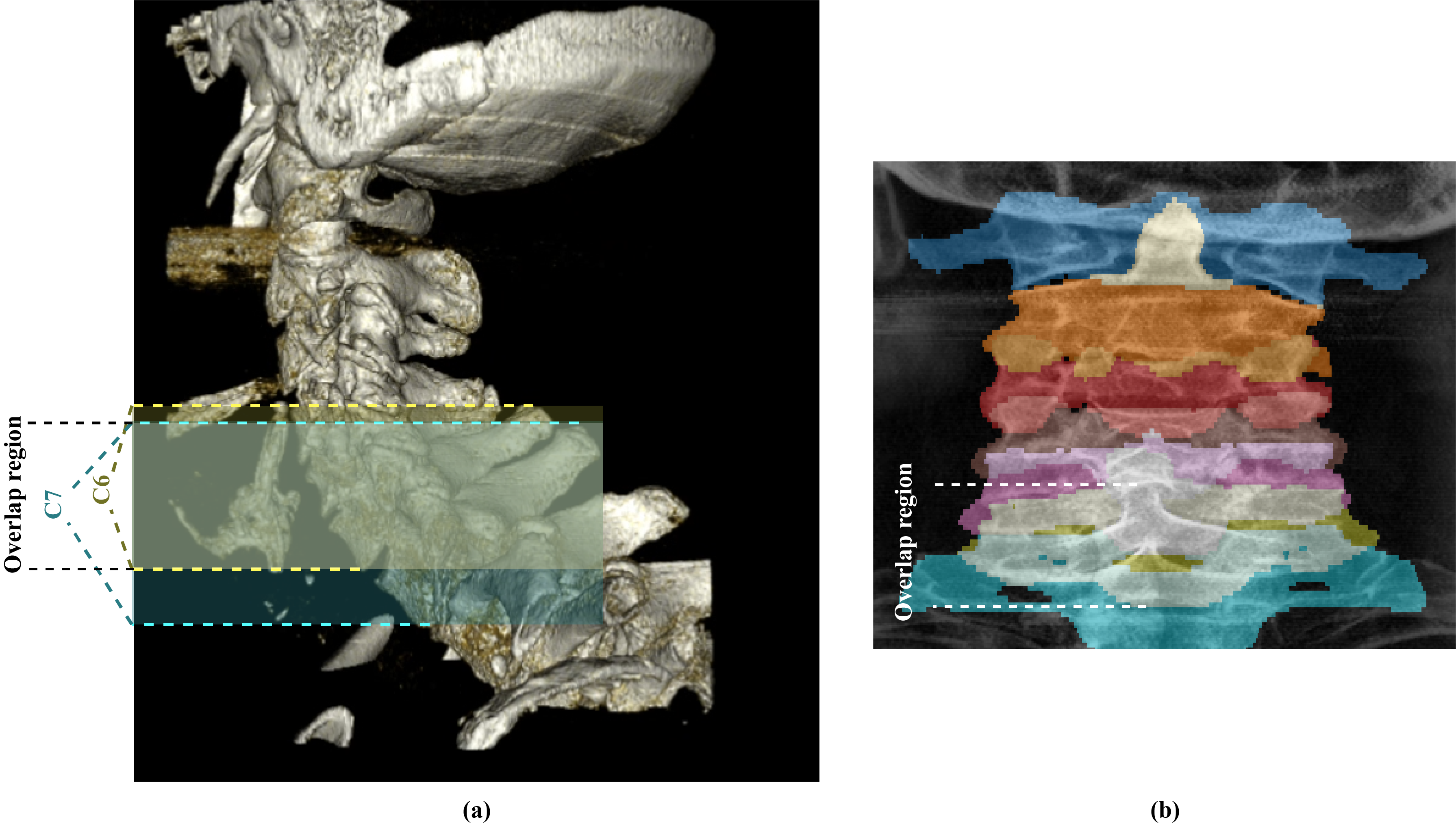}
    \caption{Overlap in the coronal projection (a) CT Volume (b) Multilabel Mask}
    \label{fig:coronal_overlap}
\end{figure}

\begin{figure}[htbp]
    \centering
    \includegraphics[width=\textwidth]{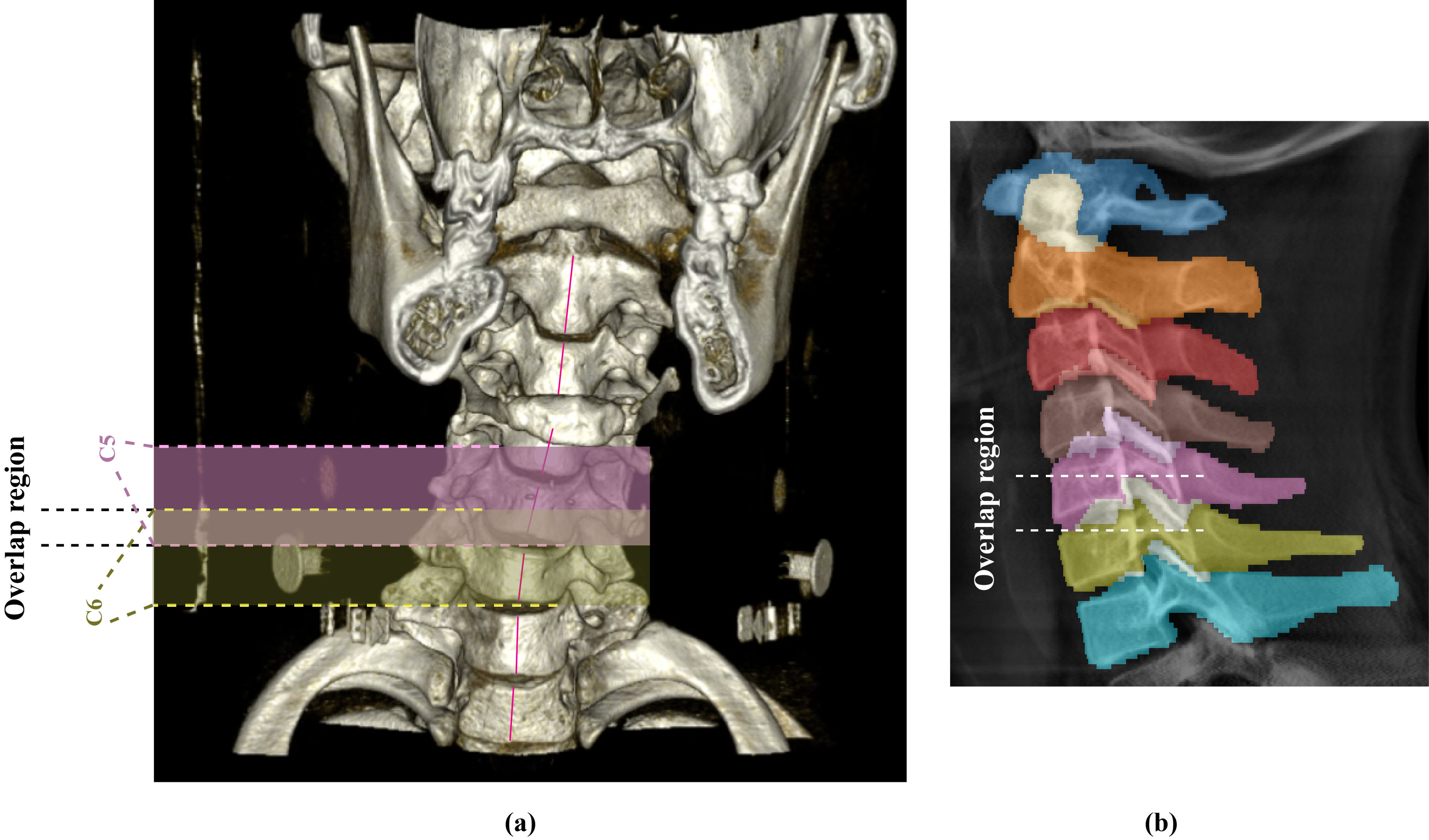}
    \caption{ Overlap in the sagittal projection (a) CT Volume (b) Multilabel Mask}
    \label{fig:sagittal_overlap}
\end{figure}

Since we are attempting to perform segmentation from the coronal and sagittal projections of the cervical spine VOI, some of the anatomical features of the spine overlap when viewed in orthogonal projections. Due to this overlap, a single pixel in the projection image can belong to two vertebrae that are overlapping from the orthogonal perspective. Many patients in the RSNA dataset do not have straight necks, so the spinous process of a vertebra can overlap the vertebral body of a neighboring vertebra. Figure \ref{fig:coronal_overlap} demonstrates the overlap between C6 and C7 vertebrae. As the spinous process of C7 is directly behind the vertebral body of C6 due to the curvature of the neck, when viewed from the coronal perspective the vertebra masks will overlap. We can see the overlap in the coronal projection in Figure \ref{fig:coronal_overlap}(b). Figure \ref{fig:sagittal_overlap} demonstrates a patient with a bent neck visible from the coronal perspective. The centerline of the neck is roughly shown with red lines. The transverse processes and vertebral bodies of the neighboring C5 and C6 vertebrae overlap in the sagittal projection, as shown in Figure~\ref{fig:sagittal_overlap}(b).

Because of these overlaps, standard multi-class segmentation was not suitable for this study. We therefore used a multi-label formulation in which a projection pixel can belong to more than one vertebra. The multi-label mask was constructed by keeping only the cervical vertebrae (C1--C7) from the provided 3D multi-class segmentation mask. Then the max projection of each vertebra was taken in the direction of the projection and the binary masks of each vertebra were stacked in separate layers.

\subsection{Evaluation Metrics}
\label{sec:S4_2}

\subsubsection{ROI Detection}
The evaluation metrics for ROI detection are formulated as follows:

\begin{equation}
IoU_{i} = \frac{\text{Area of }\left( P_{i} \cap G_{i} \right)}{\text{Area of }\left( P_{i} \cup G_{i} \right)}
\label{eq:iou_roi}
\end{equation}

\begin{equation}
mIoU = \frac{1}{N}\sum_{i = 1}^{N}{IoU_{i}}
\label{eq:miou_roi}
\end{equation}

\subsubsection{Vertebra Segmentation}
Vertebra segmentation was evaluated using two spatial-overlap metrics, the Intersection over Union (IoU) and the Dice Similarity Coefficient (DSC), together with the Hausdorff distance for boundary agreement. The IoU and DSC are computed from the true positives ($TP$), false positives ($FP$), and false negatives ($FN$) of the predicted mask, whereas the Hausdorff distance $H(A, B)$ measures the maximum deviation between the predicted and reference boundary point sets $A$ and $B$. The 95th-percentile variant (HD95) reported in the main manuscript follows from this definition by taking the 95th percentile in place of the maximum, which reduces sensitivity to outlying points. These metrics are defined as:

\begin{equation}
IoU = \frac{TP}{TP + FP + FN}
\label{eq:iou_seg}
\end{equation}
\begin{equation}
DSC = \frac{2TP}{2TP + FP + FN}
\label{eq:dsc_seg}
\end{equation}
\begin{equation}
H(A, B) = \max\left( \max_{a \in A}\left\{ \min_{b \in B}{|a - b|} \right\}, \max_{b \in B}\left\{ \min_{a \in A}{|b - a|} \right\} \right)
\label{eq:hausdorff}
\end{equation}

\subsubsection{Fracture Identification}
Fracture classification was evaluated at both the vertebra and patient levels using standard confusion-matrix metrics, where $TP$, $TN$, $FP$, and $FN$ denote the true positives, true negatives, false positives, and false negatives for the fractured class.

\begin{equation}
\text{Accuracy} = \frac{TP + TN}{TP + TN + FP + FN}
\label{eq:accuracy_cls}
\end{equation}

\begin{equation}
\text{Precision} = \frac{TP}{TP + FP}
\label{eq:precision_cls}
\end{equation}

\begin{equation}
\text{Sensitivity} = \frac{TP}{TP + FN}
\label{eq:sensitivity_cls}
\end{equation}

\begin{equation}
\text{Specificity} = \frac{TN}{TN + FP}
\label{eq:specificity_cls}
\end{equation}

\begin{equation}
\text{F1} = \frac{2\,TP}{2\,TP + FP + FN}
\label{eq:f1_cls}
\end{equation}

The two threshold-independent metrics summarize performance across all decision thresholds rather than at a single operating point. The ROC-AUC is the area under the curve of the true positive rate (TPR) against the false positive rate (FPR), and the AUPRC is the area under the curve of precision against recall (Rec), both swept over the continuum of decision thresholds.

\begin{equation}
\text{ROC-AUC} = \int_{0}^{1} \text{TPR}\; d(\text{FPR})
\label{eq:rocauc_cls}
\end{equation}

\begin{equation}
\text{AUPRC} = \int_{0}^{1} \text{Precision}\; d(\text{Rec})
\label{eq:auprc_cls}
\end{equation}

\subsection{Ground-Truth Boxes and Slice Selection}
\label{sec:S4_3}

\begin{figure}[htbp]
    \centering
    \includegraphics[width=\textwidth]{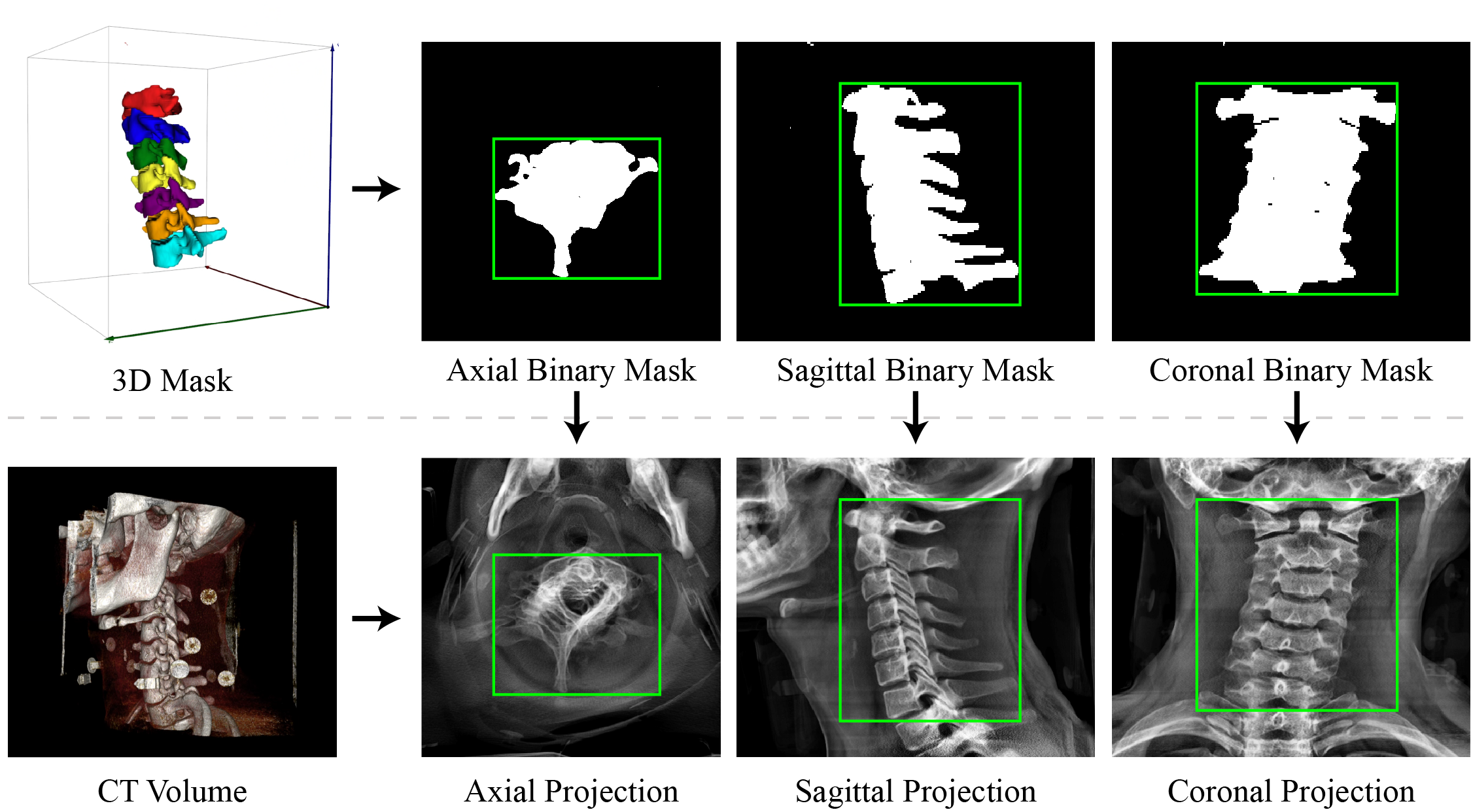}
    \caption{2D ROI bounding box calculation from a 3D mask.}
    \label{fig:roi_calculation}
\end{figure}

Ground-truth boxes for the detector were derived from the 87 patients with voxel-level masks. In each of the axial, sagittal and coronal views, the C1--C7 labels were combined into a binary cervical mask and max-projected, and the box was set to the extent of the largest connected component, so that small isolated fragments of the mask do not enlarge it (Figure~\ref{fig:roi_calculation}).

\begin{figure}[htbp]
    \centering
    \includegraphics[width=0.6\textwidth]{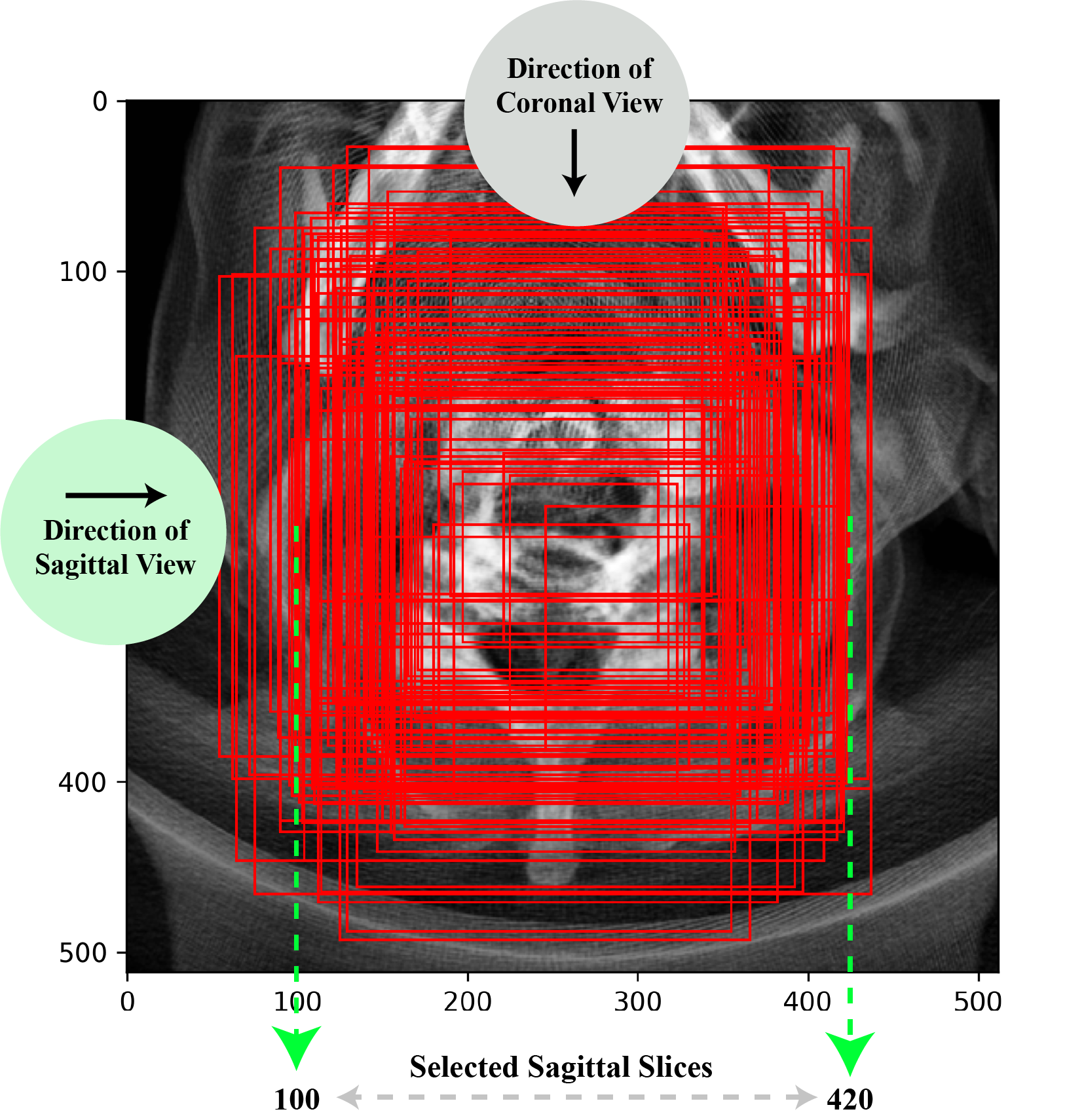}
    \caption{Heuristic selection of sagittal slices.}
    \label{fig:slice_selection}
\end{figure}

\begin{figure}[htbp]
    \centering
    \includegraphics[width=0.6\textwidth]{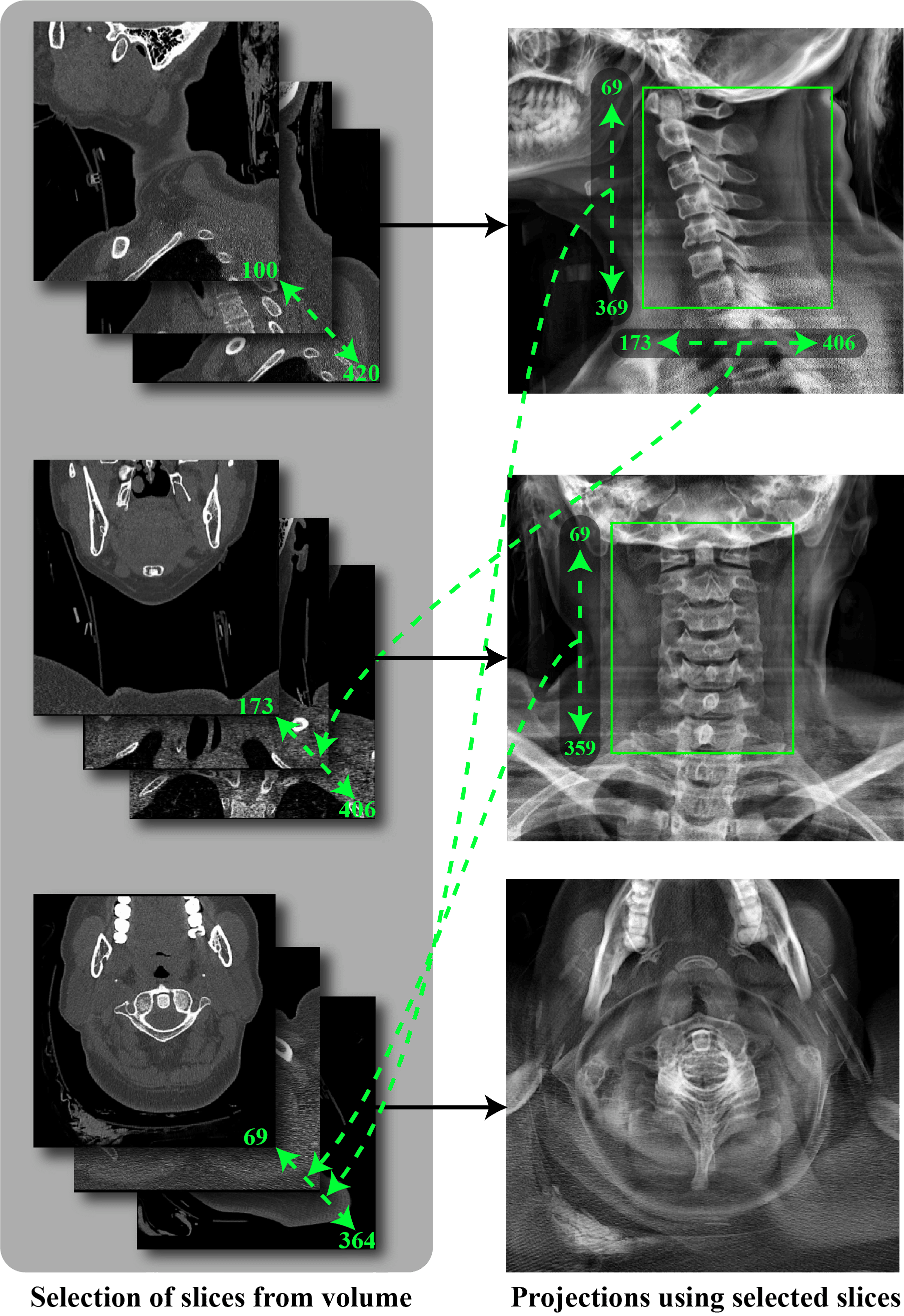}
    \caption{Sequential slice-selection process.}
    \label{fig:seq_slice_selection}
\end{figure}

Each projection was generated only from the slices that contain the cervical spine, and the slice ranges were set sequentially (Figure~\ref{fig:seq_slice_selection}). The aggregated ground-truth boxes of the 87 patients, after a few outliers were excluded, placed the cervical spine within sagittal slices 100 to 420 (Figure~\ref{fig:slice_selection}). This range excludes peripheral anatomy such as the shoulders without cutting into the vertebrae, and it was used for the sagittal projection of every patient. The coronal and axial ranges were then derived for each patient. The width of the resulting sagittal box set the coronal slice range (slices 173--406 in the example of Figure~\ref{fig:seq_slice_selection}), and the axial range was the average of the coronal and sagittal box heights.

\subsection{Baseline Fracture Identification Architectures}
\label{sec:S4_4}

For comparison with the proposed 2.5D spatio-sequential model, we implemented two alternative classifiers, a 2D network applied to projections of the vertebra and a 3D network applied to the full vertebra volume. Their results are reported in Section~4.3.1 of the main manuscript.

\subsubsection{Directly Projection-Based 2D Networks}
\label{sec:S4_4_1}

This approach classifies 2D projections of each segmented vertebra rather than the vertebra volume itself.

\begin{figure}[htbp]
    \centering
    \includegraphics[width=\textwidth]{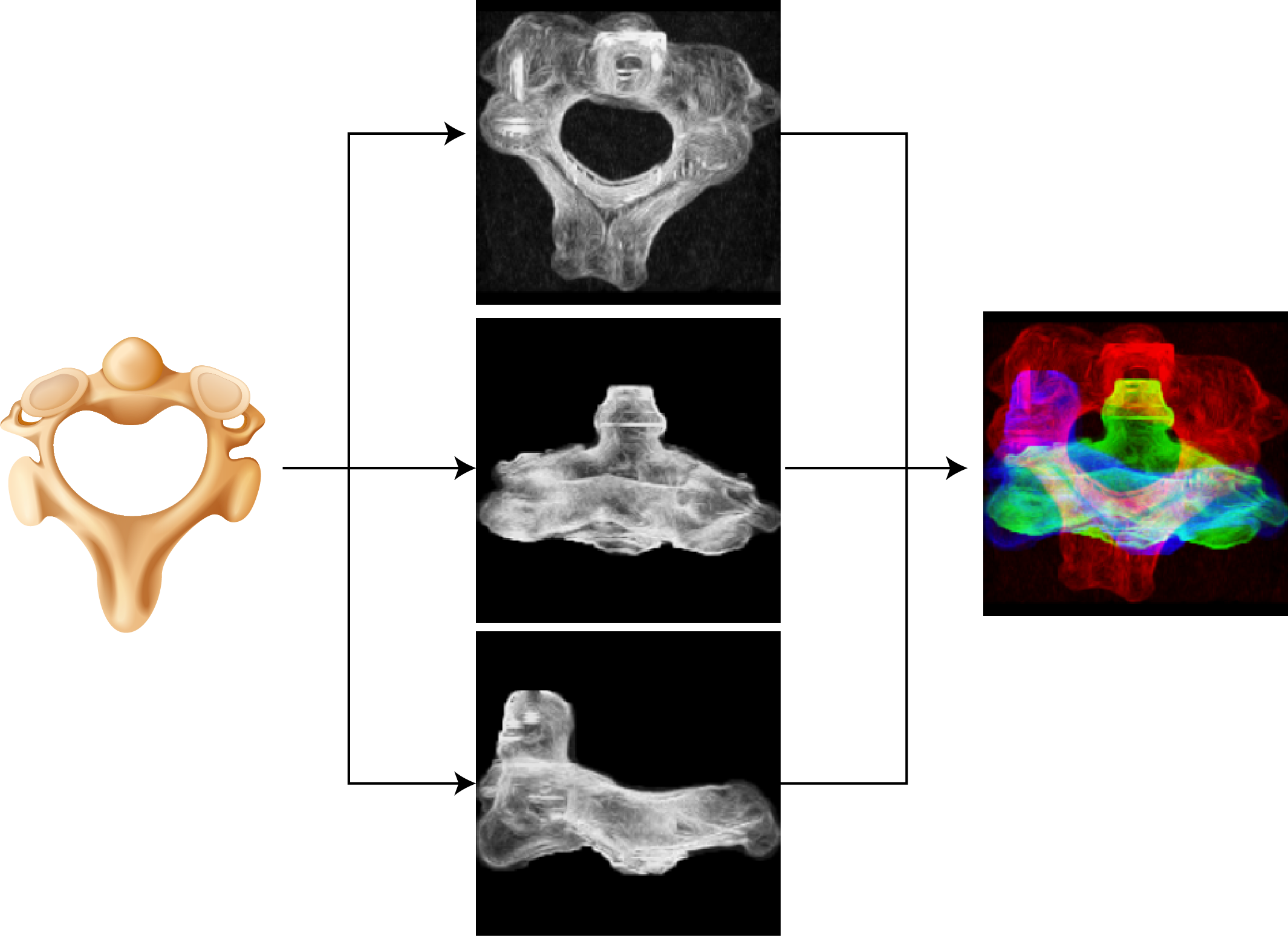}
    \caption{Multi-channel projection strategy stacking axial, sagittal, and coronal views for vertebra-level classification.}
    \label{fig:multi_view_fusion}
\end{figure}

\begin{itemize}
    \item \textbf{Input \& Preprocessing:} We utilized variance projections of the vertebrae from axial, sagittal, and coronal perspectives. A bone window (level $400$ HU, width $1800$ HU) was applied prior to projection to enhance visibility.
    \item \textbf{Fusion Strategy:} For vertebra-level classification, the three orthogonal projections were resized and stacked along the channel axis to form a composite RGB input, providing the model with simultaneous multi-view context (see Figure \ref{fig:multi_view_fusion}). For patient-level classification, whole-spine sagittal and coronal projections were also classified directly, and their predictions were combined by score-level fusion via weighted averaging (Table~8 of the main manuscript).
    \item \textbf{Architectures:} We evaluated standard CNN backbones including ResNet18, ResNet50, DenseNet121, and MobileNetV2, alongside a custom 3-layer CNN.
\end{itemize}

\subsubsection{Full-Volume 3D CNN Networks}
\label{sec:S4_4_2}

This approach utilizes the entire 3D volume of the segmented vertebra, preserving full spatial context at the cost of computational intensity.

\begin{itemize}
    \item \textbf{Preprocessing:} Vertebra volumes were extracted, preprocessed with a bone window and resampled to $64 \times 224 \times 224$ (Table~1 of the main manuscript), and 3D-specific augmentations such as random rotation and elastic deformation were applied to improve generalization.
    \item \textbf{Architecture:} We employed an EfficientNet-B0 backbone adapted for 3D inputs using 3D convolutional kernels. The model processes the full voxel grid in a channel-first format to extract hierarchical spatial features across all dimensions.
    \item \textbf{Limitation:} In our experiments this model had low sensitivity (Section~4.3.1 of the main manuscript), likely because of the high dimensionality of the input relative to the available training data.
\end{itemize}

\subsection{Input Preprocessing for 2.5D Networks}
\label{sec:S4_5}

\subsubsection{3D Mask Approximation for 2.5D}
\begin{figure}[htbp]
  \centering
  \includegraphics[width=\textwidth]{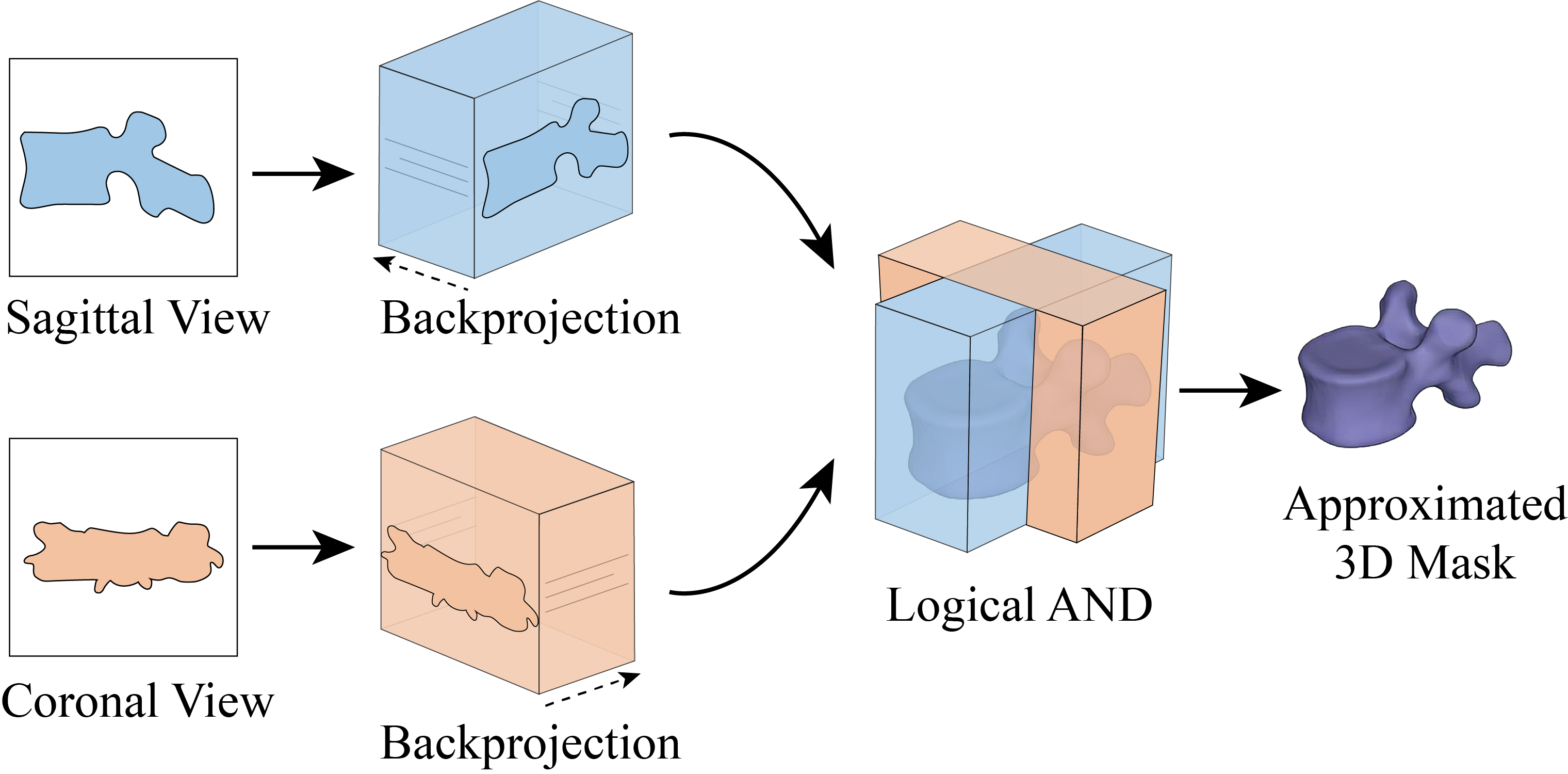}
  \caption{3D Mask Approximation Process}
  \label{fig:approximate_mask}
\end{figure}

The 2D masks from the sagittal and coronal projections were back-projected along their viewing axes, and the logical AND of the two back-projected volumes was taken as the projection-derived 3D mask of the vertebra (Figure~\ref{fig:approximate_mask}). These masks were used to extract the vertebra volumes classified by the 2.5D network.

\label{sec:S4_4_approx}

\subsubsection{Stack Generation for 2.5D}
\begin{figure}[htbp]
    \centering
    \includegraphics[width=0.6\textwidth]{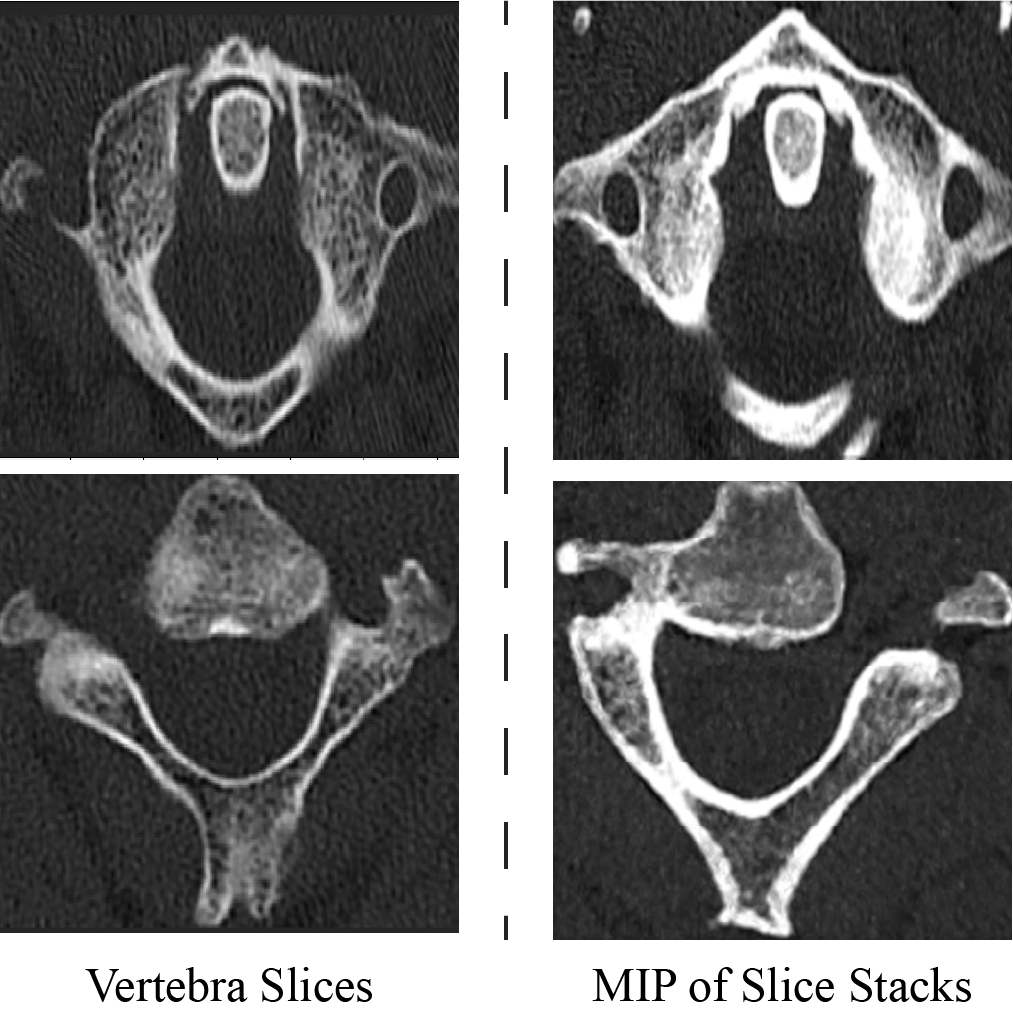}
    \caption{Example inputs to the 2.5D networks, showing raw vertebral slices and the corresponding maximum intensity projections. Single images are shown in place of full stacks.}
    \label{fig:2.5d_input_example}
\end{figure}

Figure~\ref{fig:2.5d_input_example} shows examples of the two 2.5D inputs described in Section~3.6.1 of the main manuscript, raw vertebral slices for the slice stack and maximum intensity projections for the MIP stack. Each projection combines five neighboring slices, so it shows more of the vertebra in a single image.

\subsection{Segmentation Loss}
\label{sec:S4_6}

The segmentation network (Section~3.5.3 of the main manuscript) was trained on the average of a Dice loss and a Jaccard loss. Let $y_{i,j}$ and $p_{i,j}$ denote the ground-truth label and predicted probability for pixel $i$ in channel $j$, with the sums taken over all $N$ pixels and $C$ vertebral channels, and let $\epsilon$ be a smoothing constant. Then
\begin{equation}
    L_{\text{Dice}} = 1 - \frac{2\sum_{i=1}^{N}\sum_{j=1}^{C} y_{i,j}\, p_{i,j} + \epsilon}{\sum_{i=1}^{N}\sum_{j=1}^{C}\left( y_{i,j} + p_{i,j} \right) + \epsilon},
\end{equation}
\begin{equation}
    L_{\text{Jaccard}} = 1 - \frac{\sum_{i=1}^{N}\sum_{j=1}^{C} y_{i,j}\, p_{i,j} + \epsilon}{\sum_{i=1}^{N}\sum_{j=1}^{C}\left( y_{i,j} + p_{i,j} - y_{i,j}\, p_{i,j} \right) + \epsilon}.
\end{equation}

\section{Extended Results: Cervical Spine VOI Detection}
\label{sec:S5}

This section presents the complete quantitative results for the cervical spine Region of Interest (ROI) detection experiments. All models were evaluated using 5-fold cross-validation on the subset of 87 patients with ground-truth segmentation masks.

\subsection{Quantitative Comparison of Detection Models}
\label{sec:S5_1}

Table \ref{tab:model_comparison_projections} provides the detailed performance metrics (Sagittal, Coronal, Axial, and 3D Mean IoU) for all evaluated object detection architectures across four different projection techniques.

\begin{table}[htbp]
    \centering
    \small
    \caption{Model Performance Comparison for Cervical Spine Detection across different projection types.}
    \label{tab:model_comparison_projections}
    \setlength{\tabcolsep}{6pt}
    \begin{tabular}{@{} l l c c c c @{}}
        \toprule
        Projection & Model & Sagittal mIoU & Coronal mIoU & Axial mIoU & 3D mIoU \\
        \midrule
        \multirow{11}{*}{Max} & YOLOv8m & 87.01 & 96.33 & 92.53 & 88.67 \\
        & YOLOv8l & 87.33 & 96.47 & 89.31 & 87.10 \\
        & YOLOv8x & 85.09 & 94.94 & 92.58 & 87.50 \\
        & YOLOv5m & 86.57 & 96.11 & 93.76 & 89.57 \\
        & YOLOv5l & 90.30 & 96.15 & 90.48 & 90.04 \\
        & YOLOv5x & 89.22 & 96.14 & 92.13 & 90.16 \\
        & RTDETR-l & 89.83 & 92.29 & 89.55 & 87.59 \\
        & RTDETR-x & 88.70 & 90.94 & 90.04 & 86.93 \\
        & YOLOv9c & 86.88 & 95.18 & 89.50 & 87.45 \\
        & YOLOv10m & 86.34 & 94.07 & 88.94 & 86.81 \\
        & YOLOv10x & 85.55 & 93.88 & 92.45 & 87.69 \\
        \midrule
        \multirow{11}{*}{Mean} & YOLOv8m & 90.56 & 94.90 & 92.34 & 90.46 \\
        & YOLOv8l & 88.69 & 95.17 & 90.94 & 88.85 \\
        & YOLOv8x & 92.15 & 95.37 & 91.80 & 91.40 \\
        & YOLOv5m & 89.39 & 94.88 & 91.66 & 89.73 \\
        & YOLOv5l & 90.48 & 95.36 & 93.03 & 91.31 \\
        & YOLOv5x & 90.09 & 96.17 & 92.61 & 90.78 \\
        & RTDETR-l & 90.70 & 91.35 & 87.44 & 87.36 \\
        & RTDETR-x & 90.85 & 88.68 & 88.64 & 86.72 \\
        & YOLOv9c & 90.64 & 93.55 & 90.37 & 89.22 \\
        & YOLOv10m & 88.22 & 94.02 & 90.50 & 89.05 \\
        & YOLOv10x & 89.31 & 95.03 & 90.47 & 89.07 \\
        \midrule
        \multirow{11}{*}{\textbf{Variance}} & YOLOv8m & 95.50 & 95.96 & 92.88 & 93.33 \\
        & YOLOv8l & 95.99 & 96.41 & 93.17 & 93.84 \\
        & \textbf{YOLOv8x} & 95.64 & 96.35 & 94.79 & \textbf{94.45} \\
        & YOLOv5m & 95.64 & 96.11 & 93.66 & 93.97 \\
        & YOLOv5l & 94.89 & 96.09 & 94.40 & 93.89 \\
        & YOLOv5x & 95.30 & 96.79 & 93.05 & 93.75 \\
        & RTDETR-l & 92.29 & 92.76 & 89.91 & 89.95 \\
        & RTDETR-x & 91.53 & 91.74 & 89.81 & 89.21 \\
        & YOLOv9c & 93.45 & 95.10 & 93.04 & 92.25 \\
        & YOLOv10m & 94.61 & 95.09 & 92.21 & 92.45 \\
        & YOLOv10x & 93.17 & 95.06 & 92.77 & 92.06 \\
        \midrule
        \multirow{11}{*}{Grad} & YOLOv8m & 88.92 & 96.06 & 91.93 & 89.94 \\
        & YOLOv8l & 86.93 & 95.83 & 92.97 & 88.58 \\
        & YOLOv8x & 88.94 & 95.66 & 89.66 & 89.22 \\
        & YOLOv5m & 87.34 & 95.77 & 89.72 & 88.15 \\
        & YOLOv5l & 87.93 & 95.55 & 91.92 & 89.39 \\
        & YOLOv5x & 88.06 & 95.93 & 93.01 & 89.84 \\
        & RTDETR-l & 89.08 & 92.88 & 88.62 & 87.23 \\
        & RTDETR-x & 89.50 & 92.69 & 87.40 & 87.72 \\
        & YOLOv9c & 85.48 & 94.18 & 90.97 & 87.29 \\
        & YOLOv10m & 89.32 & 92.91 & 91.99 & 89.53 \\
        & YOLOv10x & 81.86 & 95.26 & 89.41 & 84.37 \\
        \bottomrule
    \end{tabular}
\end{table}

\begin{figure}[htbp]
    \centering
    \includegraphics[width=\textwidth]{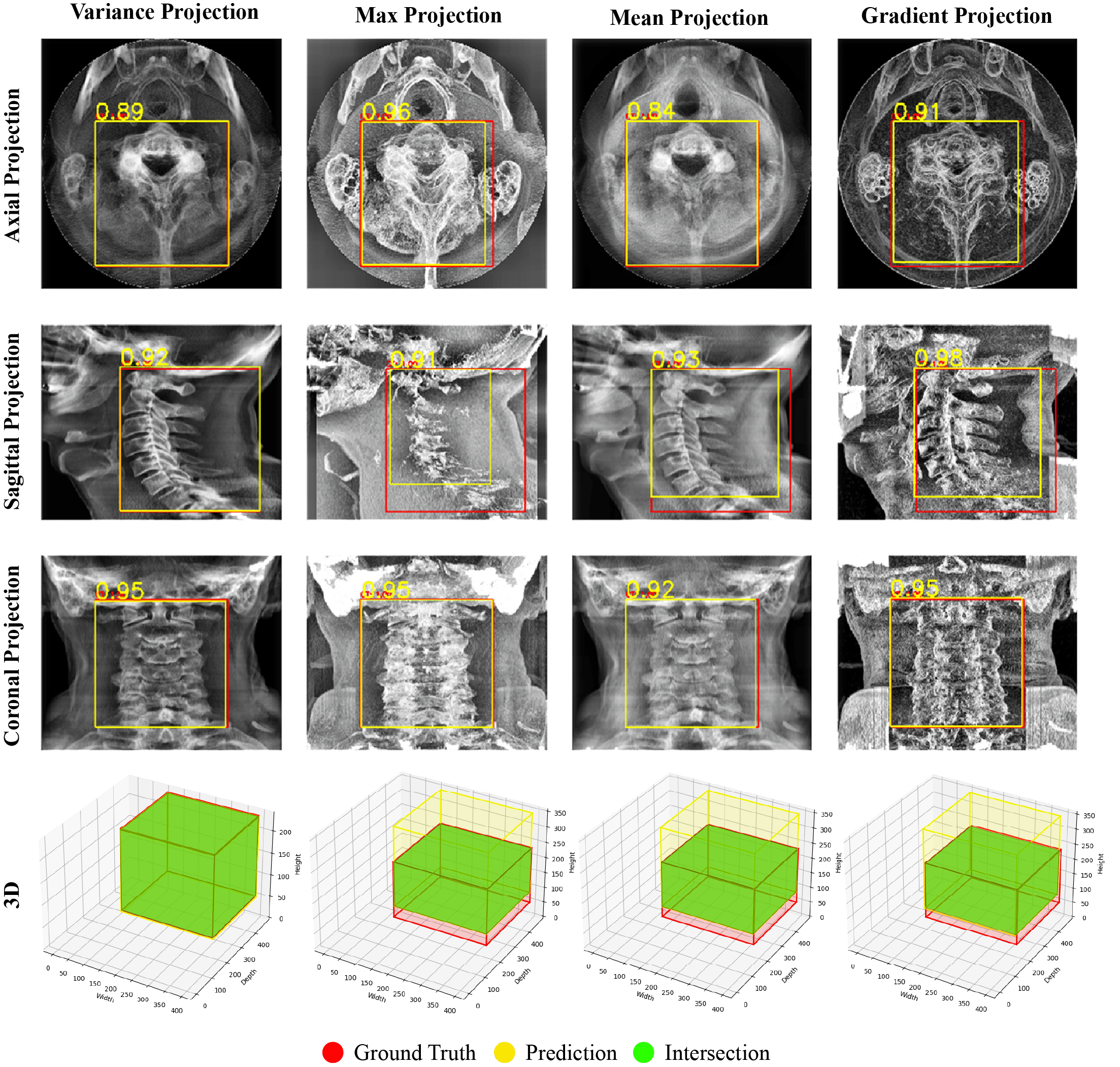}
    \caption{Cervical Spine location predictions of YOLOv8x across all projections.}
    \label{fig:projection_accuracy}
\end{figure}

\begin{figure}[htbp]
    \centering
    \includegraphics[width=\textwidth]{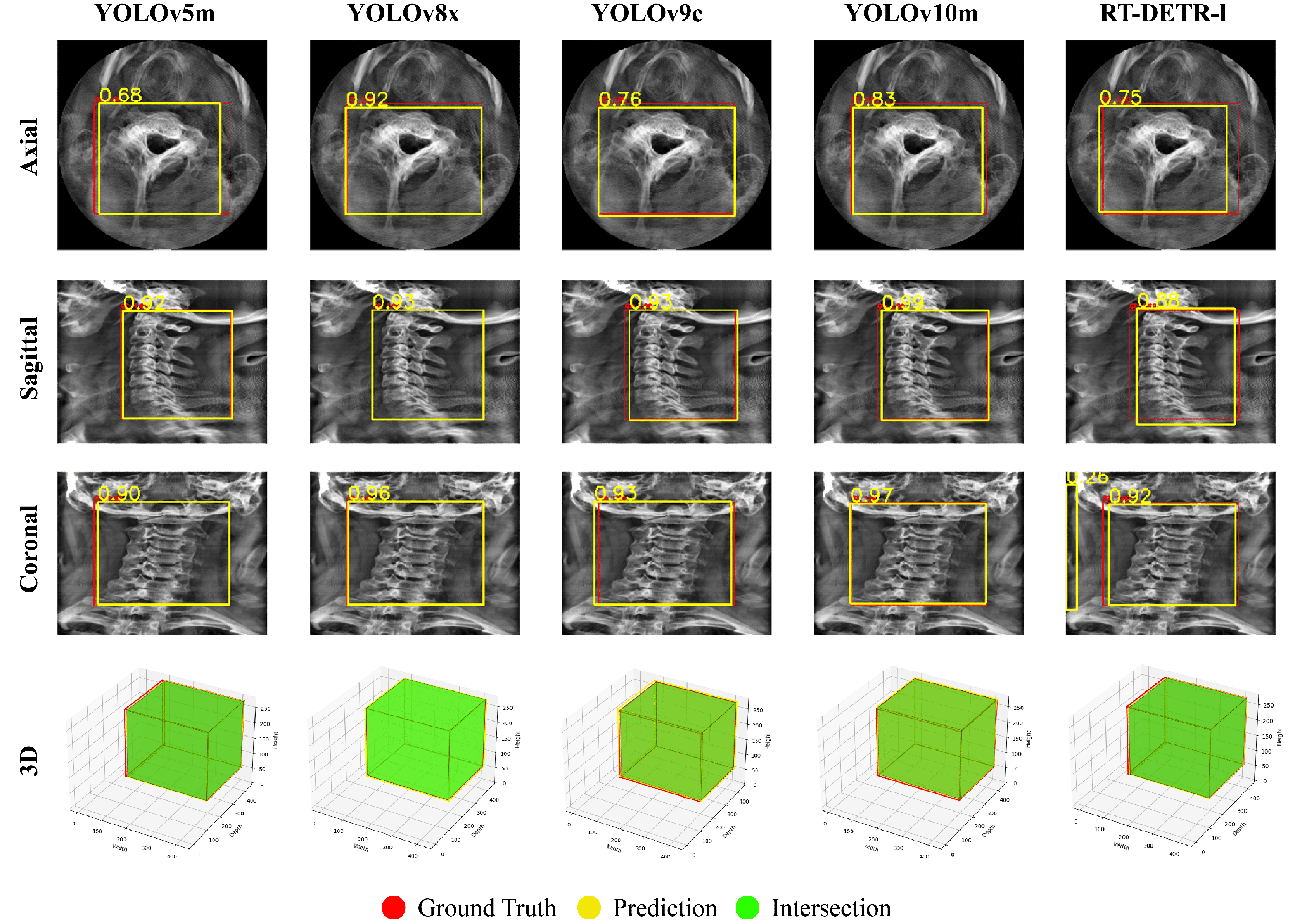}
    \caption{Qualitative spine localization by the top-performing detectors on the variance projection.}
    \label{fig:roi_qualitative_results}
\end{figure}

The variance projection gave the highest 3D mIoU for all 11 detectors in Table~\ref{tab:model_comparison_projections}. Figure~\ref{fig:projection_accuracy} shows YOLOv8x predictions on each projection for representative cases. In these examples, soft tissue around the spine is more prominent in the maximum and mean projections. Figure~\ref{fig:roi_qualitative_results} compares the top-performing detectors on the variance projection, where the YOLOv8 and YOLOv5 variants localize the spine consistently.

\subsection{Ablation Study: Effect of Interpolation}
\label{sec:S5_2}

We investigated the impact of slice interpolation on detection performance. Table \ref{tab:interpolation_effect} summarizes the results. Interpolation increased the 3D mIoU for 8 of the 11 models, including the best-performing model, YOLOv8x, which improved from 93.46\% to 94.45\%. The other three models changed by less than 0.5 points.

\begin{table}[htbp]
    \centering
    \small
    \caption{Effect of Interpolation on Performance.}
    \label{tab:interpolation_effect}
    \setlength{\tabcolsep}{4pt}
    \begin{tabular}{@{} l l c c c c c c c @{}}
        \toprule
        \multirow{2}{*}{Interp.} & \multirow{2}{*}{Model} & \multicolumn{2}{c}{Sagittal} & \multicolumn{2}{c}{Coronal} & \multicolumn{2}{c}{Axial} & 3D \\
        \cmidrule(lr){3-4} \cmidrule(lr){5-6} \cmidrule(lr){7-8}
        & & mAP & mIoU & mAP & mIoU & mAP & mIoU & mIoU \\
        \midrule
        \multirow{11}{*}{No} & YOLOv8m & 95.56 & 95.06 & 96.79 & 95.99 & 55.65 & 93.01 & 93.47 \\
        & YOLOv8l & 94.65 & 94.51 & 96.44 & 95.97 & 53.64 & 92.58 & 93.01 \\
        & YOLOv8x & 95.73 & 95.06 & 96.61 & 95.96 & 56.08 & 93.20 & 93.46 \\
        & YOLOv5m & 95.74 & 94.93 & 96.25 & 95.75 & 55.25 & 94.32 & 93.81 \\
        & YOLOv5l & 96.00 & 95.27 & 94.41 & 95.04 & 58.81 & 93.53 & 93.21 \\
        & YOLOv5x & 95.00 & 94.61 & 97.17 & 95.76 & 58.73 & 93.86 & 93.35 \\
        & RTDETR-l & 87.94 & 92.43 & 89.90 & 93.63 & 44.88 & 86.92 & 88.64 \\
        & RTDETR-x & 86.34 & 91.44 & 86.73 & 92.11 & 48.44 & 85.19 & 87.30 \\
        & YOLOv9c & 94.65 & 94.62 & 93.60 & 94.75 & 52.48 & 91.82 & 92.30 \\
        & YOLOv10m & 93.01 & 94.32 & 93.93 & 94.99 & 54.58 & 93.08 & 92.89 \\
        & YOLOv10x & 88.51 & 93.11 & 94.85 & 95.53 & 52.15 & 91.21 & 91.38 \\
        \midrule
        \multirow{11}{*}{\textbf{Yes}} & YOLOv8m & 96.78 & 95.50 & 96.51 & 95.96 & 61.26 & 92.88 & 93.33 \\
        & YOLOv8l & 97.19 & 95.99 & 96.98 & 96.41 & 59.53 & 93.17 & 93.84 \\
        & \textbf{YOLOv8x} & 95.83 & 95.64 & 97.38 & 96.35 & 60.26 & 94.79 & \textbf{94.45} \\
        & YOLOv5m & 97.34 & 95.64 & 96.78 & 96.11 & 59.07 & 93.66 & 93.97 \\
        & YOLOv5l & 95.50 & 94.89 & 96.68 & 96.09 & 59.30 & 94.40 & 93.89 \\
        & YOLOv5x & 95.97 & 95.30 & 98.50 & 96.79 & 60.95 & 93.05 & 93.75 \\
        & RTDETR-l & 87.95 & 92.29 & 88.93 & 92.76 & 53.68 & 89.91 & 89.95 \\
        & RTDETR-x & 86.98 & 91.53 & 84.29 & 91.74 & 55.32 & 89.81 & 89.21 \\
        & YOLOv9c & 92.13 & 93.45 & 94.76 & 95.10 & 58.12 & 93.04 & 92.25 \\
        & YOLOv10m & 93.24 & 94.61 & 93.71 & 95.09 & 54.17 & 92.21 & 92.45 \\
        & YOLOv10x & 90.16 & 93.17 & 93.36 & 95.06 & 58.35 & 92.77 & 92.06 \\
        \bottomrule
    \end{tabular}
\end{table}

\section{Extended Results: Multi-label Segmentation}
\label{sec:S6}

This section reports additional results for the multi-label vertebra segmentation stage, namely the per-vertebra comparison of segmentation architectures, the comparison of projection operators, and the preprocessing ablations.

\subsection{Comparative Analysis of Segmentation Models}
\label{sec:S6_1}

\begin{figure}[htbp]
    \centering
    \includegraphics[width=\textwidth]{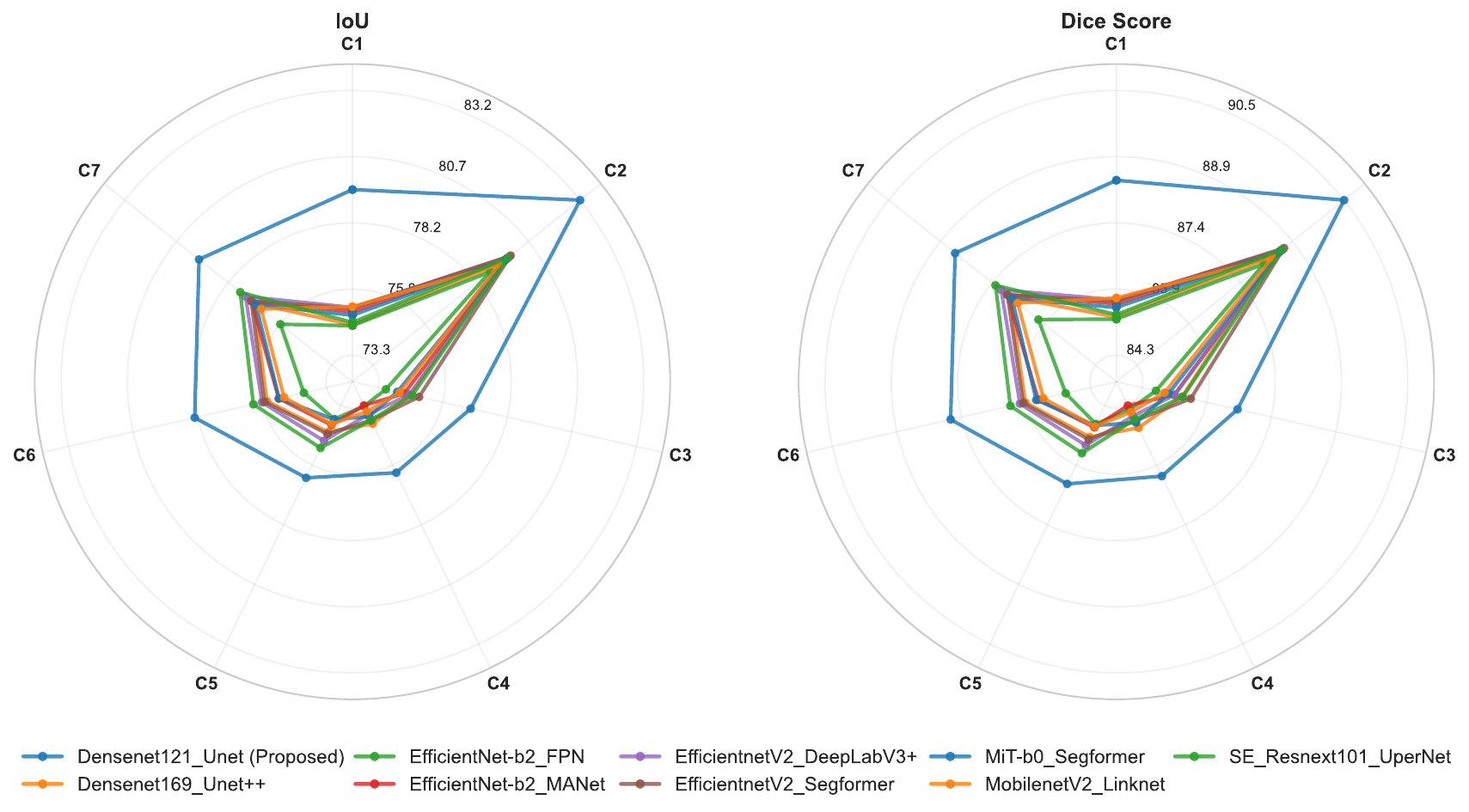}
    \caption{Vertebra-wise performance comparison radar plot.}
    \label{fig:radar_plot}
\end{figure}

\begin{figure}[htbp]
    \centering
    \includegraphics[width=\textwidth]{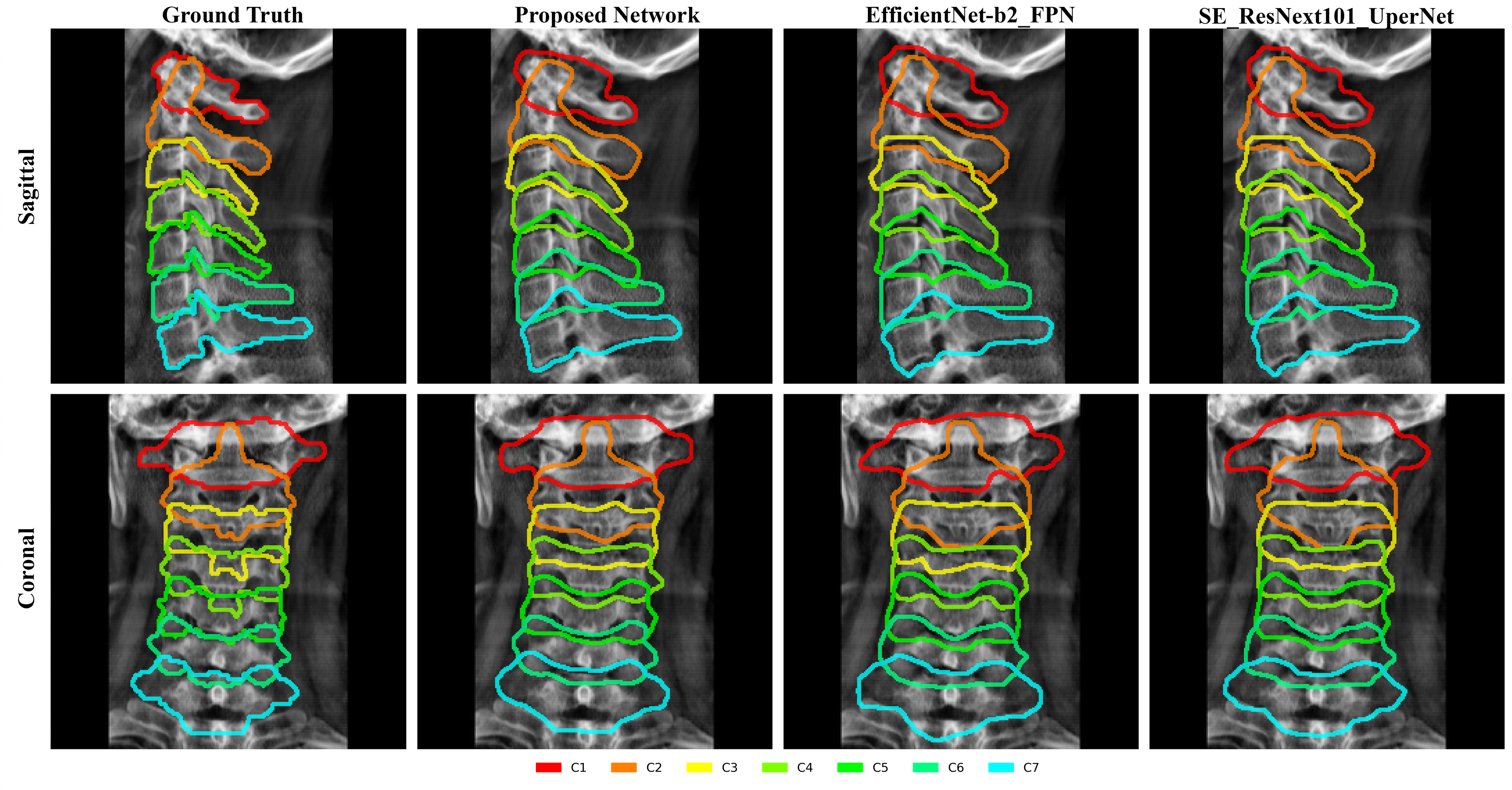}
    \caption{Multi-label vertebra masks predicted by the compared segmentation architectures on the energy projection.}
    \label{fig:seg_qualitative_results}
\end{figure}

Figure~\ref{fig:radar_plot} compares the IoU and Dice score of each segmentation architecture separately for C1--C7. The proposed DenseNet121-Unet (blue) has the highest scores at every cervical level, ahead of baselines such as SegFormer and DeepLabV3+. Several comparison models score markedly lower at C1, whereas the DenseNet121-Unet maintains its accuracy there, a difference that may relate to the distinctive ring shape of the atlas. All architectures score lower on the mid-cervical vertebrae (C3--C5), which are small and similar in shape, and the DenseNet121-Unet remains highest in this region as well. Figure~\ref{fig:seg_qualitative_results} shows example masks from each architecture on the energy projection, in which the DenseNet121-Unet masks are more compact around overlapping vertebral structures.

\subsection{Comprehensive Benchmark of Projection Techniques}
\label{sec:S6_2}

To select the 2D representation for segmentation, we compared 23 projection operators (defined in Section~S3). Table~\ref{tab:projection_benchmark} summarizes the results, with the composite score defined in Section~3.7 of the main manuscript. The energy projection achieved the best value in every column, including the highest composite score (1.48) and the lowest average HD95 (4.50 mm), although its lead in average Dice over the next operator, the standard deviation projection, was 1.07 points.

\begin{table}[H]
    \centering
    \small
    \caption{Performance comparison among different projections.}
    \label{tab:projection_benchmark}
    \setlength{\tabcolsep}{4pt}
    \begin{tabular}{@{} l ccc ccc ccc c @{}}
        \toprule
        \multirow{2}{*}{Projection} & \multicolumn{3}{c}{Sagittal} & \multicolumn{3}{c}{Coronal} & \multicolumn{4}{c}{Average} \\
        \cmidrule(lr){2-4} \cmidrule(lr){5-7} \cmidrule(lr){8-11}
        & IoU↑ & Dice↑ & HD95↓ & IoU↑ & Dice↑ & HD95↓ & IoU↑ & Dice↑ & HD95↓ & Score↑ \\
        \midrule
        Diffusion & 73.27 & 84.36 & 6.50 & 76.84 & 86.77 & 5.15 & 75.06 & 85.57 & 5.82 & 1.21 \\
        Difference & 71.74 & 83.07 & 8.82 & 75.28 & 85.70 & 6.65 & 73.51 & 84.38 & 7.74 & 0.91 \\
        Edge & 64.39 & 77.93 & 10.14 & 69.03 & 81.49 & 9.73 & 66.71 & 79.71 & 9.94 & 0.44 \\
        Energy & \textbf{78.53} & \textbf{87.93} & \textbf{4.17} & \textbf{78.36} & \textbf{87.79} & \textbf{4.83} & \textbf{78.45} & \textbf{87.86} & \textbf{4.50} & \textbf{1.48} \\
        Frangi & 69.32 & 81.58 & 8.03 & 72.66 & 84.08 & 6.60 & 70.99 & 82.83 & 7.31 & 0.91 \\
        Gabor & 71.08 & 82.76 & 7.94 & 74.25 & 84.97 & 6.59 & 72.67 & 83.86 & 7.26 & 0.96 \\
        Gradient & 62.17 & 76.36 & 11.44 & 70.66 & 82.65 & 7.60 & 66.41 & 79.51 & 9.52 & 0.50 \\
        Hessian & 60.94 & 75.39 & 11.70 & 68.24 & 80.91 & 9.40 & 64.59 & 78.15 & 10.55 & 0.31 \\
        Inversion & 76.29 & 86.50 & 5.03 & 76.27 & 86.41 & 5.16 & 76.28 & 86.46 & 5.09 & 1.35 \\
        Kurt & 69.85 & 81.88 & 9.85 & 75.35 & 85.89 & 5.98 & 72.60 & 83.89 & 7.92 & 0.86 \\
        Max Gradient & 66.44 & 79.31 & 9.60 & 70.84 & 82.83 & 7.96 & 68.64 & 81.07 & 8.78 & 0.65 \\
        Median & 67.13 & 80.04 & 8.27 & 72.23 & 83.76 & 6.74 & 69.68 & 81.90 & 7.50 & 0.86 \\
        Nonlinear & 76.95 & 86.76 & 5.10 & 75.41 & 85.74 & 6.19 & 76.18 & 86.25 & 5.65 & 1.26 \\
        Percentile Range & 70.91 & 82.45 & 8.75 & 76.17 & 86.34 & 5.46 & 73.54 & 84.40 & 7.11 & 1.00 \\
        Skew & 73.82 & 84.60 & 7.17 & 74.39 & 85.14 & 7.47 & 74.11 & 84.87 & 7.32 & 0.98 \\
        Sobel & 72.93 & 84.16 & 6.63 & 73.55 & 84.62 & 6.30 & 73.24 & 84.39 & 6.47 & 1.08 \\
        Standardized & 71.58 & 83.17 & 7.94 & 76.08 & 86.25 & 5.47 & 73.83 & 84.71 & 6.71 & 1.06 \\
        Standard Deviation & 76.51 & 86.48 & 5.70 & 77.40 & 87.10 & 5.23 & 76.95 & 86.79 & 5.47 & 1.31 \\
        Texture Energy & 74.65 & 85.42 & 5.71 & 76.04 & 86.25 & 5.51 & 75.35 & 85.84 & 5.61 & 1.25 \\
        Total Variation & 74.33 & 84.95 & 6.37 & 75.53 & 85.88 & 5.94 & 74.93 & 85.42 & 6.16 & 1.16 \\
        Variance & 75.69 & 86.00 & 6.31 & 77.43 & 87.23 & 5.34 & 76.56 & 86.62 & 5.82 & 1.25 \\
        Wavelet & 72.67 & 83.96 & 6.23 & 74.36 & 85.06 & 6.85 & 73.52 & 84.51 & 6.54 & 1.08 \\
        Z-score & 73.08 & 84.03 & 8.10 & 77.89 & 87.50 & 4.85 & 75.49 & 85.76 & 6.48 & 1.13 \\
        \bottomrule
    \end{tabular}
\end{table}

\subsection{Impact of Preprocessing on Segmentation}
\label{sec:S6_3}

We examined two preprocessing steps, bone windowing and volume interpolation. Applying a bone window (level $400$ HU, width $1800$ HU, range $[-500, 1300]$ HU) before projection increased the average Dice score by 1.82 points and reduced the average HD95 from 6.09 to 4.50 mm (Table~\ref{tab:windowing_impact}).

\begin{table}[H]
    \centering
    \small
    \caption{Impact of windowing on segmentation performance.}
    \label{tab:windowing_impact}
    \begin{tabular}{@{} l ccc ccc ccc @{}}
        \toprule
        \multirow{2}{*}{Windowing} & \multicolumn{3}{c}{Sagittal} & \multicolumn{3}{c}{Coronal} & \multicolumn{3}{c}{Average} \\
        \cmidrule(lr){2-4} \cmidrule(lr){5-7} \cmidrule(lr){8-10}
        & IoU↑ & Dice↑ & HD95↓ & IoU↑ & Dice↑ & HD95↓ & IoU↑ & Dice↑ & HD95↓ \\
        \midrule
        N/A & 76.04 & 86.20 & 5.46 & 75.67 & 85.88 & 6.72 & 75.85 & 86.04 & 6.09 \\
        Bone Window & \textbf{78.53} & \textbf{87.93} & \textbf{4.17} & \textbf{78.36} & \textbf{87.79} & \textbf{4.83} & \textbf{78.45} & \textbf{87.86} & \textbf{4.50} \\
        \bottomrule
    \end{tabular}
\end{table}

\begin{table}[H]
    \centering
    \small
    \caption{Performance impact of volume interpolation.}
    \label{tab:interpolation_impact}
    \begin{tabular}{@{} l ccc ccc ccc @{}}
        \toprule
        \multirow{2}{*}{Interpolation} & \multicolumn{3}{c}{Sagittal} & \multicolumn{3}{c}{Coronal} & \multicolumn{3}{c}{Average} \\
        \cmidrule(lr){2-4} \cmidrule(lr){5-7} \cmidrule(lr){8-10}
        & IoU↑ & Dice↑ & HD95↓ & IoU↑ & Dice↑ & HD95↓ & IoU↑ & Dice↑ & HD95↓ \\
        \midrule
        N/A & 67.43 & 80.03 & 9.07 & 73.59 & 84.67 & 5.52 & 70.51 & 82.35 & 7.30 \\
        Yes & \textbf{78.53} & \textbf{87.93} & \textbf{4.17} & \textbf{78.36} & \textbf{87.79} & \textbf{4.83} & \textbf{78.45} & \textbf{87.86} & \textbf{4.50} \\
        \bottomrule
    \end{tabular}
\end{table}

Volume interpolation, used to handle the variable slice counts across scans, increased the average Dice score from 82.35\% to 87.86\% and reduced the average HD95 from 7.30 to 4.50 mm (Table~\ref{tab:interpolation_impact}).

\section{Extended Results: Fracture Classification}
\label{sec:S7}

\subsection{Ablation Study: Mixup Augmentation}
\label{sec:S7_1}

Table~\ref{tab:mixup_ablation} compares the 2.5D spatio-sequential network (slice stack) trained with and without Mixup augmentation. Mixup improved every reported metric, raising the F1-score from 61.61\% to 66.53\% and the ROC-AUC from 88.33\% to 90.20\%.

\begin{table}[H]
    \centering
    \small
    \caption{Ablation study on the usage of Mixup augmentation.}
    \label{tab:mixup_ablation}
    \begin{tabular}{@{} l cccccc @{}}
        \toprule
        Mixup Augmentation & Accuracy & Precision & Sensitivity & Specificity & F1-Score & ROC-AUC \\
        \midrule
        No  & 92.32 & 63.00 & 60.41 & 95.95 & 61.61 & 88.33 \\
        Yes & \textbf{93.55} & \textbf{72.16} & \textbf{62.33} & \textbf{97.09} & \textbf{66.53} & \textbf{90.20} \\

        \bottomrule
    \end{tabular}
\end{table}

\subsection{Ablation Study: Square Padding of Vertebra Volumes}
\label{sec:S7_2}

Table~\ref{tab:ablation_padding} reports the effect of square-padding the extracted vertebra volumes of interest before classification. Omitting the padding improved every reported metric, raising the F1-score from 61.52\% to 66.53\%, the ROC-AUC from 89.10\% to 90.20\% and the AUPRC from 68.65\% to 72.14\%, likely because padding lowers the effective resolution of the vertebra within the fixed-size input. The proposed configuration therefore omits square padding.
\begin{table}[H]
    \centering
    \small
    \caption{Ablation study on square padding of vertebra volumes of interest.}
    \label{tab:ablation_padding}
    \begin{tabular}{@{} l ccccccc @{}}
        \toprule
        Square Padding & Accuracy & Precision & Sensitivity & Specificity & F1-Score & ROC-AUC & AUPRC \\
        \midrule
        Yes & 92.64 & 66.95 & 57.65 & 96.63 & 61.52 & 89.10 & 68.65 \\
        No  & \textbf{93.55} & \textbf{72.16} & \textbf{62.33} & \textbf{97.09} & \textbf{66.53} & \textbf{90.20} & \textbf{72.14} \\
        \bottomrule
    \end{tabular}
\end{table}

\section{Extended Results: Mask-Source and Whole-Pipeline Comparisons}
\label{sec:S8}

This section extends the comparisons of Section~4.1 of the main manuscript, in which the proposed score-fusion ensemble is compared with the same classifiers applied to VOIs from 3D-segmentation masks and with the reproduced RSNA 2022 winning solution (single model). Every pipeline is scored on all 14,133 vertebrae of the 2,019 patients, and a vertebra without a prediction is scored as non-fractured (Section~3.9 of the main manuscript). $\Delta$ denotes the proposed ensemble minus the comparator, in percentage points.

\subsection{Full Metric Profiles}
\label{sec:S8_1}

Table~\ref{tab:S8_profiles} reports all seven metrics for each classifier and VOI source, including the threshold-dependent metrics that Section~4.1 of the main manuscript omits. These are read at a decision threshold of 0.50, or at the adaptive threshold for the patient-level values of the score-fusion ensembles, and therefore reflect each model's operating point as well as its discrimination (Section~4.3.4 of the main manuscript).

\begin{table}[H]
    \centering
    \footnotesize
    \setlength{\tabcolsep}{3pt}
    \caption{Full metric profiles of the classifiers on projection-derived masks and on 3D-segmentation masks, and of the reproduced RSNA 2022 winning solution (single model). Values are the mean $\pm$ standard deviation over the five cross-validation folds, in percent. Patient-level values use the If-Any rule for the single models and the adaptive average for the score-fusion ensembles.}
    \label{tab:S8_profiles}
    \resizebox{\textwidth}{!}{%
    \begin{tabular}{@{} l l ccccccc @{}}
        \toprule
        VOI source & Classifier & Accuracy & Precision & Sensitivity & Specificity & F1 & ROC-AUC & AUPRC \\
        \midrule
        \multicolumn{9}{@{}l}{\textit{Vertebra level}} \\
        \multirow{3}{*}{\makecell[l]{Projection-derived\\masks}} & Slice stack & 93.55 $\pm$ 1.18 & 72.16 $\pm$ 8.51 & 62.33 $\pm$ 4.31 & 97.09 $\pm$ 1.41 & 66.53 $\pm$ 3.12 & 90.20 $\pm$ 1.06 & 72.14 $\pm$ 2.15 \\
         & MIP stack & 93.70 $\pm$ 0.46 & 76.82 $\pm$ 8.11 & 56.55 $\pm$ 5.70 & 97.95 $\pm$ 1.11 & 64.61 $\pm$ 1.78 & 89.85 $\pm$ 1.20 & 71.89 $\pm$ 1.76 \\
         & Score fusion & 94.51 $\pm$ 0.69 & 84.16 $\pm$ 4.82 & 57.49 $\pm$ 4.30 & 98.74 $\pm$ 0.47 & 68.15 $\pm$ 2.80 & 91.63 $\pm$ 0.99 & 75.64 $\pm$ 1.37 \\
        \addlinespace[2pt]
        \multirow{3}{*}{\makecell[l]{3D-segmentation\\masks}} & Slice stack & 93.02 $\pm$ 1.04 & 69.17 $\pm$ 10.89 & 61.11 $\pm$ 4.48 & 96.65 $\pm$ 1.64 & 64.25 $\pm$ 2.28 & 88.94 $\pm$ 1.19 & 71.11 $\pm$ 1.20 \\
         & MIP stack & 94.13 $\pm$ 0.76 & 77.37 $\pm$ 4.97 & 60.67 $\pm$ 5.78 & 97.93 $\pm$ 0.70 & 67.80 $\pm$ 3.72 & 89.46 $\pm$ 0.88 & 72.69 $\pm$ 2.93 \\
         & Score fusion & 94.57 $\pm$ 0.56 & 80.92 $\pm$ 4.60 & 61.42 $\pm$ 3.29 & 98.35 $\pm$ 0.42 & 69.78 $\pm$ 3.12 & 90.37 $\pm$ 0.95 & 74.61 $\pm$ 2.27 \\
        \addlinespace[2pt]
        \makecell[l]{Reproduced RSNA 2022\\winning solution} & Single model & 94.04 $\pm$ 0.81 & 73.01 $\pm$ 5.71 & 66.85 $\pm$ 3.02 & 97.15 $\pm$ 0.81 & 69.68 $\pm$ 3.26 & 92.15 $\pm$ 1.76 & 76.19 $\pm$ 2.39 \\
        \midrule
        \multicolumn{9}{@{}l}{\textit{Patient level}} \\
        \multirow{3}{*}{\makecell[l]{Projection-derived\\masks}} & Slice stack & 82.67 $\pm$ 1.91 & 86.15 $\pm$ 5.19 & 76.18 $\pm$ 4.95 & 88.38 $\pm$ 5.25 & 80.66 $\pm$ 2.35 & 89.09 $\pm$ 1.09 & 90.42 $\pm$ 0.90 \\
         & MIP stack & 81.18 $\pm$ 1.22 & 88.87 $\pm$ 5.61 & 69.79 $\pm$ 5.71 & 91.64 $\pm$ 5.55 & 77.87 $\pm$ 1.46 & 89.15 $\pm$ 1.71 & 90.58 $\pm$ 1.17 \\
         & Score fusion & 83.06 $\pm$ 3.13 & 82.12 $\pm$ 4.64 & 82.51 $\pm$ 1.74 & 83.56 $\pm$ 4.75 & 82.28 $\pm$ 3.05 & 90.95 $\pm$ 1.47 & 92.00 $\pm$ 1.15 \\
        \addlinespace[2pt]
        \multirow{3}{*}{\makecell[l]{3D-segmentation\\masks}} & Slice stack & 79.79 $\pm$ 3.04 & 81.09 $\pm$ 8.35 & 76.67 $\pm$ 5.40 & 82.76 $\pm$ 9.24 & 78.38 $\pm$ 2.00 & 87.93 $\pm$ 2.13 & 89.67 $\pm$ 1.84 \\
         & MIP stack & 82.27 $\pm$ 2.72 & 86.35 $\pm$ 5.03 & 75.01 $\pm$ 4.31 & 88.72 $\pm$ 5.51 & 80.13 $\pm$ 2.47 & 88.91 $\pm$ 1.96 & 90.55 $\pm$ 1.84 \\
         & Score fusion & 78.90 $\pm$ 2.35 & 73.99 $\pm$ 2.96 & 86.02 $\pm$ 1.63 & 72.31 $\pm$ 4.88 & 79.52 $\pm$ 1.98 & 89.72 $\pm$ 1.99 & 91.36 $\pm$ 1.72 \\
        \addlinespace[2pt]
        \makecell[l]{Reproduced RSNA 2022\\winning solution} & Single model & 84.35 $\pm$ 1.18 & 88.06 $\pm$ 3.12 & 77.75 $\pm$ 2.85 & 90.41 $\pm$ 2.59 & 82.52 $\pm$ 1.58 & 90.75 $\pm$ 1.26 & 91.86 $\pm$ 0.57 \\
        \bottomrule
    \end{tabular}}
\end{table}

\subsection{Per-Level Comparisons}
\label{sec:S8_3}

\begin{figure}[htbp]
    \centering
    \includegraphics[width=\textwidth]{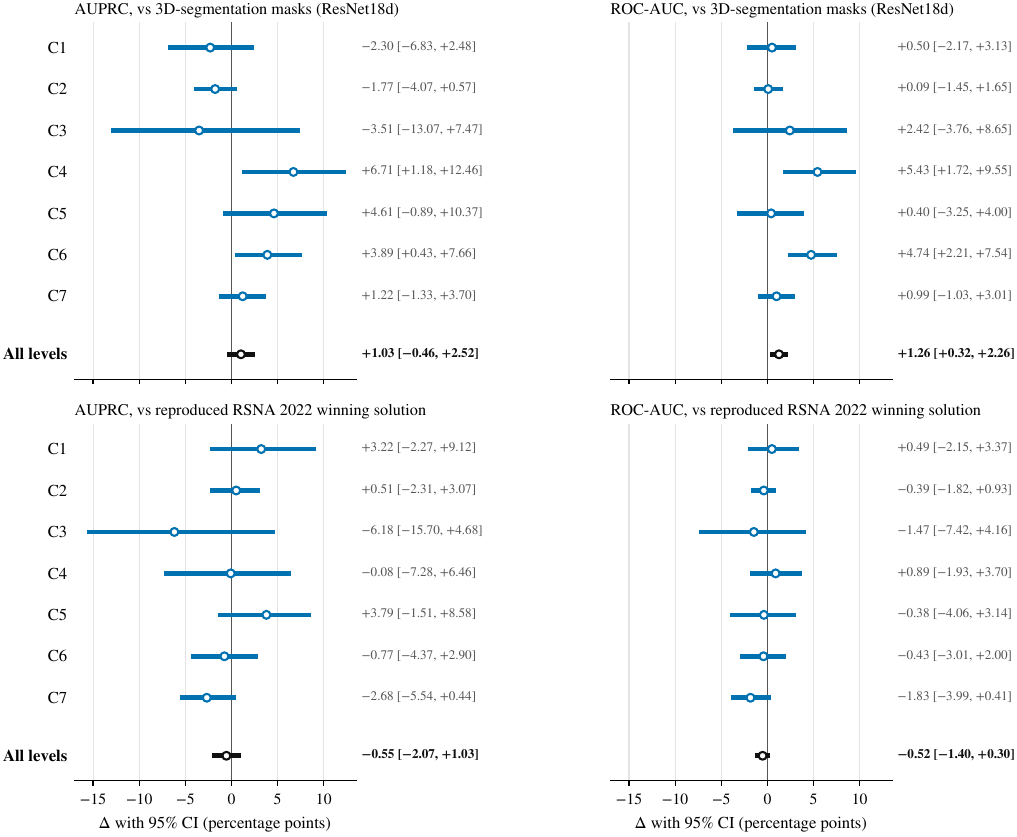}
    \caption{Per-level differences between the proposed score-fusion ensemble and each comparator, for AUPRC (left) and ROC-AUC (right). Each row gives $\Delta$ (proposed minus comparator, percentage points) with the patient-clustered paired bootstrap 95\% CI, and the bottom row of each panel pools all levels. The vertical line marks zero. Intervals are not adjusted for multiple comparisons.}
    \label{fig:S8_perlevel}
\end{figure}

Figure~\ref{fig:S8_perlevel} shows the per-level differences in AUPRC and ROC-AUC from the patient-clustered paired bootstrap against both comparators. Against the 3D-segmentation masks, the 95\% intervals exclude zero at C4 and C6 for both metrics, in favor of the projection-derived masks, and no level has a 95\% interval below zero. The AUPRC estimates at C1 to C3 lean toward the 3D-segmentation masks ($-2.30$, $-1.77$ and $-3.51$ points), with 95\% intervals that include zero and extend to $-6.83$, $-4.07$ and $-13.07$ points. Against the reproduced winning solution, whose pipeline differs in architecture and training as well as in how the VOIs are obtained, no 95\% interval excludes zero. The AUPRC estimates at C3 and C7 lean toward the winning solution ($-6.18$ and $-2.68$ points), with 95\% intervals that include zero.

\subsection{Vertebrae Without a Prediction}
\label{sec:S8_4}

Under the primary evaluation rule, a vertebra without a prediction is scored as non-fractured. This applies to 4 vertebrae for the projection-derived pipeline (one patient, none fractured), 7 for the reproduced winning solution (one patient, one fractured) and 27 for the 3D-segmentation masks (five patients, two fractured), 23 of which have an empty mask. The winning solution also had input crops but no saved prediction for 8 further vertebrae, which were predicted with the fold model for which the patient was a test patient. Table~\ref{tab:S8_missed} compares the pooled differences under this rule with those obtained after removing the union of the 27 missed vertebrae from every pipeline in the vertebra-level analyses, and the five patients who own them in the patient-level analyses. The differences change by at most 0.14 points, and the lower limits of the 95\% intervals, fold-wise or bootstrap, by at most 0.16 points. The only change in significance is in the fold-wise $t$-test, where the vertebra-level ROC-AUC difference against the 3D-segmentation masks moves from $p = 0.047$ to $p = 0.053$.

\begin{table}[H]
    \centering
    \small
    \caption{Pooled paired differences under the primary evaluation rule, in which a vertebra without a prediction is scored as non-fractured, and after removing the 27 vertebrae that any compared pipeline missed from the vertebra-level analyses, and the five patients who own them from the patient-level analyses. $\Delta$ is the proposed score-fusion ensemble minus the comparator, in percentage points, with the patient-clustered bootstrap 95\% CI.}
    \label{tab:S8_missed}
    \begin{tabular}{@{} l l c c @{}}
        \toprule
        Comparator & Level, metric & \makecell{Every vertebra scored\\(14,133 vertebrae)} & \makecell{Missed vertebrae removed\\(14,106 vertebrae)} \\
        \midrule
        \multirow{4}{*}{\makecell[l]{3D-segmentation\\masks}} & Vertebra AUPRC & +1.03 [$-$0.46, +2.52] & +1.00 [$-$0.45, +2.55]\\
         & Vertebra ROC-AUC & +1.26 [+0.32, +2.26] & +1.20 [+0.34, +2.15] \\
         & Patient AUPRC & +0.65 [$-$0.32, +1.67] & +0.60 [$-$0.32, +1.60] \\
         & Patient ROC-AUC & +1.23 [$-$0.04, +2.53] & +1.09 [$-$0.08, +2.41] \\
        \addlinespace[2pt]
        \multirow{4}{*}{\makecell[l]{Reproduced RSNA 2022\\winning solution}} & Vertebra AUPRC & $-$0.55 [$-$2.07, +1.03] & $-$0.61 [$-$2.23, +1.02] \\
         & Vertebra ROC-AUC & $-$0.52 [$-$1.40, +0.30] & $-$0.59 [$-$1.42, +0.29] \\
         & Patient AUPRC & +0.15 [$-$0.93, +1.20] & +0.12 [$-$0.91, +1.09] \\
         & Patient ROC-AUC & +0.19 [$-$1.01, +1.42] & +0.11 [$-$1.08, +1.24] \\
        \bottomrule
    \end{tabular}
\end{table}

\subsection{Segmentation by Fracture Status}
\label{sec:S8_5}

\begin{figure}[htbp]
    \centering
    \includegraphics[width=0.6\textwidth]{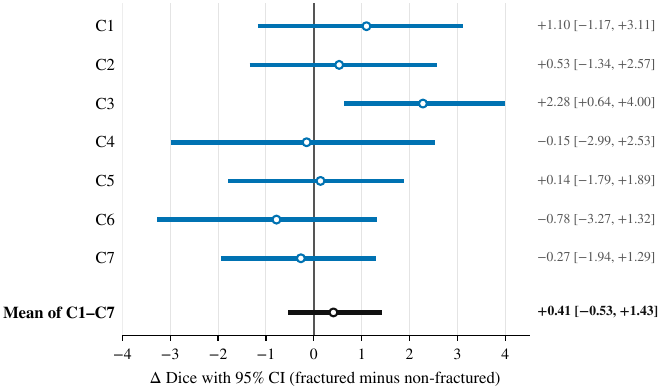}
    \caption{Dice of fractured minus non-fractured vertebrae on the 87-patient subset, per level and as the level-matched mean over C1--C7 (bottom row), with patient-clustered bootstrap 95\% CIs (patients resampled within each fold, 2,000 replicates). The vertical line marks zero.}
    \label{fig:S8_fracture}
\end{figure}

Section~4.2.2 of the main manuscript reports the Dice score of fractured and non-fractured vertebrae on the 87-patient subset. Figure~\ref{fig:S8_fracture} shows the per-level differences with their 95\% intervals.

\subsection{Ankylosed Vertebrae}
\label{sec:S8_6}

Three readers independently labeled each of the 609 cervical vertebrae of the 87 voxel-annotated patients as ankylosed or not ankylosed, and rated their confidence for each case. Ankylosis was defined broadly as a substantial bony connection between adjacent vertebrae, whatever its cause (for example ankylosing spondylitis, diffuse idiopathic skeletal hyperostosis, degenerative spondylosis with bridging osteophytes, ossification of the posterior longitudinal ligament or rheumatoid arthritis), and did not require the vertebrae to be completely fused or to appear as a single bone. Agreement among the three was moderate (Fleiss $\kappa = 0.584$ at the vertebra level and 0.580 at the patient level). Two readers agreed closely with each other ($\kappa = 0.844$), whereas the third called ankylosis far less often (66 of 607 assessable vertebrae, against 156 and 154 of 609) and was the lone dissenter on 89 of the 124 vertebrae on which the readers split. When all three readers rated a case with high confidence, they agreed on 97.2\% of its vertebrae. The label used here is the majority vote, which gives 141 ankylosed and 467 non-ankylosed vertebrae. One vertebra, assessed by only two readers who disagreed, was excluded.

\begin{figure}[htbp]
    \centering
    \includegraphics[width=0.6\textwidth]{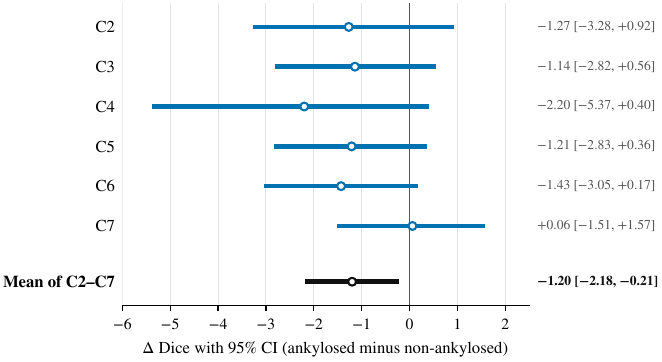}
    \caption{Dice of ankylosed minus non-ankylosed vertebrae (majority-vote labels) on the 87-patient subset, per level for C2--C7 and as the level-matched mean (bottom row), with patient-clustered bootstrap 95\% CIs (patients resampled within each fold, 2,000 replicates). C1 is not shown because only four of its vertebrae are ankylosed. The vertical line marks zero.}
    \label{fig:S8_ankylosis}
\end{figure}

Table~\ref{tab:S8_ankylosis_seg} compares the Dice of ankylosed and non-ankylosed vertebrae. Ankylosed vertebrae are segmented 1.20 to 1.45 Dice points lower, depending on how the levels are combined. Every 95\% interval lies below zero, and losses larger than 2.34 points are not supported by the data. No single level's 95\% interval excludes zero (Figure~\ref{fig:S8_ankylosis}).

Table~\ref{tab:S8_ankylosis_cls} compares fracture recognition by the proposed score-fusion ensemble on the same patients, each predicted by the classifier of the fold in which it was held out. No significant difference was detected between ankylosed and non-ankylosed vertebrae or between patients with and without an ankylosed level, although the intervals are wide because only 30 ankylosed vertebrae and 19 patients with an ankylosed level are fractured. Because fracture prevalence is higher among ankylosed vertebrae (21\% against 12\%), ROC-AUC, which does not depend on prevalence, is the more appropriate basis for comparing the two groups.

\begin{table}[H]
    \centering
    \small
        \caption{Dice of ankylosed and non-ankylosed vertebrae (majority-vote labels) on the 87-patient subset, in percent, computed as in Table~6 of the main manuscript. $\Delta$ is ankylosed minus non-ankylosed, in Dice points, with patient-clustered bootstrap 95\% CIs (patients resampled within each fold, 2,000 replicates). The level-matched mean averages C2--C7, because C1 has too few ankylosed vertebrae.}
    \label{tab:S8_ankylosis_seg}
    \begin{tabular}{@{} l c c c c @{}}
        \toprule
        Scope & \makecell{Vertebrae\\(ank. / non-ank.)} & Ankylosed & Non-ankylosed & $\Delta$ [95\% CI] \\
        \midrule
        Level-matched (C2--C7) & 137 / 384 & 86.74 & 87.94 & $-$1.20 [$-$2.18, $-$0.21] \\
        Pooled & 141 / 467 & 86.73 & 88.05 & $-$1.32 [$-$2.26, $-$0.37] \\
        Fold-wise & 141 / 467 & 86.57 & 88.02 & $-$1.45 [$-$2.34, $-$0.47] \\
        \bottomrule
    \end{tabular}
\end{table}

\begin{table}[H]
    \centering
    \small
    \caption{Fracture recognition by the proposed score-fusion ensemble on ankylosed and non-ankylosed vertebrae (majority-vote labels) and on patients with and without an ankylosed level, on the 87-patient subset, in percent. Each patient is predicted by the classifier of the fold in which it was held out. $\Delta$ is ankylosed minus non-ankylosed, in percentage points, with patient-clustered bootstrap 95\% CIs over the 87 patients (2,000 replicates).}
    \label{tab:S8_ankylosis_cls}
    \begin{tabular}{@{} l l c c c @{}}
        \toprule
        Level (fractured / total) & Metric & Ankylosed & Non-ankylosed & $\Delta$ [95\% CI] \\
        \midrule
        \multirow{2}{*}{\makecell[l]{Vertebra\\(30 / 141 against 55 / 467)}} & ROC-AUC & 90.24 & 94.15 & $-$3.91 [$-$12.82, +3.25] \\
         & AUPRC & 80.03 & 80.87 & $-$0.85 [$-$23.68, +17.07] \\
        \addlinespace[2pt]
        \multirow{2}{*}{\makecell[l]{Patient\\(19 / 36 against 21 / 51)}} & ROC-AUC & 93.81 & 95.24 & $-$1.43 [$-$11.65, +8.13] \\
         & AUPRC & 95.49 & 94.10 & +1.39 [$-$8.32, +11.57] \\
        \bottomrule
    \end{tabular}
\end{table}

\section{Per-Fold Results}
\label{sec:S9}

Tables~\ref{tab:S9_folds_vertebra} and \ref{tab:S9_folds_patient} list the results of each cross-validation fold for the classifiers on projection-derived masks and on 3D-segmentation masks, and for the reproduced RSNA 2022 winning solution. All three pipelines use the same five folds, so a fold of one model can be paired with the same fold of another, as in the fold-wise tests of Section~4.1 of the main manuscript.

\begin{table}[H]
    \centering
    \footnotesize
    \setlength{\tabcolsep}{4pt}
    \caption{Per-fold vertebra-level results of the main models, in percent. Fold means and standard deviations are given in Table~\ref{tab:S8_profiles}.}
    \label{tab:S9_folds_vertebra}
    \resizebox{\textwidth}{!}{%
    \begin{tabular}{@{} l l c ccccccc @{}}
        \toprule
        VOI source & Classifier & Fold & Accuracy & Precision & Sensitivity & Specificity & F1 & ROC-AUC & AUPRC \\
        \midrule
        \multirow{5}{*}{\makecell[l]{Projection-derived\\masks}} & Slice stack & 0 & 91.83 & 62.61 & 66.77 & 94.98 & 64.62 & 90.98 & 69.60 \\
         &  & 1 & 93.46 & 74.03 & 57.77 & 97.63 & 64.90 & 88.81 & 71.21 \\
         &  & 2 & 93.21 & 66.14 & 60.81 & 96.67 & 63.36 & 90.26 & 71.87 \\
         &  & 3 & 94.91 & 84.69 & 59.29 & 98.82 & 69.75 & 91.41 & 75.44 \\
         &  & 4 & 94.33 & 73.33 & 67.03 & 97.32 & 70.04 & 89.52 & 72.60 \\
        \midrule
        \multirow{5}{*}{\makecell[l]{Projection-derived\\masks}} & MIP stack & 0 & 93.53 & 81.82 & 54.11 & 98.49 & 65.14 & 90.26 & 72.52 \\
         &  & 1 & 93.74 & 84.80 & 48.99 & 98.97 & 62.10 & 88.85 & 70.15 \\
         &  & 2 & 93.00 & 63.64 & 64.10 & 96.09 & 63.87 & 89.42 & 70.85 \\
         &  & 3 & 94.17 & 76.26 & 59.64 & 97.96 & 66.93 & 91.76 & 74.64 \\
         &  & 4 & 94.04 & 77.61 & 55.91 & 98.23 & 65.00 & 88.97 & 71.29 \\
        \midrule
        \multirow{5}{*}{\makecell[l]{Projection-derived\\masks}} & Score fusion & 0 & 93.39 & 79.72 & 54.75 & 98.25 & 64.92 & 92.36 & 74.31 \\
         &  & 1 & 94.38 & 91.02 & 51.35 & 99.41 & 65.66 & 91.06 & 75.44 \\
         &  & 2 & 94.66 & 79.61 & 60.07 & 98.36 & 68.48 & 91.04 & 74.54 \\
         &  & 3 & 95.08 & 86.53 & 59.64 & 98.98 & 70.61 & 92.99 & 77.66 \\
         &  & 4 & 95.04 & 83.90 & 61.65 & 98.70 & 71.07 & 90.70 & 76.26 \\
        \midrule
        \multirow{5}{*}{\makecell[l]{3D-segmentation\\masks}} & Slice stack & 0 & 93.28 & 76.47 & 57.59 & 97.77 & 65.70 & 90.45 & 71.68 \\
         &  & 1 & 91.90 & 60.44 & 65.54 & 94.98 & 62.88 & 87.35 & 71.07 \\
         &  & 2 & 92.72 & 62.64 & 60.81 & 96.13 & 61.71 & 88.25 & 69.13 \\
         &  & 3 & 92.54 & 61.54 & 65.71 & 95.49 & 63.56 & 89.56 & 71.36 \\
         &  & 4 & 94.65 & 84.78 & 55.91 & 98.90 & 67.39 & 89.06 & 72.31 \\
        \midrule
        \multirow{5}{*}{\makecell[l]{3D-segmentation\\masks}} & MIP stack & 0 & 93.00 & 71.69 & 61.71 & 96.93 & 66.33 & 90.21 & 72.90 \\
         &  & 1 & 94.24 & 84.10 & 55.41 & 98.78 & 66.80 & 89.26 & 74.08 \\
         &  & 2 & 93.81 & 74.75 & 54.21 & 98.04 & 62.85 & 88.02 & 67.62 \\
         &  & 3 & 94.94 & 80.72 & 64.29 & 98.31 & 71.57 & 89.93 & 74.91 \\
         &  & 4 & 94.65 & 75.60 & 67.74 & 97.60 & 71.46 & 89.86 & 73.95 \\
        \midrule
        \multirow{5}{*}{\makecell[l]{3D-segmentation\\masks}} & Score fusion & 0 & 94.02 & 81.55 & 60.13 & 98.29 & 69.22 & 91.45 & 75.32 \\
         &  & 1 & 94.80 & 83.71 & 62.50 & 98.58 & 71.57 & 89.42 & 75.08 \\
         &  & 2 & 93.99 & 74.88 & 56.78 & 97.96 & 64.58 & 89.33 & 70.70 \\
         &  & 3 & 94.77 & 77.97 & 65.71 & 97.96 & 71.32 & 90.68 & 75.35 \\
         &  & 4 & 95.29 & 86.50 & 62.01 & 98.94 & 72.23 & 90.98 & 76.62 \\
        \midrule
        \multirow{5}{*}{\makecell[l]{Reproduced RSNA 2022\\winning solution}} & Single model & 0 & 93.81 & 75.78 & 63.19 & 97.54 & 68.92 & 92.46 & 76.89 \\
         &  & 1 & 94.19 & 75.49 & 65.19 & 97.55 & 69.96 & 91.80 & 75.67 \\
         &  & 2 & 93.74 & 69.85 & 66.67 & 96.78 & 68.22 & 90.78 & 73.76 \\
         &  & 3 & 95.33 & 79.20 & 71.22 & 97.96 & 75.00 & 95.01 & 79.93 \\
         &  & 4 & 93.14 & 64.75 & 67.97 & 95.92 & 66.32 & 90.68 & 74.72 \\
        \bottomrule
    \end{tabular}}
\end{table}

\begin{table}[H]
    \centering
    \footnotesize
    \setlength{\tabcolsep}{4pt}
    \caption{Per-fold patient-level results of the main models, in percent. Fold means and standard deviations are given in Table~\ref{tab:S8_profiles}. Patient-level values use the If-Any rule for the single models and the adaptive average for the score-fusion ensembles.}
    \label{tab:S9_folds_patient}
    \resizebox{\textwidth}{!}{%
    \begin{tabular}{@{} l l c ccccccc @{}}
        \toprule
        VOI source & Classifier & Fold & Accuracy & Precision & Sensitivity & Specificity & F1 & ROC-AUC & AUPRC \\
        \midrule
        \multirow{5}{*}{\makecell[l]{Projection-derived\\masks}} & Slice stack & 0 & 82.43 & 83.74 & 81.73 & 83.16 & 82.73 & 88.79 & 90.18 \\
         &  & 1 & 84.16 & 87.35 & 77.13 & 90.28 & 81.92 & 90.48 & 91.80 \\
         &  & 2 & 79.46 & 80.35 & 73.94 & 84.26 & 77.01 & 87.86 & 89.33 \\
         &  & 3 & 83.42 & 94.24 & 68.95 & 96.26 & 79.64 & 88.39 & 90.61 \\
         &  & 4 & 83.87 & 85.06 & 79.14 & 87.96 & 81.99 & 89.92 & 90.16 \\
        \midrule
        \multirow{5}{*}{\makecell[l]{Projection-derived\\masks}} & MIP stack & 0 & 79.70 & 92.57 & 65.87 & 94.39 & 76.97 & 88.69 & 90.72 \\
         &  & 1 & 80.94 & 91.73 & 64.89 & 94.91 & 76.01 & 88.36 & 89.72 \\
         &  & 2 & 80.45 & 79.14 & 78.72 & 81.94 & 78.93 & 86.99 & 89.21 \\
         &  & 3 & 82.67 & 88.96 & 72.11 & 92.06 & 79.65 & 91.33 & 92.21 \\
         &  & 4 & 82.13 & 91.97 & 67.38 & 94.91 & 77.78 & 90.37 & 91.05 \\
        \midrule
        \multirow{5}{*}{\makecell[l]{Projection-derived\\masks}} & Score fusion & 0 & 83.17 & 83.98 & 83.17 & 83.16 & 83.57 & 90.65 & 91.86 \\
         &  & 1 & 82.92 & 81.48 & 81.91 & 83.80 & 81.70 & 91.85 & 92.79 \\
         &  & 2 & 77.97 & 74.38 & 80.32 & 75.93 & 77.24 & 88.49 & 90.11 \\
         &  & 3 & 85.40 & 86.19 & 82.11 & 88.32 & 84.10 & 91.79 & 93.00 \\
         &  & 4 & 85.86 & 84.57 & 85.03 & 86.57 & 84.80 & 91.96 & 92.27 \\
        \midrule
        \multirow{5}{*}{\makecell[l]{3D-segmentation\\masks}} & Slice stack & 0 & 78.96 & 84.75 & 72.12 & 86.22 & 77.92 & 85.53 & 88.29 \\
         &  & 1 & 75.99 & 70.59 & 82.98 & 69.91 & 76.28 & 88.70 & 90.29 \\
         &  & 2 & 78.71 & 78.02 & 75.53 & 81.48 & 76.76 & 85.95 & 87.45 \\
         &  & 3 & 81.19 & 79.08 & 81.58 & 80.84 & 80.31 & 90.58 & 92.12 \\
         &  & 4 & 84.12 & 93.01 & 71.12 & 95.37 & 80.61 & 88.86 & 90.22 \\
        \midrule
        \multirow{5}{*}{\makecell[l]{3D-segmentation\\masks}} & MIP stack & 0 & 78.71 & 81.44 & 75.96 & 81.63 & 78.61 & 87.32 & 89.68 \\
         &  & 1 & 84.16 & 93.06 & 71.28 & 95.37 & 80.72 & 91.33 & 92.51 \\
         &  & 2 & 80.20 & 84.62 & 70.21 & 88.89 & 76.74 & 86.73 & 88.39 \\
         &  & 3 & 85.15 & 90.12 & 76.84 & 92.52 & 82.95 & 90.41 & 92.43 \\
         &  & 4 & 83.13 & 82.51 & 80.75 & 85.19 & 81.62 & 88.76 & 89.72 \\
        \midrule
        \multirow{5}{*}{\makecell[l]{3D-segmentation\\masks}} & Score fusion & 0 & 77.72 & 73.79 & 87.98 & 66.84 & 80.26 & 88.23 & 90.35 \\
         &  & 1 & 77.48 & 71.18 & 86.70 & 69.44 & 78.18 & 90.85 & 92.40 \\
         &  & 2 & 76.49 & 71.04 & 83.51 & 70.37 & 76.77 & 87.05 & 88.99 \\
         &  & 3 & 81.68 & 77.62 & 85.79 & 78.04 & 81.50 & 91.81 & 93.39 \\
         &  & 4 & 81.14 & 76.30 & 86.10 & 76.85 & 80.90 & 90.64 & 91.66 \\
        \midrule
        \multirow{5}{*}{\makecell[l]{Reproduced RSNA 2022\\winning solution}} & Single model & 0 & 83.42 & 89.87 & 73.58 & 92.42 & 80.91 & 90.72 & 91.73 \\
         &  & 1 & 84.62 & 91.07 & 76.50 & 92.61 & 83.15 & 89.20 & 91.50 \\
         &  & 2 & 84.41 & 86.98 & 78.19 & 89.81 & 82.35 & 91.76 & 92.54 \\
         &  & 3 & 86.14 & 89.20 & 80.93 & 90.95 & 84.86 & 92.21 & 92.32 \\
         &  & 4 & 83.17 & 83.15 & 79.57 & 86.24 & 81.32 & 89.88 & 91.18 \\
        \bottomrule
    \end{tabular}}
\end{table}

\section{Additional Implementation Details}
\label{sec:S10}

\subsection{Data Augmentation Parameters}
\label{sec:S10_1}

To improve generalization of the 2.5D spatio-sequential network, we applied the augmentation pipeline summarized in Section~3.6.1 of the main manuscript. The complete set of operations and their parameters is listed in Table \ref{tab:augmentation_params}. Image-level operations were applied with the \texttt{albumentations} library, while slice-order permutation and MixUp were applied at the sequence level during batch construction.

\begin{table}[H]
    \centering
    \small
    \setlength{\tabcolsep}{5pt}
    \begin{threeparttable}
    \caption{Data augmentation operations and parameters used for fracture classification.}
    \label{tab:augmentation_params}
    \begin{tabular}{@{} l l l c @{}}
        \toprule
        Category & Operation & Key parameters & Probability \\
        \midrule
        Geometric                       & Horizontal flip               & --                                      & 0.5 \\
        Geometric                       & Vertical flip                 & --                                      & 0.5 \\
        Geometric                       & Transpose                     & --                                      & 0.5 \\
        Geometric                       & Shift-scale-rotate            & shift 0.3, scale 0.3, rotate $45^\circ$ & 0.7 \\
        Photometric                     & Brightness-contrast           & brightness limit 0.1                    & 0.7 \\
        Spatial distortion$^{\dagger}$  & Optical or grid distortion    & distort limit 1.0; grid 5 steps         & 0.5 \\
        Noise and blur$^{\ddagger}$     & Motion, median, or Gaussian blur & kernel 3--5                          & 0.5 \\
        Noise and blur$^{\ddagger}$     & Gaussian noise                & variance 3.0--9.0                       & 0.5 \\
        Region dropout                  & Coarse dropout                & 1 hole, up to $128\times128$ px         & 0.5 \\
        Slice-order permutation         & Random slice and stack reordering & two variants                        & 0.2 / 0.3 \\
        Sample mixing                   & MixUp                         & --                                      & 0.55 \\
        \bottomrule
    \end{tabular}
    \begin{tablenotes}[flushleft]
      \small
      \item $\dagger$, $\ddagger$ each denote a mutually exclusive group, from which a single operation is sampled at the listed probability.
    \end{tablenotes}
    \end{threeparttable}
\end{table}

\subsection{Network Layer Dimensions}
\label{sec:S10_2}

Table \ref{tab:layer_dims} provides the complete layer-by-layer output dimensions of the proposed 2.5D spatio-sequential classification network, complementing the per-stage dimensions annotated in Figure~6 of the main manuscript. The network processes each vertebra as 15 slice groups of 5 adjacent slices, encodes each group with an EfficientNetV2-S backbone, models the resulting 15-token sequence with a two-layer transformer encoder, and predicts one fracture output per slice group.

\begin{table}[H]
    \centering
    \footnotesize
    \setlength{\tabcolsep}{4pt}
    \caption{Layer-by-layer output dimensions of the proposed CNN-Transformer classification network.}
    \label{tab:layer_dims}
    \begin{tabular}{@{} p{2.7cm} p{3.0cm} l p{4.0cm} @{}}
        \toprule
        Stage / Layer & Type / Operation & Output shape & Notes \\
        \midrule
        \multicolumn{4}{@{}l}{\textbf{Model input and output}} \\
        Input  & -- & $\mathrm{B}\times15\times5\times256\times256$ & Input CT slice groups \\
        Output & -- & $\mathrm{B}\times15$ & Fracture output per slice group \\
        \midrule
        \multicolumn{4}{@{}l}{\textbf{Stack input}} \\
        Input tensor         & -- & $15\times5\times256\times256$ & 15 slice groups; each group is 5 adjacent axial slices stacked as channels \\
        Reshape for backbone & -- & $15\times5\times256\times256$ & Each group is encoded independently; batch and slice-group dimensions are merged \\
        \midrule
        \multicolumn{4}{@{}l}{\textbf{EfficientNetV2-S backbone}} \\
        Input per group        & --                        & $5\times256\times256$   & 5 slices as channels \\
        conv\_stem + BN + act  & Conv $3\times3$, stride 2 & $24\times128\times128$  & -- \\
        blocks.0               & Fused-MBConv, stride 1    & $24\times128\times128$  & 2 layers \\
        blocks.1               & Fused-MBConv, stride 2    & $48\times64\times64$    & 4 layers \\
        blocks.2               & Fused-MBConv, stride 2    & $64\times32\times32$    & 4 layers \\
        blocks.3               & MBConv + SE, stride 2     & $128\times16\times16$   & 6 layers \\
        blocks.4               & MBConv + SE, stride 1     & $160\times16\times16$   & 9 layers \\
        blocks.5               & MBConv + SE, stride 2     & $256\times8\times8$     & 15 layers \\
        conv\_head + BN + act  & Conv $1\times1$           & $1280\times8\times8$    & -- \\
        global\_pool           & Global average pooling    & $1280$                  & Spatial dimensions removed \\
        Backbone output        & --                        & $15\times1280$          & 1280-dimensional feature vector per slice group \\
        \midrule
        \multicolumn{4}{@{}l}{\textbf{Transformer encoder}} \\
        Reshape to sequence  & --                & $1\times15\times1280$ & 15 tokens, one per slice group \\
        Positional encoding  & Sinusoidal PE     & $1\times15\times1280$ & Added to token embeddings \\
        Transformer layer 1  & MHSA + FFN        & $1\times15\times1280$ & 8 heads, head dim 160; FFN $1280\to5120\to1280$ \\
        Transformer layer 2  & MHSA + FFN        & $1\times15\times1280$ & Same as layer 1 \\
        Reshape for head     & --                & $15\times1280$        & Batch and slice-group dimensions are merged again \\
        \midrule
        \multicolumn{4}{@{}l}{\textbf{Classification head}} \\
        Linear      & FC $1280\to256$       & $15\times256$ & Applied independently to each token \\
        BatchNorm1d & BN over 256 features  & $15\times256$ & -- \\
        Dropout     & $p = 0.3$             & $15\times256$ & Applied during training \\
        LeakyReLU   & slope 0.1             & $15\times256$ & -- \\
        Linear      & FC $256\to1$          & $15\times1$   & -- \\
        Output      & Reshape + sigmoid     & $15$          & One fracture output per slice group \\
        \bottomrule
    \end{tabular}
\end{table}

\subsection{Per-Step Computational Runtime}
\label{sec:S10_3}

Table \ref{tab:runtime} reports the complete per-patient runtime of the pipeline at inference. The three core stages, detection, segmentation, and classification, are listed with their associated preprocessing and post-processing sub-steps. The CPU-bound DICOM loading and projection-generation sub-steps dominate the wall-clock time, while the three neural stages together account for roughly a fifth of the total, so the pipeline-level runtime is governed mainly by input and output handling rather than by the networks themselves.

\begin{table}[H]
    \centering
    \small
    \setlength{\tabcolsep}{5pt}
    \caption{Complete per-step computational runtime of the pipeline, measured per patient at inference, grouped by pipeline stage. Mean and standard deviation are reported in milliseconds.}
    \label{tab:runtime}
    \begin{tabular}{@{} p{5.6cm} l r r @{}}
        \toprule
        Step & Hardware & Mean (ms) & Std (ms) \\
        \midrule
        \multicolumn{4}{@{}l}{\textbf{Stage 1: Cervical spine VOI detection}} \\
        \hspace{1em}1.1 DICOM loading and preprocessing & CPU & 16,202 & 7,582 \\
        \hspace{1em}1.2 Projection generation & CPU & 18,395 & 3,703 \\
        \hspace{1em}1.3 Cervical spine VOI detection and extraction (3 views) & GPU & 7,202 & 1,476 \\
        \midrule
        \multicolumn{4}{@{}l}{\textbf{Stage 2: Cervical vertebra VOI segmentation}} \\
        \hspace{1em}2.1 Multi-label vertebra segmentation (2 views) & GPU & 900 & 222 \\
        \hspace{1em}2.2 Vertebra VOI extraction & CPU & 2,856 & 832 \\
        \midrule
        \multicolumn{4}{@{}l}{\textbf{Stage 3: Cervical vertebra fracture identification}} \\
        \hspace{1em}3.1 Fracture classification (2 models $\times$ 7 vertebrae) & GPU & 2,288 & 271 \\
        \hspace{1em}3.2 Ensemble fusion & CPU & $<$1 & -- \\
        \midrule
        \textbf{Total} & -- & 47,844 & 5,292 \\
        \bottomrule
    \end{tabular}
\end{table}

\section{Interobserver Variability}
\label{sec:S11}

Agreement was measured with Cohen's $\kappa$ for pairs of raters and Fleiss' $\kappa$ ($\kappa_F$) for more than two raters, defined as
\begin{equation}
    \kappa = \frac{p_{o} - p_{e}}{1 - p_{e}}, \qquad \kappa_{F} = \frac{\overline{P} - \overline{P_{e}}}{1 - \overline{P_{e}}},
\end{equation}
where $p_{o}$ is the observed and $p_{e}$ the chance-expected agreement between two raters, and $\overline{P}$ and $\overline{P_{e}}$ are the mean observed and mean expected agreement across all subjects. The figures below show the agreement between the expert radiologists, the reference standard (GT) and the model.

\begin{figure}[H]
    \centering
    \includegraphics[width = \textwidth]{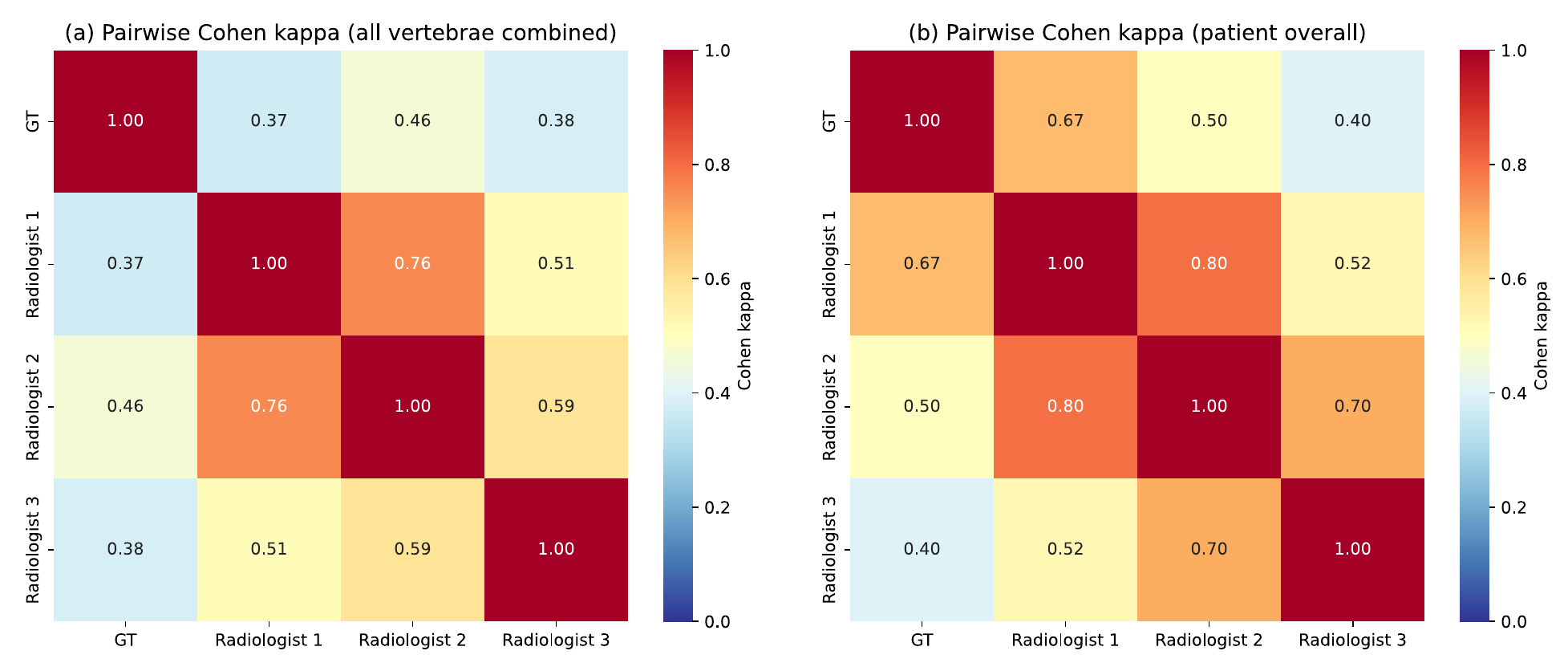}
    \caption{Pairwise Cohen's $\kappa$ among the three radiologists and the reference standard (GT), (a) across all vertebrae combined and (b) at the patient level (30 patients).}
    \label{fig:kappa_overall}
\end{figure}

\begin{figure}[htbp]
    \centering
    \includegraphics[width = \textwidth]{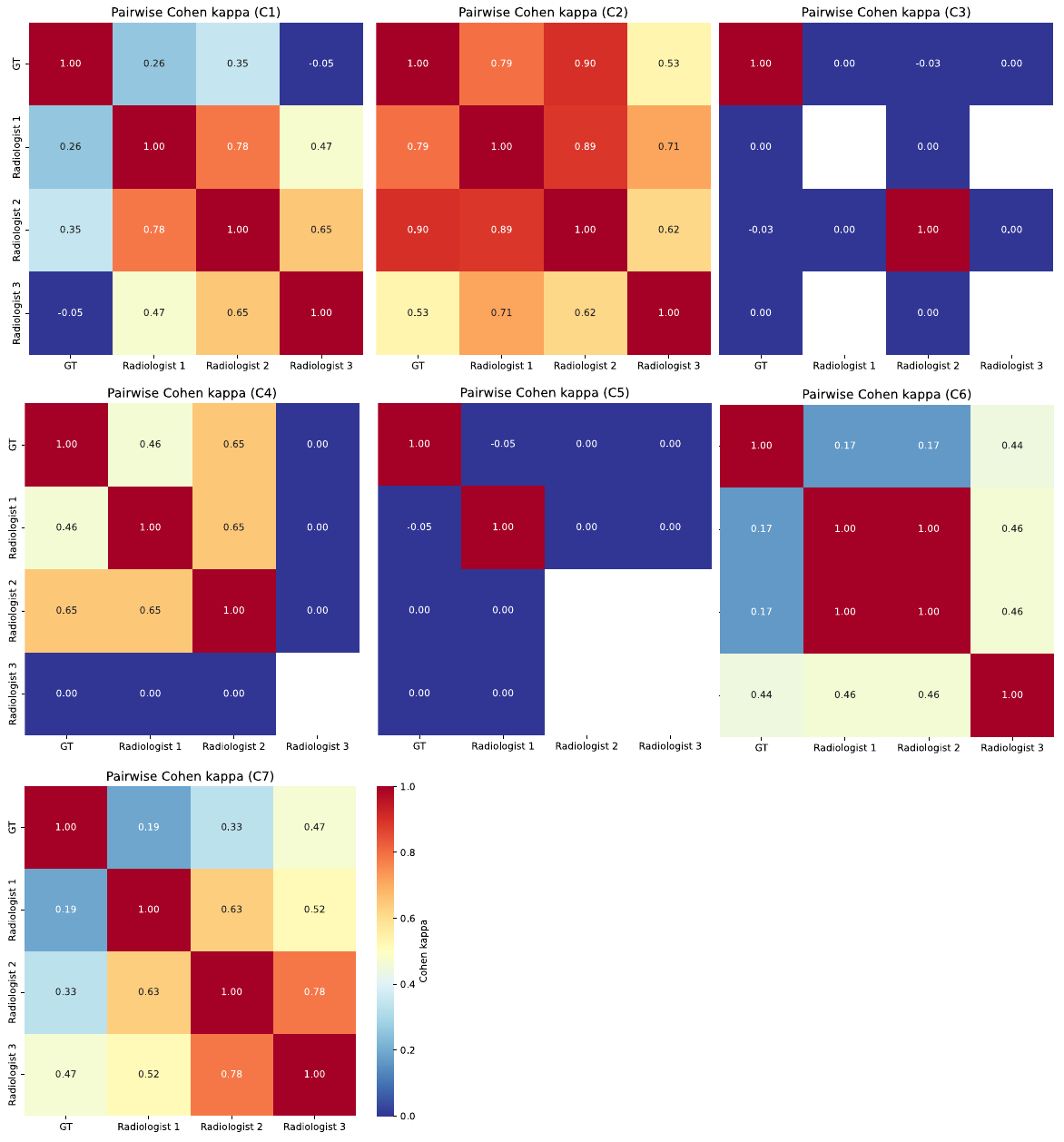}
    \caption{Pairwise Cohen's $\kappa$ among the three radiologists and the reference standard (GT) for C1--C7.}
    \label{fig:kappa_heatmaps}
\end{figure}

Figure~\ref{fig:kappa_overall} presents the pairwise Cohen's $\kappa$ across all vertebrae combined and at the patient level, and Figure~\ref{fig:kappa_heatmaps} gives the same comparison for each vertebra. The analysis reveals that radiologists often had higher agreement with each other than with the reference standard, particularly for the difficult mid-cervical vertebrae (C3--C5). Radiologist 3 demonstrated the most distinct interpretive style, often showing lower agreement with peers.

\begin{figure}[htbp]
    \centering
    \includegraphics[width = \textwidth]{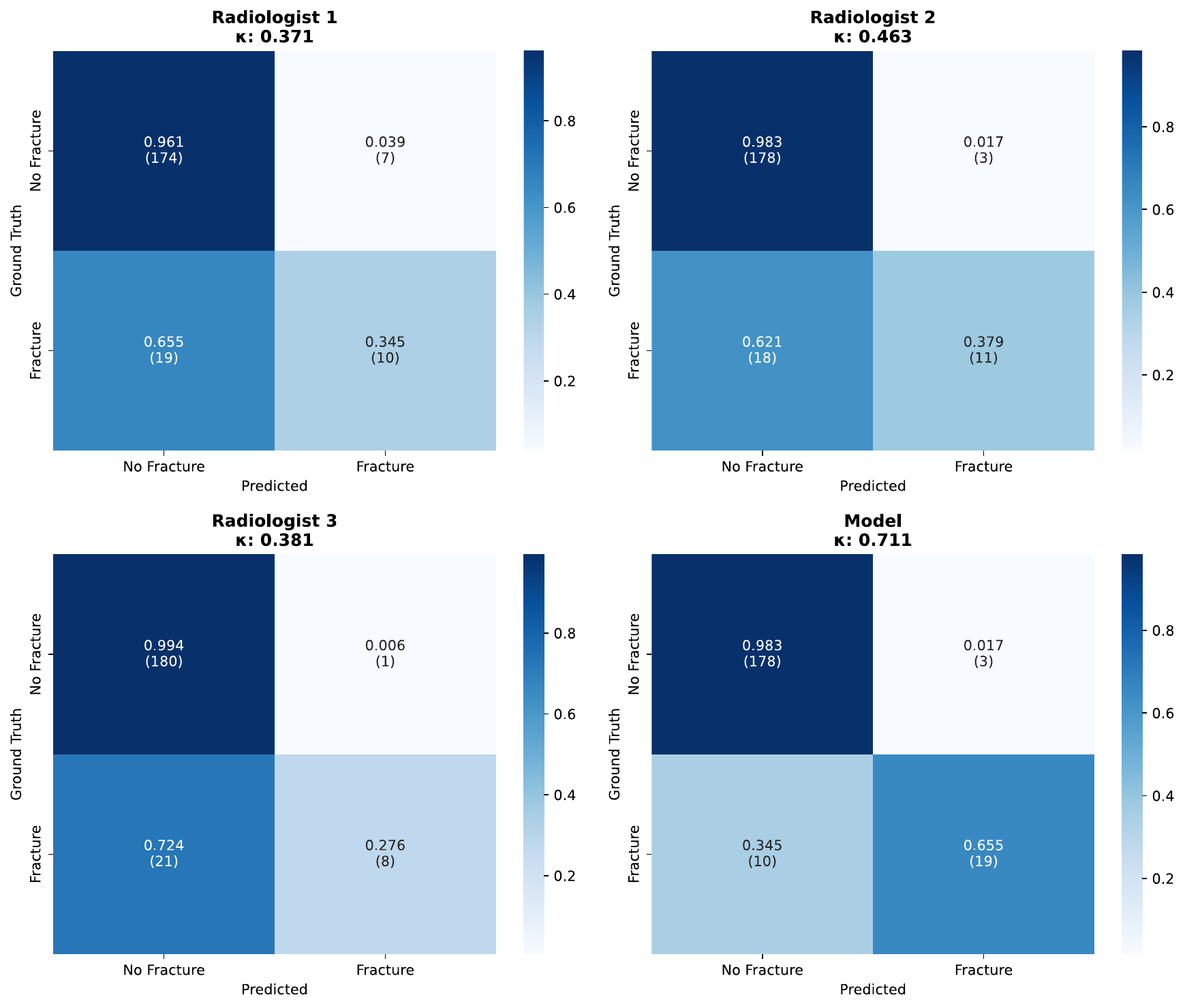}
    \caption{Confusion matrices comparing radiologist and model performance for cervical spine fracture detection at the vertebra level.}
    \label{fig:confusion_vertebra}
\end{figure}

Figure \ref{fig:confusion_vertebra} displays the confusion matrices for the model versus each radiologist against the reference standard at the vertebra level. The model correctly identified 19 of the 29 fractures in the test subset, whereas the radiologists identified 10, 11, and 8, respectively. The model therefore recovered more of the reference fractures than any radiologist on this subset.

\begin{figure}[htbp]
    \centering
    \includegraphics[width = \textwidth]{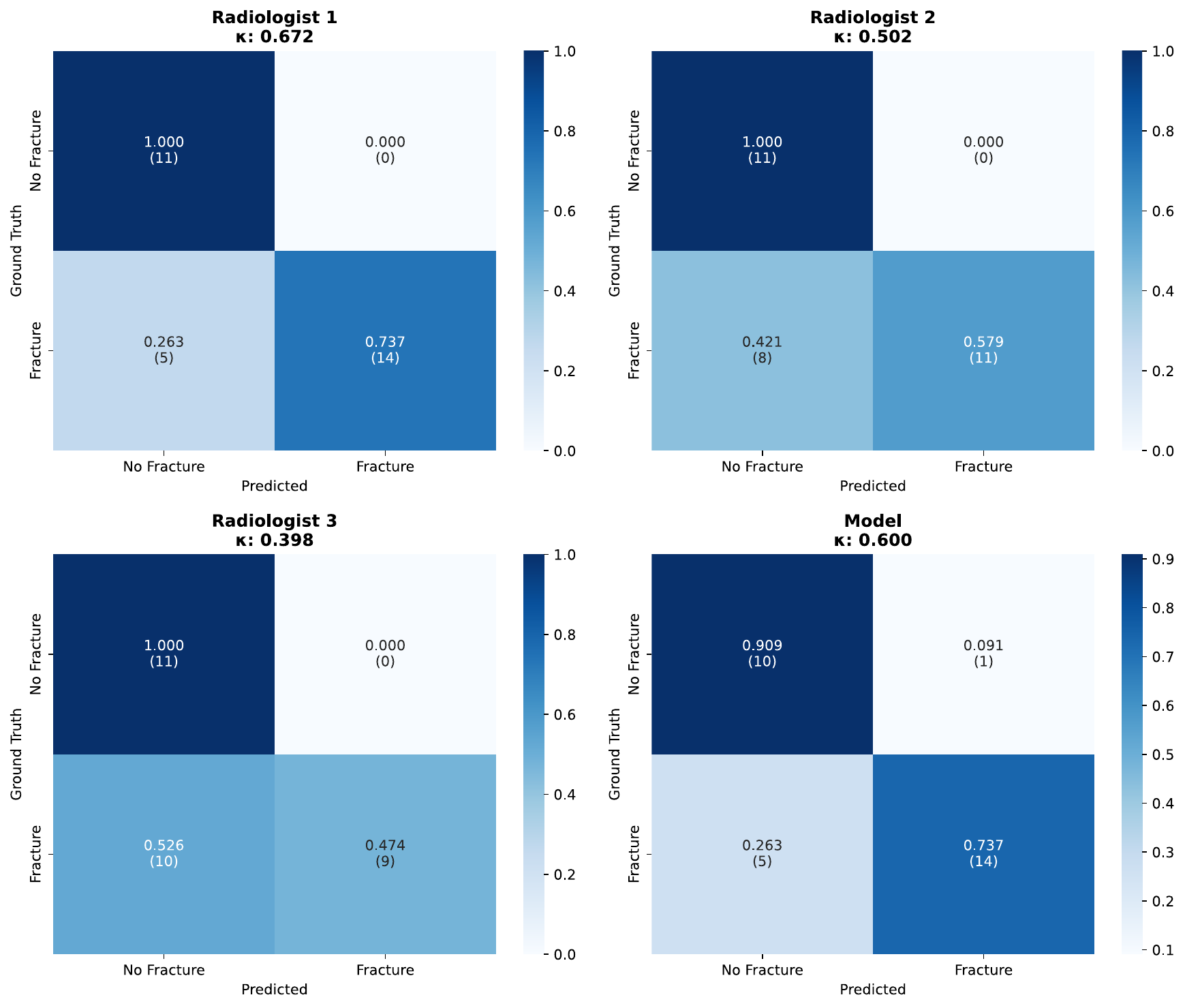}
    \caption{Confusion matrices comparing radiologist and model performance for cervical spine fracture detection at the patient level.}
    \label{fig:confusion_patient}
\end{figure}

Figure \ref{fig:confusion_patient} displays the confusion matrices at the patient level. The model correctly identified 14 of the 19 fractured patients, as many as Radiologist 1 and more than the other two.

\end{document}